\documentclass{article}

\PassOptionsToPackage{numbers, compress}{natbib}

\usepackage[final]{include/neurips_data_2021}

\usepackage[utf8]{inputenc} %
\usepackage[T1]{fontenc}    %
\usepackage{include/titletoc}
\usepackage{hyperref}       %
\usepackage{url}            %
\usepackage{booktabs}       %
\usepackage{amsfonts}       %
\usepackage{nicefrac}       %
\usepackage{microtype}      %
\usepackage{xcolor}         %

\usepackage[utf8]{inputenc} %
\usepackage[T1]{fontenc}    %
\usepackage{hyperref}       %
\usepackage{url}            %
\usepackage{booktabs}       %
\usepackage{amsfonts}       %
\usepackage{nicefrac}       %
\usepackage{xfrac}       %
\usepackage{microtype}      %
\usepackage{wrapfig}

\usepackage{pgfplotstable}
\usepackage{pgfplots}
\usepackage{collcell}
\usepackage{xstring}
\usepackage{graphicx}
\usepackage{tabu}
\usepackage{environ}
\usepackage{fp}
\usepackage{upgreek}
\usepackage{amsmath}
\usepackage{enumitem}
\usepackage{bm}
\usepackage{subfiles}
\usepackage{array,multirow}
\usepackage{caption}
\usepackage{makecell}
\usepackage{tikz, color}
\usetikzlibrary{matrix}
\usetikzlibrary{calc}
\usetikzlibrary{positioning}
\usetikzlibrary{graphs}
\usetikzlibrary{shapes.geometric}

\usepackage{xcolor,colortbl}
\definecolor{matlab-blue}{rgb}{0,0.4470,0.7410}
\definecolor{matlab-orange}{rgb}{0.8500,0.3250,0.0980}
\definecolor{matlab-yellow}{rgb}{0.9290,0.6940,0.1250}

\definecolor{matlab-green}{rgb}{0.4660,0.6740,0.1880}

\definecolor{matlab-red}{rgb}{0.6350,0.0780,0.1840}

\definecolor{limegreen}{HTML}{32CD32}

\DeclareMathOperator*{\argmax}{argmax}

\newcounter{daggerfootnote}

\usepackage{tabularx}  %
\usepackage{amsthm}  %
\usepackage{subcaption}  %

\usepackage[utf8]{inputenc}
\usepackage{booktabs} %

\usepackage{bbm} 
\usepackage{bm}
\usepackage{verbatim}
\usepackage{float}
\usepackage{color,soul}
\usepackage{enumitem}
\usepackage{mathtools}
\usepackage{hhline}
\usepackage[nameinlink]{cleveref} %
\usepackage[font=small,labelfont=bf,justification=justified,format=plain]{caption}
\usepackage{tikz}
\usepackage{wrapfig,booktabs}
\usepackage{xspace}
\usepackage{cancel}
\usepackage{sidecap}
\usepackage{placeins}

\graphicspath{ {../figures/} }

\DeclarePairedDelimiterX{\infdivx}[2]{(}{)}{%
  #1\;\delimsize\|\;#2%
}

\newcommand{\QF}[1]{\textcolor{blue}{QF}}

\newcommand{\edit}[1]{\textcolor{black}{#1}}
\newcommand{\finalblue}[1]{\textcolor{blue}{#1}}

\usepackage{ragged2e}

\newcommand{\bnn}{\textsc{bnn}\xspace}
\newcommand{\bnns}{\textsc{bnn}s\xspace}

\newcommand{\fsvi}{\textsc{fsvi}\xspace}
\newcommand{\mfvi}{\textsc{mfvi}\xspace}

\crefname{appsec}{appendix}{appendices}
\Crefname{appsec}{Appendix}{Appendices}

\usetikzlibrary{shapes.geometric, arrows, bayesnet, calc, positioning}

\newcommand{\bX}{\mathbf X}
\newcommand{\by}{\mathbf y}
\newcommand{\bx}{\mathbf x}

\newcommand{\calN}{\mathcal{N}}

\newcommand{\calI}{\mathcal{I}}

\newcommand{\bmu}{{\boldsymbol{\mu}}}
\newcommand{\btheta}{{\boldsymbol{\theta}}}

\newcommand{\closer}[3]{{\kern-#1ex{#2}\kern-#3ex}}

\DeclareMathOperator*{\argmin}{arg\,min}

\mathchardef\mhyphen="2D

\DeclareMathOperator{\E}{\mathbb{E}}

\usepackage{algorithm}
\usepackage{algorithmic}

\usepackage{tcolorbox}

\definecolor{azure}{rgb}{0.0, 0.5, 1.0}
\definecolor{airforceblue}{rgb}{0.36, 0.54, 0.66}
\definecolor{darkgreen}{rgb}{0.0, 0.2, 0.13}

\newcommand\defines{\,\dot{=}\,}

\newcommand{\map}{\textsc{map}\xspace}

\newcommand{\mcd}{\textsc{mc dropout}\xspace}

\newcommand{\retina}{\textsc{Retina}\xspace}

\newcommand{\vbar}{\,|\,}

\newcommand{\calD}{\mathcal{D}}

\newcommand{\calQ}{\mathcal{Q}}

\newcommand{\DD}{\mathbb{D}}

\newcommand{\bTheta}{\boldsymbol{\Theta}}

\newcommand{\pms}[1]{\ensuremath{{\scriptstyle\pm #1}}}

\usepackage{standalone}
\usepackage{tikz}
\usepackage{pgfplots}
\usepackage{pgf}
\usetikzlibrary{calc}
\usetikzlibrary{positioning}
\usetikzlibrary{angles,quotes}
\usetikzlibrary{backgrounds}
\usetikzlibrary{fit}
\usetikzlibrary{arrows}
\usetikzlibrary{arrows.meta}
\usetikzlibrary{shapes.symbols}
\usetikzlibrary{shadings}
\usetikzlibrary{shapes}
\usetikzlibrary{fadings}
\usetikzlibrary{bayesnet}
\usetikzlibrary{matrix}
\usetikzlibrary{plotmarks}
\usetikzlibrary{intersections}
\usetikzlibrary{pgfplots.fillbetween}
\pgfplotsset{compat=1.14}

\usepackage{colortbl}
\definecolor{mediumgray}{gray}{0.7}
\definecolor{lightgray}{gray}{0.85}
\definecolor{lightlightgray}{gray}{0.9}
\definecolor{C1}{HTML}{1F77B4}
\definecolor{C2}{HTML}{FF7F0E}
\definecolor{C3}{HTML}{2CA02C}
\definecolor{C4}{HTML}{D62728}
\definecolor{C5}{HTML}{9467BD}
\colorlet{C1light}{C1!70!white}
\colorlet{C2light}{C2!70!white}
\colorlet{C3light}{C3!70!white}
\colorlet{C4light}{C4!70!white}
\colorlet{C5light}{C5!70!white}
\colorlet{C1lighter}{C1!50!white}
\colorlet{C2lighter}{C2!50!white}
\colorlet{C3lighter}{C3!50!white}
\colorlet{C4lighter}{C4!50!white}
\colorlet{C5lighter}{C5!50!white}
\colorlet{C1vlight}{C1!20!white}
\colorlet{C2vlight}{C2!20!white}
\colorlet{C3vlight}{C3!20!white}
\colorlet{C4vlight}{C4!20!white}
\colorlet{C5vlight}{C5!20!white}
\colorlet{linkcolor}{violet}

\usepackage{siunitx}  %

\usepackage[american]{babel}

\usepackage{natbib} %
\usepackage{tikz}
\usepackage{arydshln}
\usepackage{todonotes}

\usepackage[title]{appendix}

\hypersetup{
    colorlinks,
    citecolor=[rgb]{0.00, 0.45, 0.70},
    linkcolor=[rgb]{0.01, 0.62, 0.45},
    urlcolor=[rgb]{0.00, 0.45, 0.70},
    }

\title{
    Benchmarking Bayesian Deep Learning on\\Diabetic Retinopathy Detection Tasks
}

\author{%
  Neil Band\thanks{Equal contribution.}~~\thanks{Corresponding authors: \href{mailto:neil.band@cs.ox.ac.uk}{\texttt{neil.band@cs.ox.ac.uk}} and \href{mailto:neil.band@cs.ox.ac.uk}{\texttt{tim.rudner@cs.ox.ac.uk}}.}
  \\
  University of Oxford
  \And
  Tim G. J. Rudner$^\ast$$^\dagger$
  \\
  University of Oxford
  \And
  Qixuan Feng
  \\
  University of Oxford
  \AND
  Angelos Filos
  \\
  University of Oxford
  \And
  Zachary Nado
  \\
  Google Research
  \And
  Michael W. Dusenberry
  \\
  Google Research
  \And
  Ghassen Jerfel
  \\
  Google Research
  \AND
  Dustin Tran
  \\
  Google Research
  \And
  Yarin Gal
  \\
  University of Oxford
}

\begin{document}

\maketitle

\vspace*{-10pt}
\begin{abstract}
Bayesian deep learning seeks to equip deep neural networks with the ability to precisely quantify their predictive uncertainty, and has promised to make deep learning more reliable for safety-critical real-world applications.
Yet, existing Bayesian deep learning methods fall short of this promise; new methods continue to be evaluated on unrealistic test beds that do not reflect the complexities of downstream real-world tasks that would benefit most from reliable uncertainty quantification.
We propose the \finalblue{\textbf{\retina Benchmark},} a set of real-world tasks that accurately reflect such complexities and are designed to assess the reliability of predictive models in safety-critical scenarios.
Specifically, we curate two publicly available datasets of high-resolution human retina images exhibiting varying degrees of diabetic retinopathy, a medical condition that can lead to blindness, and use them to design a suite of automated diagnosis tasks that require reliable predictive uncertainty quantification.
We use these tasks to benchmark well-established and state-of-the-art Bayesian deep learning methods on task-specific evaluation metrics.
We provide an easy-to-use codebase for fast and easy benchmarking following reproducibility and software design principles.
We provide implementations of all methods included in the benchmark as well as results computed over 100 TPU days, 20 GPU days, 400 hyperparameter configurations, and evaluation on at least 6 random seeds each.
\end{abstract}

\vspace*{-5pt}
\section{Introduction}
\label{sec:introduction}

Bayesian deep learning has been applied successfully to a wide range of real-world prediction problems such as \textit{medical diagnosis}~\citep{ching2018opportunities,kamnitsas2017efficient,leibig2017leveraging,worrall2016automated}, \textit{computer vision}~\citep{kampffmeyer2016semantic, kendall2016modelling, kendall2015bayesian}, \textit{scientific discovery}~\citep{levasseur2017uncertainties, mcgibbon2017improving}, and \textit{autonomous driving}~\citep{amodei2016concrete,filos2020can,kahn2017uncertainty,kendall2016modelling, KendallGal2017Uncertainties,kendall2015bayesian, malinin2021shifts}.

Despite the demonstrated usefulness of Bayesian deep learning for such practical applications and a growing literature on inference methods~\citep{blundell2015mfvi,farquhar2020radial,gal2016dropout,graves2011practical,neklyudov2018variance,osawa2019practical,rudner2021fsvi,wen2018flipout,wu2018deterministic}, there exists no standardized benchmarking task that reflects the complexities and challenges of safety-critical real-world tasks while adequately accounting for the reliability of models' predictive uncertainty estimates.

To make meaningful progress in the development and successful deployment of reliable Bayesian deep learning methods, we need easy-to-use benchmarking tasks that reflect the real world and hence serve as a legitimate litmus test for practitioners that aim to deploy their models in safety-critical settings.
Further, such tasks ought to be usable without the extensive domain expertise often necessary for appropriate experiment design and data preprocessing.
Lastly, any such benchmarking task must include evaluation methods that test for predictive performance and assess different properties of models' predictive uncertainty estimates, while taking into account application-specific constraints.

\begin{figure}[htb!]
\vspace{-5pt}
\centering
\begin{subfigure}[c]{0.95\linewidth}
  \includegraphics[height=0.127\linewidth]{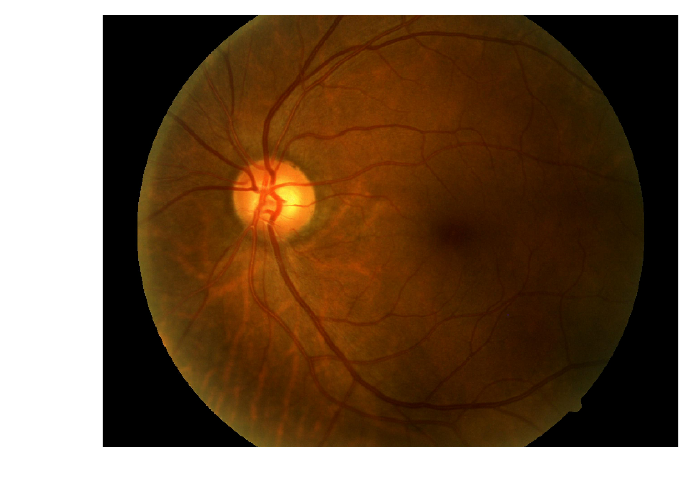}
  \includegraphics[height=0.127\linewidth]{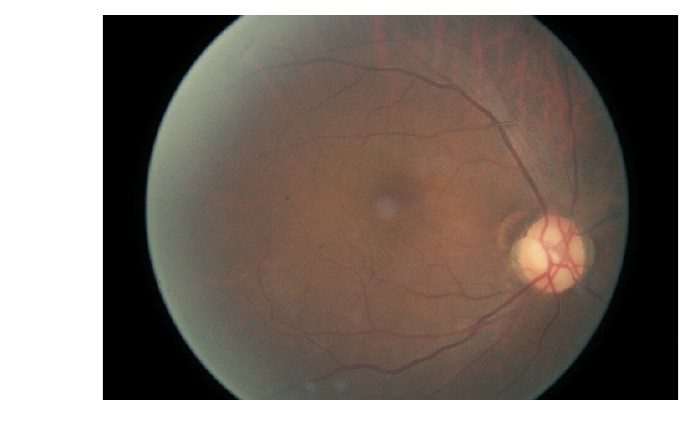}
  \includegraphics[height=0.127\linewidth]{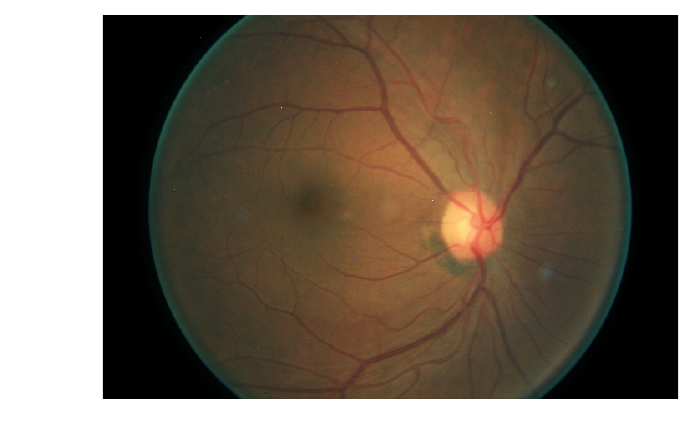}
  \includegraphics[height=0.127\linewidth]{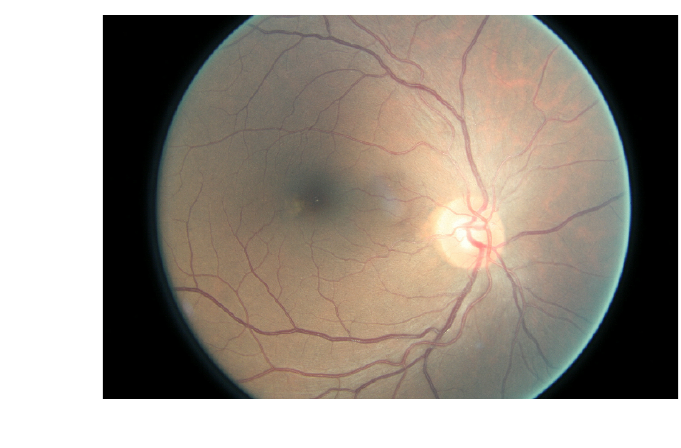}
  \includegraphics[height=0.127\linewidth]{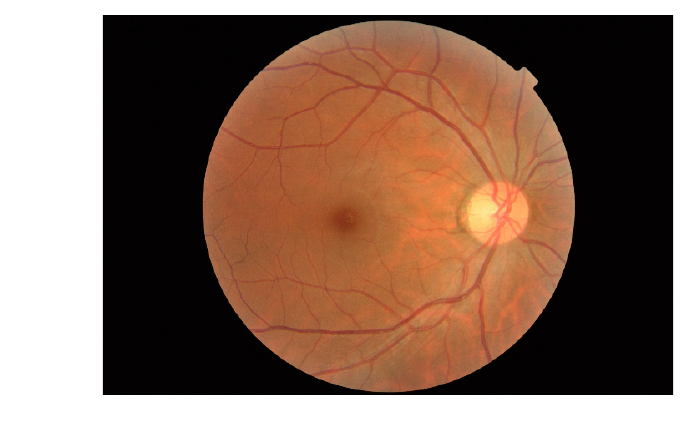}
  \caption{Retina images exhibiting non-sight-threatening diabetic retinopathy ($y=0$).}
\end{subfigure}
\\
\vspace*{5pt}
\centering
\begin{subfigure}[c]{0.95\linewidth}
  \includegraphics[height=0.127\linewidth]{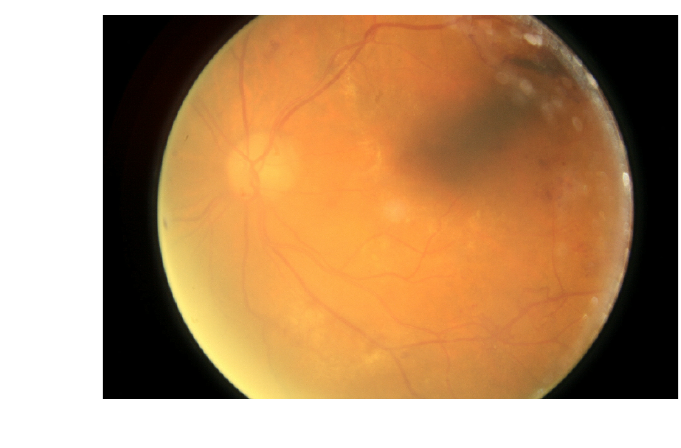}
  \includegraphics[height=0.127\linewidth]{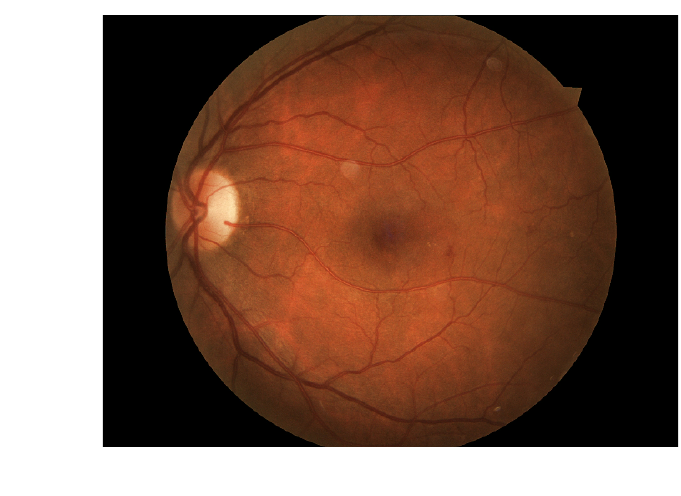}
  \includegraphics[height=0.127\linewidth]{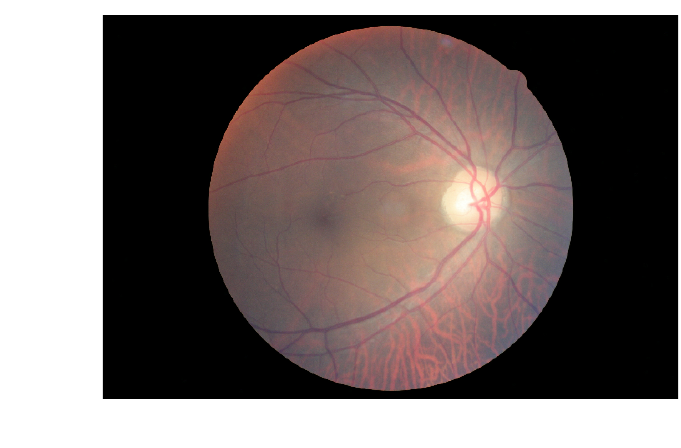}
  \includegraphics[height=0.127\linewidth]{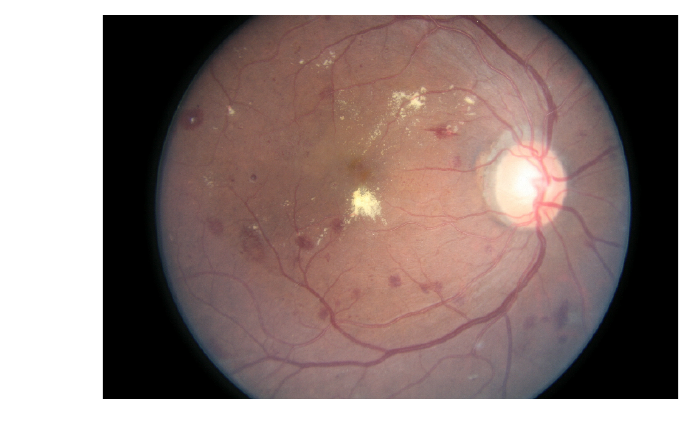}
  \includegraphics[height=0.127\linewidth]{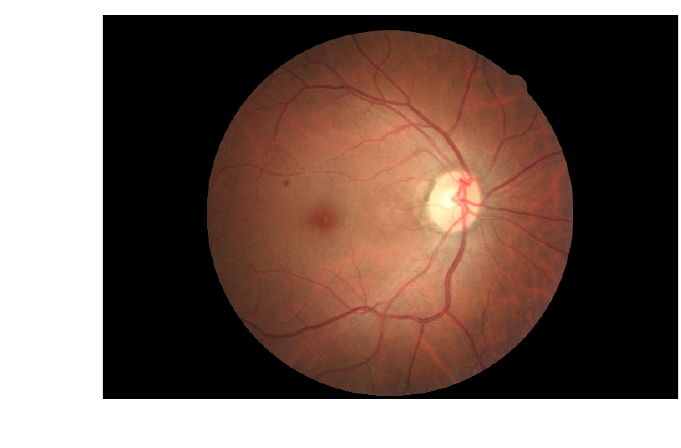}
  \caption{Retina images exhibiting sight-threatening diabetic retinopathy ($y=1$).}
\end{subfigure}
\caption{
	Samples of retina scans from the EyePACS dataset showing varying degrees of diabetic retinopathy.}
\label{fig:samples}
\vspace*{10pt}
\end{figure}
\vspace{-14pt}

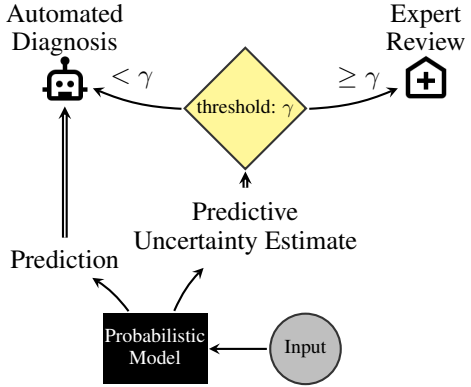
\begin{wrapfigure}{r}{0.445\textwidth}
\vspace{-14pt}
\centering
\begin{tikzpicture}[scale=0.8, >=stealth]

\tikzstyle{data}=[
  circle,
  fill =black!25,
  inner sep=1pt,
  minimum size = 12.5mm,
  thick, draw =black!80,
  node distance = 20mm,
  scale=0.75]

\tikzstyle{model}=[
  rectangle,
  fill =black,
  inner sep=1pt,
  text=white,
  minimum size = 12.5mm,
  draw=none,
  node distance = 20mm,
  scale=0.75]

\tikzstyle{thres}=[
  diamond,
  fill =yellow!50,
  inner sep=1pt,
  minimum size = 12.5mm,
  thick,
  draw =black!80,
  node distance = 20mm,
  scale=0.75]

\tikzstyle{directed}=[
  ->,
  thick,
  shorten >=0.5 pt,
  shorten <=1 pt]
  
\tikzstyle{image}=[
  inner sep=0pt]

\node[model] at (12.5,2.0) (model)     {\makecell[c]{Probabilistic\\Model}};
\node[data]  at (15.0,2.0) (x_test)    {Input};
\node        at (11.0,3.5) (y_pred)    {Prediction};
\node        at (14.0,4.0) (y_uncert)  {\makecell[c]{Predictive\\Uncertainty Estimate}};
\node[thres] at (14.0,6.0) (threshold) {threshold: $\gamma$};
\node[image] at (11.0,6.5) (machine)   {\includegraphics[width=.05\textwidth]{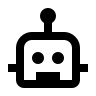}};
\node[right = 0.125cm of machine]
                          (less)      {$< \gamma$};
\node[image] at (17.0,6.5) (hospital)  {\includegraphics[width=.05\textwidth]{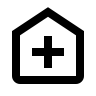}};
\node[left = 0.125cm of hospital]
                          (more)      {$\geq \gamma$};
\node[image] at (17.0,7.3) (hospital_descr)  {\makecell[c]{Expert\\Review}};
\node[image] at (11.0,7.3) (more_descr)  {\makecell[c]{Automated\\Diagnosis}};

\path
        (x_test)    edge [directed]                 (model)
        (model)     edge [directed, bend right=10]  (y_pred)
        (model)     edge [directed, bend left=10]   (y_uncert)
        (y_uncert)  edge [directed, double]         (threshold)
        (threshold) edge [directed, bend left=10]   (machine)
        (threshold) edge [directed, bend right=10]  (hospital)
        (y_pred)    edge [directed, double]         (machine)
        ;
\end{tikzpicture}

\caption{
    Automated diagnosis pipeline:
    For a given input, a model provides a prediction and a corresponding uncertainty estimate;
    if the uncertainty estimate is below a certain reference threshold $\gamma$ (indicating a low degree of uncertainty) the diagnosis is processed without further review; otherwise, it is referred to a medical expert.
}
\label{fig:diagnosis}
\vspace*{-15pt}
\end{wrapfigure}

In this paper, we propose a set of realistic safety-critical downstream tasks that respect these desiderata and use them to benchmark well-established and state-of-the-art Bayesian deep learning methods.
To do so, we consider the problem of using machine learning to detect diabetic retinopathy, a medical condition considered the leading cause of vision impairment and blindness~\citep{Sabanayagam2019incidence}.
Unlike in prior works on diabetic retinopathy detection, the benchmarking tasks presented in this paper are specifically designed to assess the reliability of machine learning models and the quality of their predictive uncertainty estimates using both aleatoric and epistemic uncertainty estimates.

Medical diagnosis problems are particularly well-suited to assess reliability due to the severe harm caused by predictive models that make confident but poor predictions (for example, when a disease is not recognized).
As a general desideratum, we want a model's predictive uncertainty to correlate with the correctness of its predictions.
Good predictive uncertainty estimates can be a fail-safe against incorrect predictions.
If a given data point might result in an incorrect prediction because it is meaningfully different from data in the training set---for example, because it shows signs of the disease not captured there, exhibits visual artifacts, or was obtained using different measurement devices---a good predictive model will express a high level of predictive uncertainty and flag the example for further review by a medical expert.

\textbf{Contributions.}$~~~$
We present the \finalblue{\textbf{\retina Benchmark}}: an easy-to-use, expert-guided, open-source \emph{suite of diabetic retinopathy detection benchmarking tasks} for Bayesian deep learning.
In particular, we design safety-critical downstream tasks from publicly available datasets.
On these downstream tasks, we evaluate well-established and state-of-the-art Bayesian and non-Bayesian methods on a set of task-specific reliability and performance metrics.
Lastly, we provide a modular and extensible implementation of the benchmarking tasks and methods, as well as pre-trained models obtained from an extensive hyperparameter optimization over more than 400 total configurations and evaluation, using over 100 TPU days and 20 GPU days of compute.
Code to reproduce our results and benchmark new methods is available at:
\vspace*{5pt}
\begin{tcolorbox}
\small
\begin{center}
\href{https://github.com/google/uncertainty-baselines/tree/main/baselines/diabetic_retinopathy_detection}{\texttt{github.com/google/uncertainty-baselines/.../diabetic\_retinopathy\_detection}}.
\end{center}
\end{tcolorbox}

\clearpage

\section{Downstream Benchmarking Tasks for Diabetic Retinopathy Detection}
\label{sec:tasks}

In this section, we present two real-world scenarios in diabetic retinopathy detection and describe how we merge two publicly available datasets to design corresponding prediction tasks.

\subsection{Data and Preprocessing}
\label{subsec:data}

\begin{wrapfigure}{r}{0.27\textwidth}
\vspace*{-25pt}
\centering
\begin{minipage}{\linewidth}
\includegraphics[width=\linewidth]{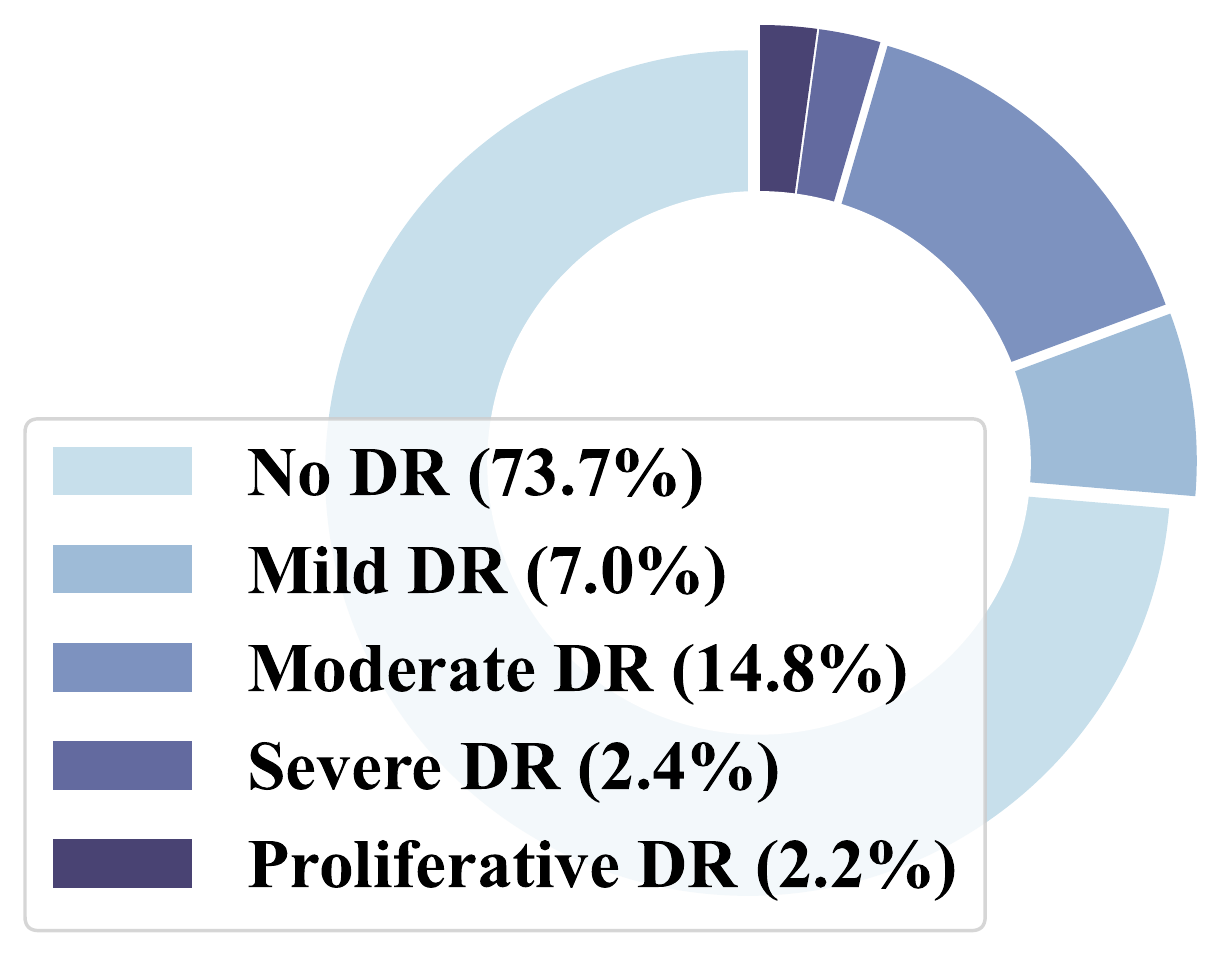}
\subcaption{
    EyePACS~\citep{kaggle_2015}.
}
\end{minipage}
\begin{minipage}{\linewidth}
\vspace*{5pt}
\includegraphics[width=\linewidth]{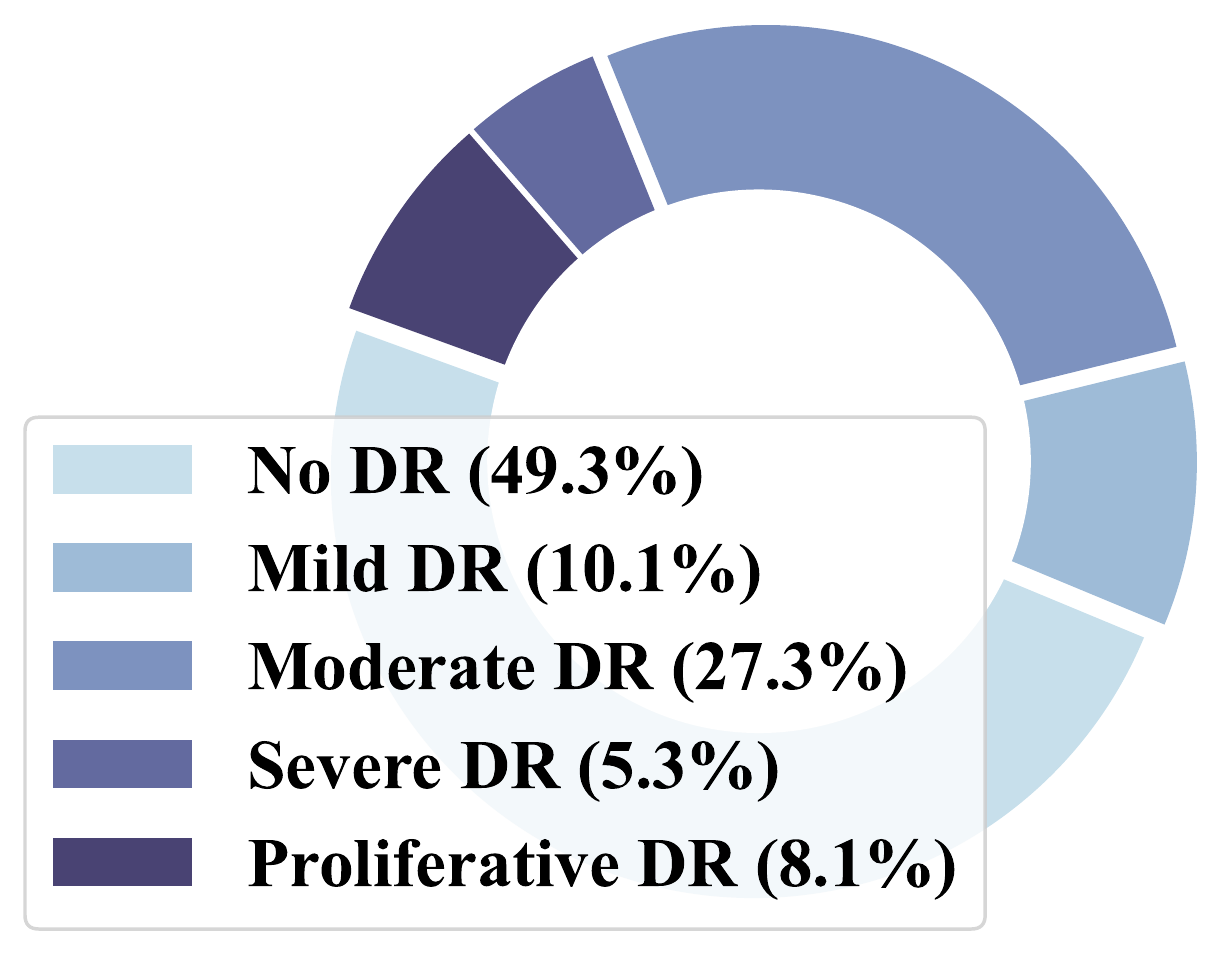}
\subcaption{
    APTOS~\citep{APTOS_2019}.
}
\end{minipage}
\caption{
    Data class labels.
}
\vspace*{-15pt}
\label{fig:labels}
\end{wrapfigure}

\textbf{EyePACS Dataset.}$~~~$
\label{subsec:datasets}
We construct training datasets for different tasks from the EyePACS dataset, previously used for the Kaggle Diabetic Retinopathy Detection Challenge~\citep{kaggle_2015}.
It contains high-resolution labeled images of human retinas exhibiting varying degrees of diabetic retinopathy.
The dataset consists of 35,126 training, 10,906 validation, and 42,670 test images, each an RGB image of a human retina graded by a medical expert on the following scale: 0 (no diabetic retinopathy), 1 (mild diabetic retinopathy), 2 (moderate diabetic retinopathy), 3 (severe diabetic retinopathy), and 4 (proliferative diabetic retinopathy).

\textbf{APTOS Dataset.}$~~~$
To construct tasks that assess model performance under distribution shift, we use the APTOS 2019 Blindness Detection dataset~\citep{APTOS_2019}.
The dataset also contains labeled images of human retinas exhibiting varying degrees of diabetic retinopathy, but was collected in India, from a different patient population, using different medical equipment.
We use 80\% of the images (2,929 images) as a test set and the other 20\% (733 images) as a secondary validation set.
Moreover, the images are significantly noisier than the images in the EyePACS dataset, with distinct visual artifacts (cf.~\Cref{fig:aptos_images},~\Cref{subsec:app_input_data_examples}).
Each image was graded on the same 0-to-4 scale as the EyePACS dataset.

\textbf{Prediction Targets.}$~~~$
We follow~\citet{leibig2017leveraging} and binarize \edit{all examples from both the EyePACS and APTOS datasets} by dividing the classes up into sight-threatening diabetic retinopathy---defined as moderate diabetic retinopathy or worse (classes $\{2,3,4\}$)---and non-sight-threatening diabetic retinopathy---defined as no or mild diabetic retinopathy (classes $\{0,1\}$).
By international guidelines, this is the threshold at which a case should be referred to an ophthalmologist~\citep{wong2016guidelines}.
Example EyePACS retina images from the two classes are shown in~\Cref{fig:samples}.
Reflecting real-world challenges, \edit{the datasets are} unbalanced---e.g., for EyePACS, only $19.6\%$ of the training set and $19.2\%$ of the test set have a positive label---and images have visual artifacts and noisy labels (some labels are incorrect).

\textbf{Data Preprocessing.}$~~~$
Data preprocessing on examples from both the EyePACS and APTOS datasets follows the winning entry of the Kaggle Challenge~\cite{kaggle_2015}: 
Images are rescaled such that retinas have a radius of 300 pixels, are smoothed using local Gaussian blur, and finally, are clipped to 90\% size to remove boundary effects.
Examples of original and corresponding processed images are provided in \Cref{fig:processed} (\Cref{subsec:app_input_data_examples}).
We conduct an empirical study investigating how varying the strength of the Gaussian blur smoothing affects downstream performance and uncertainty quality in \Cref{subsec:app_preprocessing_exps}.

\vspace*{-2pt}
\subsection{Diabetic Retinopathy Detection under Severity Shift}
\label{subsec:severity_shift}

Diabetes and diabetes-related illnesses such as diabetic retinopathy are becoming widespread.
Yet cases of sight-threatening diabetic retinopathy are still relatively rare, and scans of retinas exhibiting signs of no or mild diabetic retinopathy are more easily obtainable.
As a result, predictive models for detecting diabetic retinopathy may be trained on only a very small number of retina images showing signs of severe or proliferative retinopathy.

We design a prediction task that simulates this setting and allows us to assess the reliability of predictive models when they are evaluated on images that have been assigned a severity higher than any encountered in the training data.
Specifically, we train models only on retina images showing signs of at most moderate diabetic retinopathy and evaluate them on retina images showing signs of severe or proliferative diabetic retinopathy.
Given that many signs of moderate diabetic retinopathy are similar in appearance to signs of severe or proliferative diabetic retinopathy (just weaker), we would expect a good predictive model to be able to correctly classify the latter, but to exhibit increased predictive uncertainty.
There are certain features of diabetic retinopathy progression that are unique to more severe cases, such as vitreous hemorrhage, or bleeding into the vitreous humor \citep{elannan2014current}. 
However, we consider uncertainty-aware downstream tasks that tolerate such unfamiliar cases (cf.~\Cref{subsec:downstream_task}).

In this \textit{Severity Shift} task, we partition the EyePACS dataset into a subset containing all retina images labeled as no, mild, or moderate diabetic retinopathy (original classes $\{0,1,2\}$) and a subset of retina images labeled as severe or proliferative diabetic retinopathy (original classes $\{3,4\}$). 
Next, the samples in each subset are binarized (cf.~\Cref{subsec:data}): The subset of retina images showing signs of at most moderate diabetic retinopathy (subset ``moderate'') contains images of binarized classes $\{0,1\}$; and the subset of retina images showing signs of severe or proliferative diabetic retinopathy (subset ``severe'') only contains the binarized class 1.
This results in 33,545 images in the training set, and 40,727 and 3,524 images in the in-domain and distributionally shifted evaluation sets, respectively.

\vspace*{-2pt}
\subsection{Diabetic Retinopathy Detection under Country Shift}
\label{subsec:population_shift}

Similar to the scarcity of scans of sight-threatening diabetic retinopathy, the availability of retina scans is limited in countries without widespread screening.
Hence, a predictive model may be trained on images collected in the United States---where many scans are performed---and used to evaluate scans from another country, where scans are rarer and performed using different medical devices.

We design a prediction task that simulates this setting and allows us to evaluate the reliability of predictive models when the training and test data are not obtained from the same patient population nor collected with the same medical equipment.
In this \textit{Country Shift} task, we train models on retina images from the EyePACS dataset and evaluate them on retina images from the APTOS dataset.
We use the entire training and test data provided in the EyePACS dataset and convert the task into binary classification as described in \Cref{subsec:data}.
This results in 35,126 images in the training set, and 42,670 and 2,929 images in the in-domain and distributionally shifted evaluation sets, respectively.

\vspace*{-2pt}
\subsection{Downstream Task: Selective Prediction and Expert Referral}
\label{subsec:downstream_task}

In real-world settings where the evaluation data may be sampled from a shifted distribution, incorrect predictions may become increasingly likely.
To account for that possibility, predictive uncertainty estimates can be used to identify datapoints where the likelihood of an incorrect prediction is particularly high and refer them for further review as described in~\Cref{fig:diagnosis}.
We consider a corresponding selective prediction task, where the predictive performance of a given model is evaluated for varying expert referral rates.
That is, for a given referral rate of $\tau \in [0, 1]$, a model's predictive uncertainty is used to identify the $\tau$ proportion of images in the evaluation set for which the model's predictions are most uncertain.
Those images are referred to a medical professional for further review, and the model is assessed on its predictions on the remaining $(1 - \tau)$ proportion of images.
By repeating this process for all possible referral rates and assessing the model's predictive performance on the retained images, we estimate how reliable it would be in a safety-critical downstream task, where predictive uncertainty estimates are used in conjunction with human expertise to avoid harmful predictions.

Importantly, selective prediction tolerates out-of-distribution examples.
For example, even if unfamiliar vitreous hemorrhages appear in certain \textit{Severity Shift} images (cf. \Cref{subsec:severity_shift}), a model with reliable uncertainty estimates will perform better in selective prediction by assigning these images high epistemic (and predictive) uncertainty, therefore referring them to an expert at a lower $\tau$.
\Cref{subsec:app_sel_pred_details} discusses best- and worst-case uncertainty estimates for the selective prediction task.

To assess how well different models' predictive uncertainty estimates can be used to separate correct from incorrect diagnoses, we perform selective prediction on three different evaluation settings for the prediction problems described in Sections~\ref{subsec:severity_shift}~and~\ref{subsec:population_shift}, to account for the possibility that the evaluation dataset may contain samples from the in-domain distribution, a shifted distribution, or both.

\subsection{Model Diagnostic: Predictive Uncertainty Histograms}
\label{subsec:pred_uncert_hist}

We may also investigate how a model's predictive uncertainty estimates vary with respect to the ground-truth clinical label (0-to-4).
For each task (\emph{Country} or \textit{Severity Shift}) and each uncertainty quantification method (cf. \Cref{sec:methods}), we bin examples by their ground-truth clinical label.
Then, for each (task, method, clinical label) tuple, we plot the distribution of predictive uncertainty estimates for correctly and incorrectly predicted examples (in blue and red, respectively).
See~\Cref{sec:app_single_model_pred_uncert_hist} for further setup details and plots for both tasks.
A model that produces reliable uncertainty estimates should assign low predictive uncertainty to examples that it classifies correctly (the blue distribution should have most of its mass near $x=0$) and high predictive uncertainty to examples that it classifies incorrectly (the red distribution should have its mass concentrated at a higher $x$-value).

\section{Related Work}
\label{sec:related_work}

\retina builds on prior works that demonstrated the usefulness of predictive uncertainty estimates in diabetic retinopathy detection and related downstream tasks~\citep{leibig2017leveraging}.
We significantly extend the empirical evaluation in~\citet{leibig2017leveraging} by designing new prediction problems and corresponding safety-critical downstream tasks for diabetic retinopathy detection, benchmarking a wide array of Bayesian deep learning methods, and providing a modular, extensible, and easy-to-use codebase.
We also significantly extend~\citet{filos2019systematic} (of which this paper is a direct extension; with contributions from some of the authors), which does not consider severity shifts, only compares two variational inference methods, uses an outdated neural network architecture (with only $\approx$10\% of the parameters of the ResNet-50 architecture used in this work), and considers only a small subset of the evaluation procedures included in \retina (cf.~\Cref{subsec:more_experiments} for the full set of results).

Previous works have evaluated methods by predictive performance and quality of their predictive uncertainty estimates on curated datasets such as CIFAR-10 and FashionMNIST~\citep{osawa2019practical,ovadia2019uncertainty,henne2020benchmarking,rudner2021fsvi}.
Some prior works provide datasets and benchmarks for robustness and uncertainty quantification in real-world settings but have significant shortcomings.
\citet{le2018uncertainty} considers object detection using a real-world dataset~\citep{Geiger2012CVPR} but benchmarks only two methods, neither of which can quantify epistemic uncertainty (cf.~\Cref{sec:uncertainty_estimation}), and does not consider distribution shifts. 
Other works \citep{blum2019fishyscapes, Feng_2021} use methods which quantify both epistemic and aleatoric uncertainty, and consider distribution shifts, but use performance metrics which do not assess quality of uncertainty estimates, such as average precision and log-likelihood (cf.~\Cref{subsubsec:country_shift}).
Finally,~\citet{koh2021wilds} considers real-world datasets in domain adaptation problems, but restrictively assumes that the training data is composed of multiple training distributions with domain labels, and does not take into account models' predictive uncertainty.

In contrast, \retina \textbf{(i)} considers real-world safety-critical tasks and accompanying uncertainty-aware metrics in an important application domain, \textbf{(ii)} is composed of large amounts of high-dimensional data (>$80$ GB), \textbf{(iii)} compares a larger set of methods than prior works and incorporates both aleatoric and epistemic uncertainty, and is implemented in adherence to the \textit{Uncertainty Baselines} repository\footnote{See \url{https://github.com/google/uncertainty-baselines}.} practices for easy future use and extension, making it easier to benchmark other Bayesian deep learning methods not only on the tasks presented but also on a range of other datasets.

\section{Uncertainty Estimation}
\label{sec:uncertainty_estimation}

Predictive models' total uncertainty can be decomposed into aleatoric and epistemic uncertainty.
A model's aleatoric uncertainty is an estimate of the uncertainty inherent in the data (e.g., due to noisy inputs or targets), whereas a model's epistemic uncertainty is an estimate of the uncertainty due to constraints on the model (e.g., due to model misspecification) or the training process (e.g., due to convergence to bad local optima)~\citep{depeweg2018decomposition}.
Optimal uncertainty estimates would be perfectly correlated with the model error.
Hence, because both aleatoric and epistemic uncertainty may contribute to an incorrect prediction, total uncertainty is our uncertainty measure of choice.
For a model with stochastic parameters $\bTheta$, pre-likelihood outputs $f(\bX ; \bTheta)$, and a likelihood function $p(\by_\ast \vbar \bx_\ast ;\, \btheta)$, the model's predictive uncertainty can be decomposed as
\begin{align}
\label{eq:uncertainty}
    \underbrace{\mathcal{H}(\E [ p(\by_\ast \vbar f(\bx_\ast ; \btheta)) ] ) }_{\text{Total Uncertainty}}
    = \underbrace{ \E [ \mathcal{H}( p(\by_\ast \vbar f(\bx_\ast ; \btheta) )) ]}_{\text{Aleatoric Uncertainty}} \hspace*{4pt}+\hspace*{-8pt} \underbrace{\mathcal{I}(\by_\ast ;\, \bTheta)}_{\text{Epistemic Uncertainty}} \hspace{-10pt},
\end{align}
where the expectation is taken with respect to the distribution over the model parameters, $\mathcal{H}(\cdot)$ is the entropy functional, and $\mathcal{I}(\by_\ast ;\, \bTheta)$ is the mutual information between the model parameters and its predictions~\citep{cover1991elements,shannon1949mathematical}.

In binary classification settings with classes $\{0,1\}$, the total predictive uncertainty is given by
\begin{align}
\label{eq:predictive-entropy}
    \mathcal{H}(\E [ p(\by_\ast \vbar f(\bx_\ast ; \btheta) ) ] )
    =
    -\hspace*{-5pt} \sum_{c \in \{0,1\}} \E[p(\by_\ast=c \vbar f(\bx_\ast ; \btheta) )] \log \E[ p(\by_\ast=c \vbar f(\bx_\ast ; \btheta) )] ,
\end{align}
where $f(\bx_\ast ; \btheta)$ are logits and $p(\by_\ast=c \vbar f(\bx_\ast ; \btheta) )$ is a binary cross-entropy likelihood function.
The total predictive uncertainty is high when either the aleatoric uncertainty is high (e.g., because the input is noisy), or when the epistemic uncertainty is high (e.g., because the model has many possible explanations for the input).
In practice, the total predictive uncertainty $\mathcal{H}(\E [ p(\by_\ast \vbar f(\bx_\ast ; \btheta) ) ])$ is computed with a Monte Carlo estimator $\E[p(\by_\ast \vbar f(\bx_\ast ; \btheta) )] \approx \frac{1}{S}\sum_{i}^S p(\by_\ast \vbar f(\bx_\ast ; \btheta^{(i)}) )$, where parameter realizations $\{ \btheta^{(i)} \}_{i=1}^S$ are sampled from some distribution over the network parameters, and $p(\by_\ast \vbar f(\bx_\ast ; \btheta^{(i)}) )$ denotes the predictive distribution given parameter realization $\btheta^{(i)}$.

\section{Methods}
\label{sec:methods}

Estimating a model's predictive uncertainty in terms of both aleatoric and epistemic uncertainty requires a \emph{distribution over predictive functions}.
Such a distribution over predictive functions can be obtained by treating the parameters of a neural network as random variables and inferring a posterior distribution $p(\btheta \vbar \calD)$---a distribution over the network parameters conditioned on a set of training data $\calD = (\bX_{\calD}, \by_{\calD})$---according to the rules of Bayesian inference.
Neural networks with such distributions over the network parameters---referred to as Bayesian neural networks (\bnn)---induce distributions over functions that are able to capture both aleatoric and epistemic uncertainty~\citep{gal2016uncertainty,mackay1992practical,neal1995bayesian}.
Unfortunately, computing a posterior distribution over the parameters of a neural network according to the rule of Bayesian inference is analytically intractable and requires the use of approximate inference methods~\citep{gal2016uncertainty,graves2011practical,hinton1993keeping,neal1995bayesian,peterson1987mean}.
Below, we describe baseline and state-of-the-art methods for which we implemented standardized and optimized runscripts that are readily extensible for experimentation and deployment in application settings.

\vspace*{-2pt}
\subsection{Maximum A Posteriori Estimation in Bayesian Neural Networks}

As an alternative to inferring a posterior distribution over neural network parameters, maximum a posteriori (\map) estimation yields network parameter values equal to the mode of the exact posterior distribution.
For a prior distribution over network parameters with zero mean and precision $\lambda$, the maximum a posteriori estimate is equal to the solution of the $\ell_2$-regularized optimization problem
\mbox{$
    \argmin_{\btheta} \{ -\log p(\by_{\calD} \vbar f(\bX_{\mathcal{D}} ; \btheta) ) + \lambda ||\btheta||_{2}^2 \},
$}
and as such is equivalent to parameter values obtained by training a neural network with weight decay.
Since \map estimation yields a point estimate of the \map parameters, the \map solution defines a deterministic neural network and is thus unable to capture any epistemic uncertainty.
In classification tasks, they represent aleatoric uncertainty estimates via the predicted class probabilities~\citep{KendallGal2017Uncertainties}.
We use neural networks with \map estimation as a baseline for the benchmark.

\vspace*{-2pt}
\subsection{Variational Inference in Bayesian Neural Networks}\label{subsec:vi_methods}

Variational inference is an approximate inference method that seeks to sidestep the intractability of exact posterior inference over the network parameters by framing posterior inference as a variational optimization problem.
In particular, variational inference in neural networks seeks to find an approximation to the posterior distribution over parameters by solving the optimization problem
\begin{align}
\label{eq:objective_elbo}
    \argmax\nolimits_{q \in \calQ} \{ \mathbb{E}_{q}[\log p(\by_{\calD} \vbar f(\bX_{\calD} ; \btheta) )]
    - \DD_\textrm{KL}(q \,\|\, p ) \} ,
\end{align}
where $\calQ$ is a variational family of distributions and $p$ is a prior distribution.

\textbf{Gaussian Mean-Field Variational Inference.}$~~~$
If $p \defines p_{\bTheta}$ and $q \defines q_{\bTheta}$ are distributions over parameters, $\calQ$ is the family of mean-field (i.e., fully-factorized) Gaussian distributions, and the prior distribution over parameters $p_{\bTheta}$ is also a diagonal Gaussian, the resulting variational objective is amenable to stochastic variational inference and can be optimized using stochastic gradient methods~\citep{blundell2015mfvi,graves2011practical,hinton1993keeping,hoffman2013svi,wainwright2008vi}.
Henceforth, we refer to \bnn inference methods that make these variational assumptions as mean-field variational inference.
To optimize this objective, the expectation is estimated using Monte Carlo sampling and the network parameters are reparameterized as $\bTheta \defines \bmu + \bm{\sigma} \odot \bm{\epsilon}$ with $\bm{\epsilon} \sim \calN(\mathbf{0}, \mathbf{I})$.
Throughout, we use the flipout estimator~\citep{wen2018flipout} to reduce the variance of the gradient estimates, and temper the Kullback-Leibler divergence term in the variational objective~\citep{wenzel2020cold}.

\textbf{Radial-Gaussian Mean-Field Variational Inference.}$~~~$
Radial-Gaussian mean-field variational inference~\citep{farquhar2020radial} uses the same variational objective, prior, and variational distribution as standard Gaussian mean-field variational inference, but uses an alternative gradient estimator to obtain an improved signal-to-noise ratio in the gradient estimates.
Specifically, the network parameters are reparameterized as $\bTheta \defines \bmu + \bm{\sigma} \odot \frac{\bm{\epsilon}}{||\bm{\epsilon}||_{2}} \cdot |r|$ with $\bm{\epsilon} \sim \calN(\mathbf{0}, \mathbf{I})$ and $r \sim \calN(0, 1)$.

\textbf{Function-Space Variational Inference.}$~~~$
\citet{rudner2021fsvi} proposed a tractable function-space variational objective for Bayesian neural networks.
If $p \defines p_{f([\bX_{\calD}, \bX_{\calI}] ; \bTheta)}$ and $q \defines q_{f([\bX_{\calD}, \bX_{\calI}] ; \bTheta)}$ are distributions over functions evaluated at the training inputs $\bX_{\calD}$ and at a set of inducing inputs $\bX_{\calI}$, $\calQ$ is the family of distributions over functions induced by some distribution over network parameters, and the Kullback-Leibler divergence between distributions over functions evaluated at $[\bX_{\calD}, \bX_{\calI}]$ is approximated by a linearization of the neural network mapping, then the resulting variational objective is amenable to stochastic variational inference~\citep{rudner2021fsvi,rudner2021cfsvi}.
In \retina, we define a Gaussian mean-field distribution over the final layer of the neural network and reparameterize the parameters as $\bTheta \defines \bmu + \bm{\sigma} \odot \bm{\epsilon}$ with $\bm{\epsilon} \sim \calN(\mathbf{0}, \mathbf{I})$.

\textbf{Monte Carlo Dropout.}$~~~$
\citet{gal2016dropout} showed that training a deterministic neural network with $\ell_2$- and dropout regularization~\citep{srivastava2014dropout}, that is, solving the optimization problem
\mbox{$
    \argmin_{\btheta} \{ - \E_{q}[ \log p(\by_{\calD} \vbar f(\bX ; \btheta) ) ] + \lambda ||\btheta||_{2}^2 \},
$}
where $q_{\bTheta}$ is the distribution over parameters obtained by applying dropout with a given dropout rate, approximately corresponds to variational inference in a Bayesian neural network.
To sample from the approximate posterior predictive distribution, dropout is applied to the deterministic network parameters.
To optimize the objective above, the expectation is estimated using a single Monte Carlo sample (i.e., by applying dropout).

\textbf{Rank-1 Parameterization.}$~~~$
\citet{dusenberry2020rank1} propose a rank-1 parameterization of Bayesian neural networks, where each weight matrix involves only a distribution on a rank-1 subspace, that is, each stochastic weight matrix is defined as $\mathbf{W}'_k = \mathbf{W}_k \odot \mathbf{r}_k\mathbf{s}_k^\top$, where $ \mathbf{W}_k$ is a deterministic set of weights, and $\mathbf{r}_k$ and $\mathbf{s}_k$ are random vectors of parameters.
Variational distributions over $\mathbf{r}_k$ and $\mathbf{s}_k$ and a Dirac delta distribution over $\mathbf{W}_k$ for all layers $k$ are obtained by optimizing a variational objective.

\vspace*{-2pt}
\subsection{Model Ensembling}
\label{subsec:ensembling}

\textbf{Deep Ensembles.}$~~~$
A deep ensemble~\citep{lakshminarayanan2017simple} is a mixture of multiple independently-trained deterministic neural networks.
As such, unlike \bnns, deep ensembles do not explicitly infer a distribution over the parameters of a single neural network.
Instead, they marginalize over multiple deterministic models to obtain a predictive distribution that captures both aleatoric and epistemic uncertainty.
We construct deep ensembles from multiple \map neural networks trained with different random seeds.

\textbf{Ensembles of Bayesian Neural Networks.}$~~~$
Ensembles of Bayesian neural networks~\citep{filos2019systematic,rudner2021fsvi,smith2018understanding} are mixtures of multiple independently-trained Bayesian neural networks.
They can account for the possibility that any individual approximate posterior distribution obtained via variational inference may be a poor approximation to the exact posterior distribution and may hence yield a poor predictive distribution.
A common issue in the Bayesian deep learning literature is that ensembles are frequently compared to single models, often due to computational constraints.
In \retina, we provide a unified comparison and construct ensembles for all predictive models, including \bnns.

\vspace*{-5pt}
\section{\retina Benchmark}
\label{sec:benchmark}

\vspace*{-5pt}
\subsection{Evaluation Protocol}\label{subsec:eval_protocol}

\textbf{Network Architecture.}$~~~$
We use a ResNet-50 architecture for all experiments~\citep{he2016deep}.
A sigmoid transformation is applied to the final linear layer of all networks to obtain class probabilities corresponding to the outcomes of the binary classification problems described in Sections \ref{subsec:severity_shift} and \ref{subsec:population_shift}.

\textbf{Validation Data, Hyperparameter Tuning, and Monte Carlo Estimation.}$~~~$
Reliable uncertainty estimation on data points from shifted distributions is the central challenge for Bayesian deep learning methods.
In training and evaluating such methods, practitioners must decide how they should choose validation data: specifically, in which settings they would benefit from using ``out-of-distribution'' data points for hyperparameter tuning.
We consider two real-world settings:
\textbf{(i)} No distributionally shifted data is available during hyperparameter tuning. 
This setting reflects scenarios in which practitioners do not know what data or distributional shift they might encounter during deployment and hence cannot make assumptions about it at training time.
\textbf{(ii)} Shifted validation data \emph{is} available for hyperparameter tuning.
This setting reflects scenarios in which practitioners may intend to train a model on data collected from one subpopulation and deploy it on data collected from another subpopulation, but are able to acquire a small number of examples from the deployment subpopulation for use in tuning to improve generalization.
Prior works on out-of-distribution detection~\citep{hendrycks2019deep} and uncertainty quantification~\citep{malinin-pn-2018} have considered setting \textbf{(ii)}, but have not provided a comparative analysis, which would inform practitioners on when they ought to collect shifted validation data for tuning.
We rigorously investigate the two settings across downstream tasks in~\Cref{subsec:more_experiments}.
In the main paper, we report results for models tuned under setting $\textbf{(i)}$. Lastly, for all evaluations, we use five Monte Carlo samples per model to estimate predictive means (e.g., the \mcd \textsc{ensemble} with $K = 3$ ensemble members uses a total of $S=15$ Monte Carlo samples).

\label{sub:metrics}

The aim of the \retina Benchmark is to adequately represent the challenges of real-world distributional shift, and rigorously assess the reliability of (Bayesian) uncertainty quantification in deep learning.
Our selective prediction downstream tasks demonstrate two real-world use cases:\vspace*{-5pt}
\begin{itemize}[leftmargin=*]
    \item \textbf{Tuning Referral Thresholds.}$~~$ 
    On the \textit{Severity Shift} task, models demonstrate reasonable uncertainty estimates:
    predictive performance increases monotonically with an increasing referral rate $\tau$. 
    Therefore, practitioners can infer which referral rate will lead to a desired predictive performance, or infer the performance for a predetermined referral rate (i.e., respecting a budget of expert time).
    \item \textbf{Detecting Low-Quality Predictive Uncertainty.}$~~~$
    On the \textit{Country Shift} task, most methods fail:
    predictive performance on the shifted dataset \emph{declines} as $\tau$ increases, indicating that the quality of uncertainty estimates is no better than random referral.
    Importantly, this failure is \emph{not reflected} in the standard performance measure for retinopathy diagnosis, the receiver operating characteristic (ROC) curve~\citep{leibig2017leveraging}---the area under the ROC curve (AUC) is \emph{higher} on the shifted evaluation dataset than the in-domain dataset---meaning that a practitioner using only AUC might wrongly conclude that these models would perform well as part of an automated diagnosis pipeline (cf. \Cref{fig:diagnosis}) on distributionally shifted data.
\end{itemize}
\vspace{-2pt}

For each method, we assess both AUC and accuracy as a function of the referral rate $\tau$, evaluating the models' predictions for the $(1 - \tau)$ proportion of cases on which they are most certain, as indicated by their predictive uncertainty estimates.
We additionally examine predictive uncertainty histograms for each task, method, and ground-truth clinical label (cf.~\Cref{subsec:pred_uncert_hist}, \Cref{sec:app_single_model_pred_uncert_hist}) to determine if methods have particularly good or bad uncertainty estimates at particular severity levels.
We also investigate other metrics to assess the reliability of models' uncertainty estimates, including \emph{expected calibration error} and \emph{out-of-distribution detection AUC}, in \Cref{subsec:more_experiments}.

\begin{figure}[t!]
\centering
\begin{subfigure}{\linewidth}
    \includegraphics[width=\linewidth]{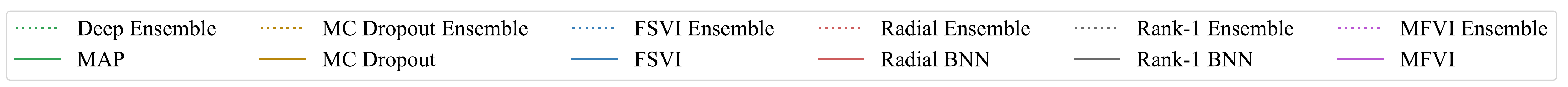}
\end{subfigure}
\begin{subfigure}[l]{0.24\linewidth}
    \hspace*{-10pt}
    \includegraphics[width=\linewidth]{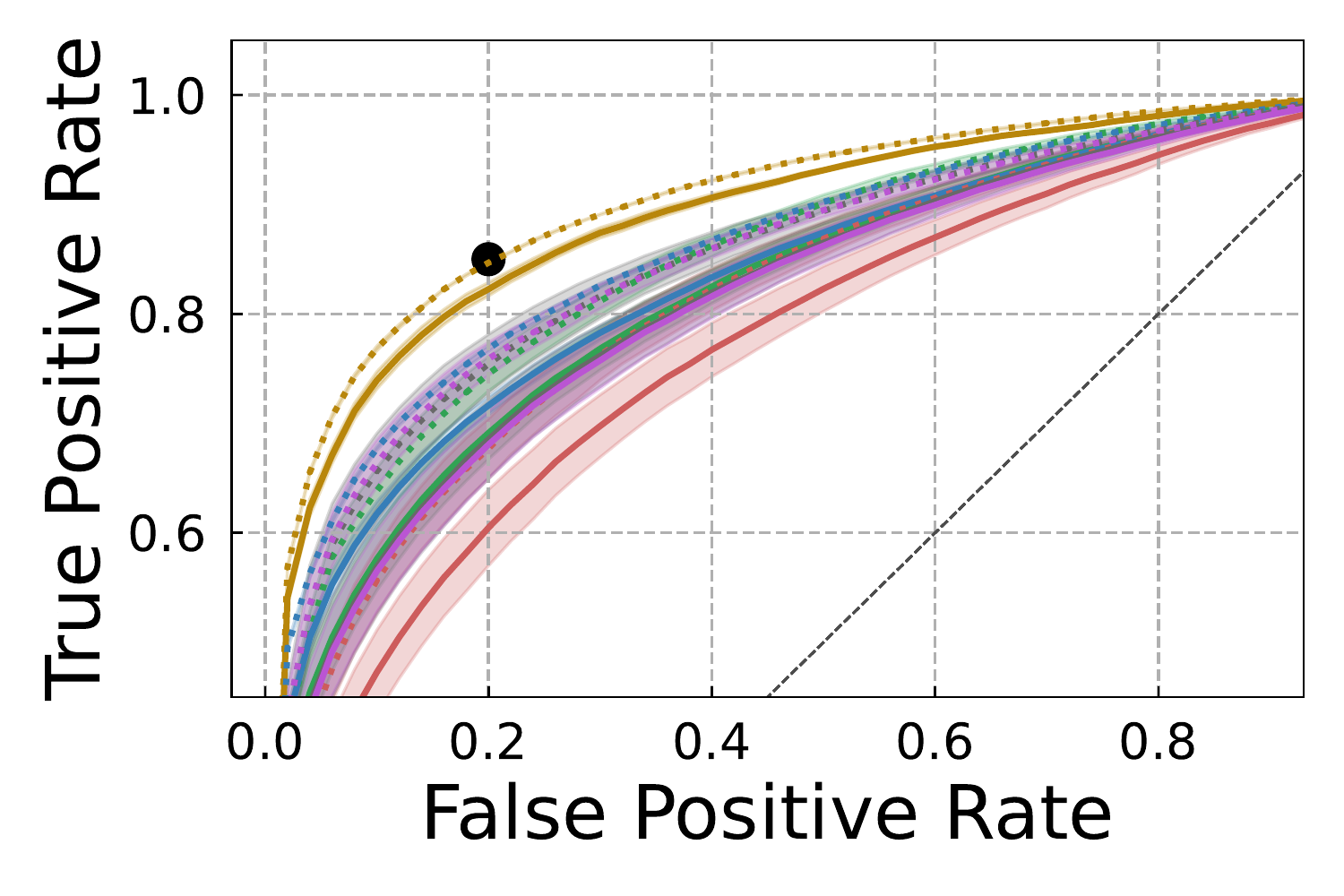}
    \vspace*{-7pt}
    \caption{
        \textbf{ROC: In-Domain\\$~$}
    }
\end{subfigure}
\begin{subfigure}[l]{0.24\linewidth}
    \hspace*{-10pt}
    \includegraphics[width=\linewidth]{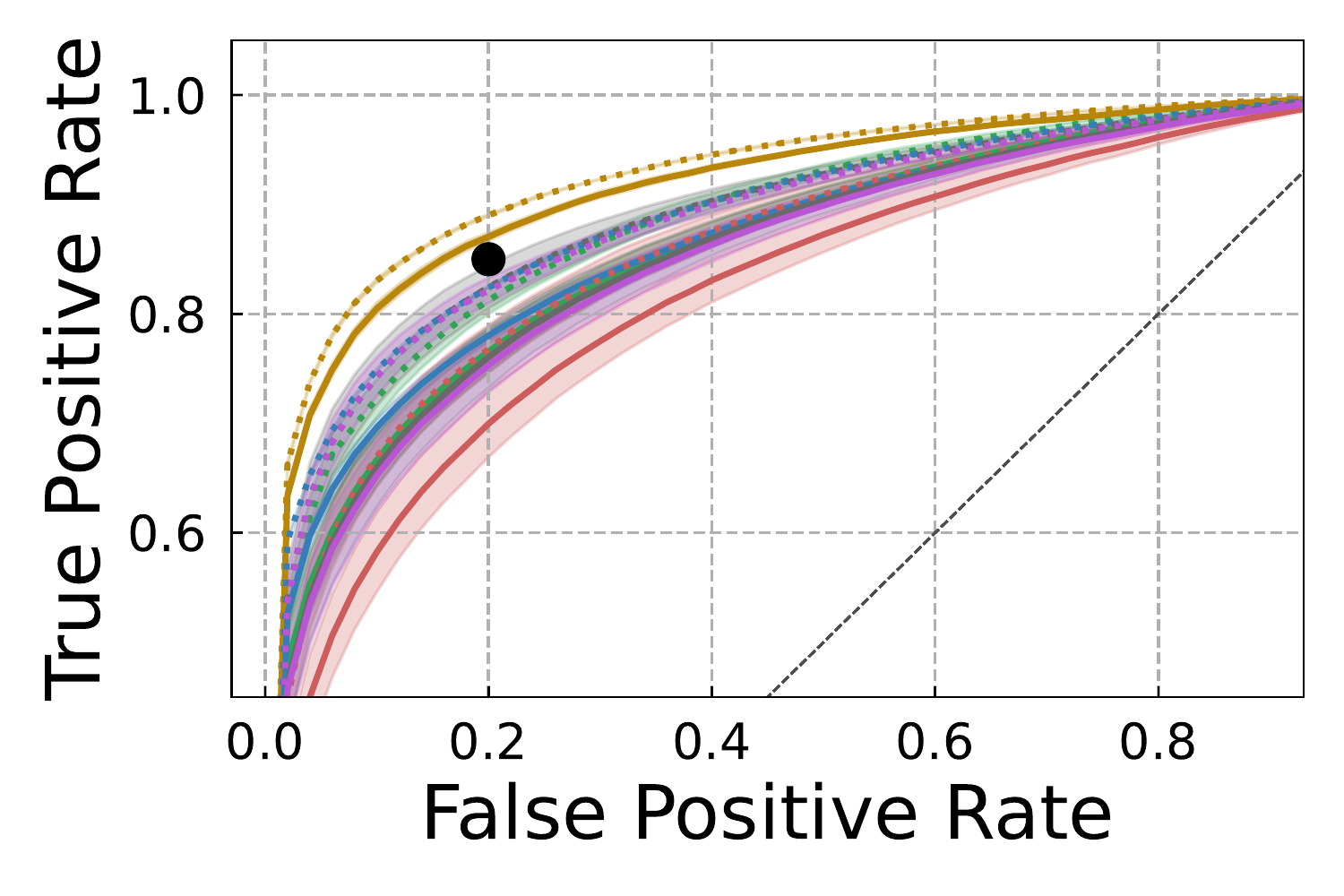}
    \vspace*{-7pt}
    \caption{
        \textbf{ROC: Joint\\$~$}
    }
\end{subfigure}
\begin{subfigure}[l]{0.24\linewidth}
    \hspace*{-10pt}
    \includegraphics[width=\linewidth]{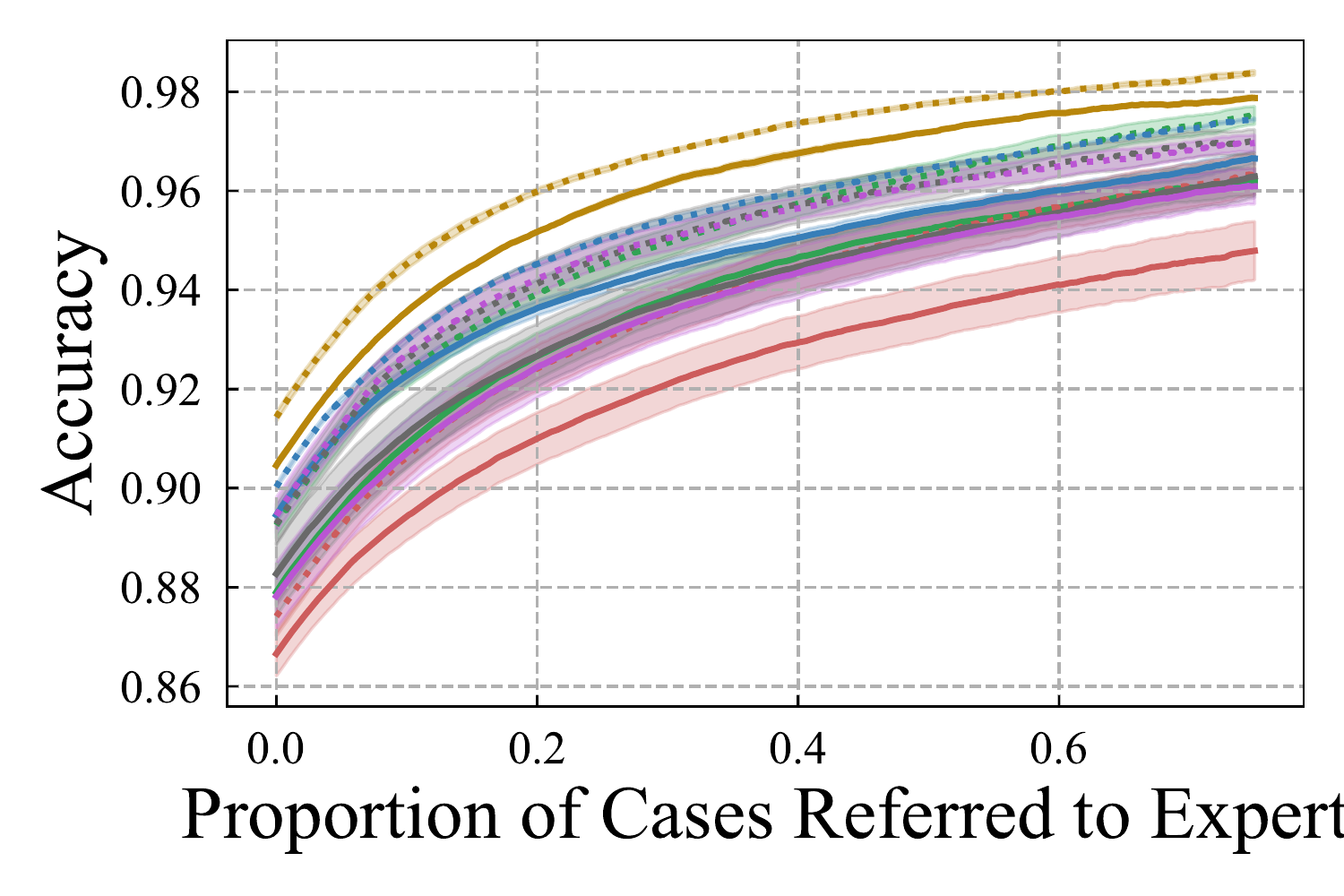}
    \vspace*{-7pt}
    \caption{
        \centering\textbf{Selective Prediction Accuracy: In-Domain}
    }
\end{subfigure}
\begin{subfigure}[r]{0.24\linewidth}
    \hspace*{-10pt}
    \includegraphics[width=\linewidth]{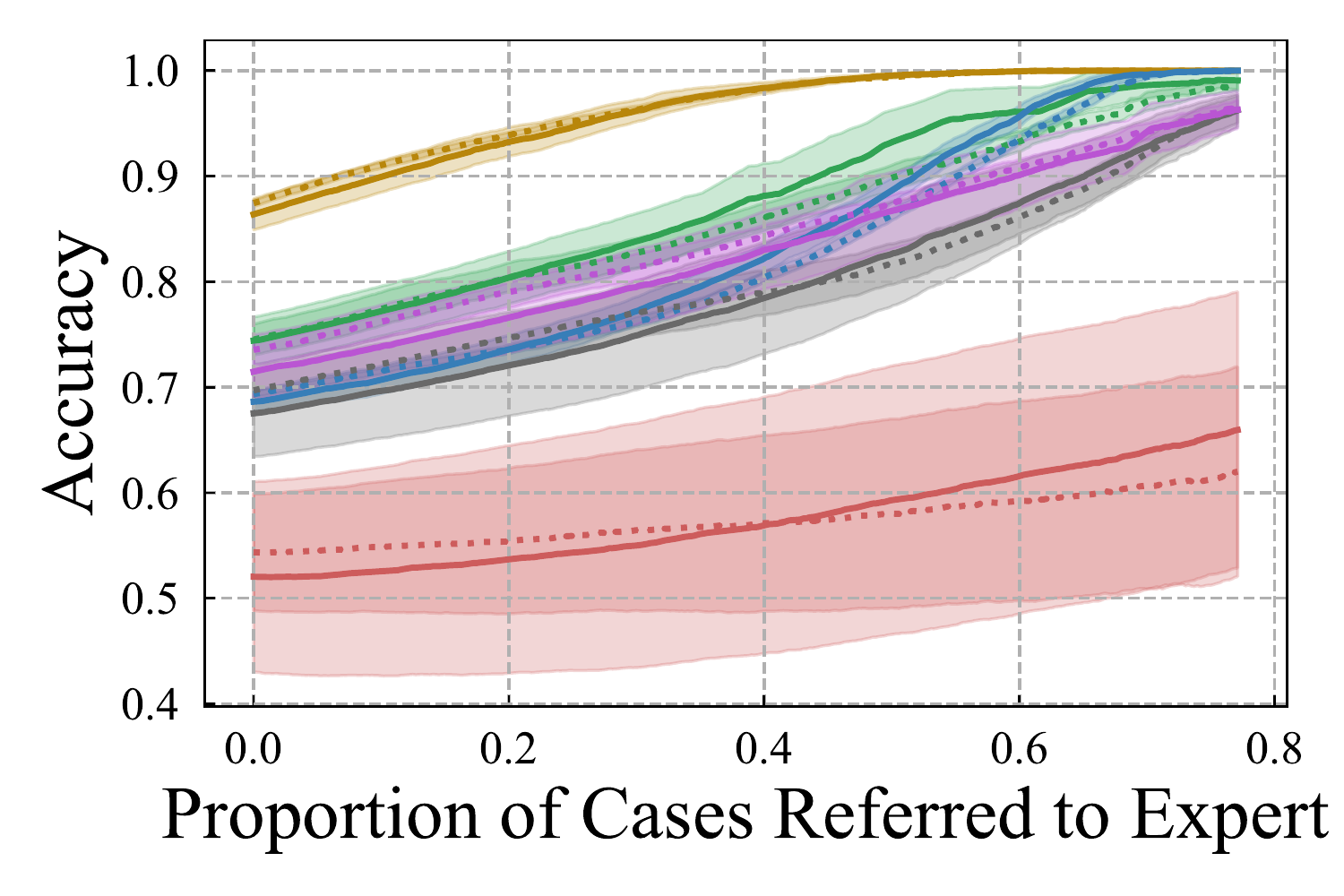}
    \vspace*{-7pt}
    \caption{
        \centering\textbf{Selective Prediction Accuracy: Severity Shift}
    }
\end{subfigure}
\vspace*{2pt}
\vspace*{-5pt}  %
\caption{
    \textbf{Severity Shift.} 
    We jointly assess model predictive performance and uncertainty quantification on the in-domain test dataset composed only of cases with either no, mild, or moderate diabetic retinopathy, and the \textit{Severity Shift} evaluation set composed only of severe and proliferative cases.
    \textbf{Left:} The \textit{receiver operating characteristic curve} (ROC) for (\textbf{a}) in-domain diagnosis and for (\textbf{b}) a joint dataset composed of examples from both the in-domain and \textit{Severity Shift} evaluation sets.
    The dot in 
    \textbf{black}
    denotes the NHS-recommended 85\% sensitivity and 80\% specificity ratios~\citep{widdowson2016management}.
    \textbf{Right:} Selective prediction on accuracy in the (\textbf{c}) in-domain and (\textbf{d})  \textit{Severity Shift} settings.
    Shading denotes standard error computed over six random seeds.
    See~\Cref{subsubsec:severity_shift}.
}
\label{fig:severity_shift}
\vspace*{-5pt}
\end{figure}

\begin{figure}[t!]
\centering
\begin{subfigure}[l]{0.24\linewidth}
    \hspace*{-10pt}
    \includegraphics[width=\linewidth]{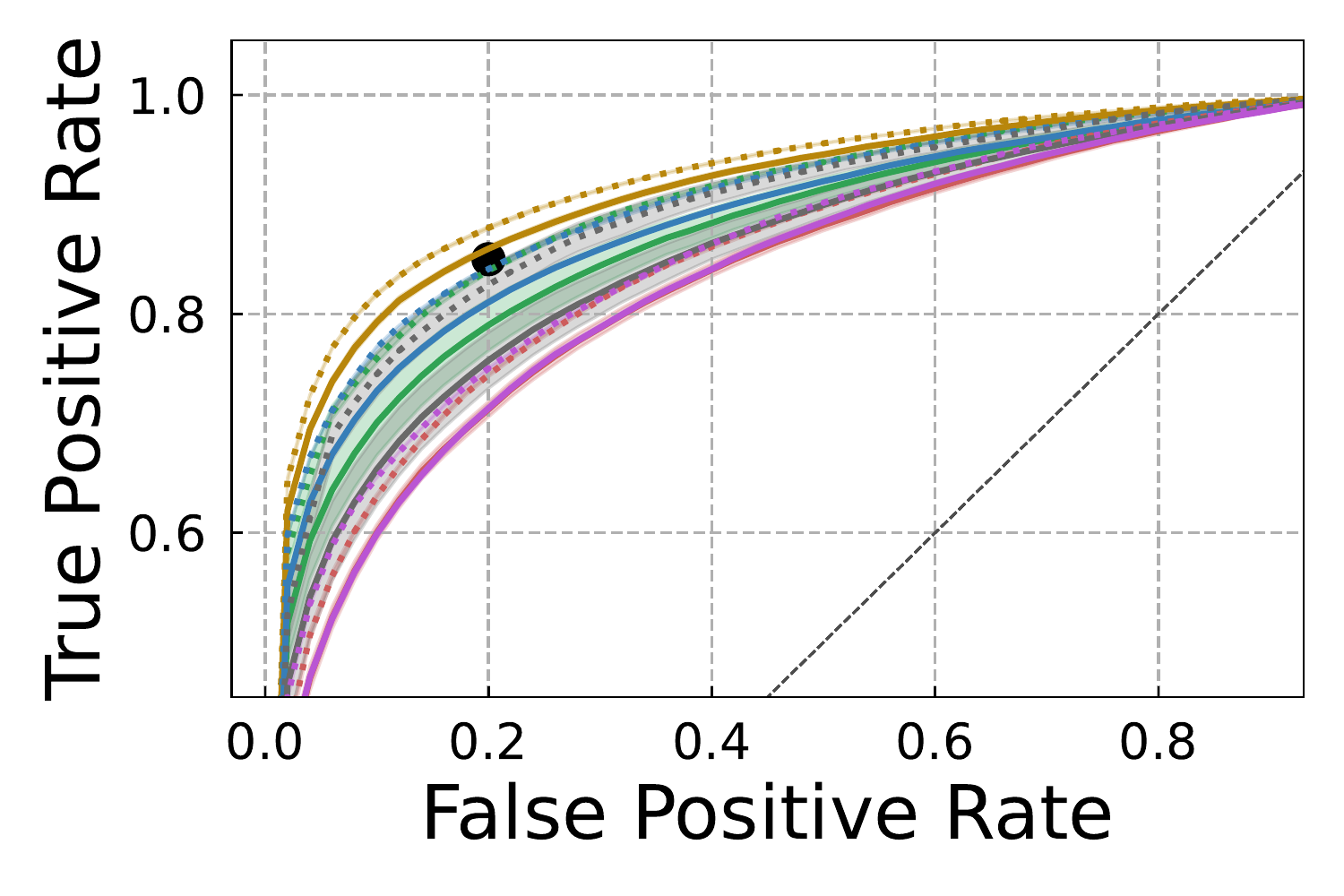}
    \vspace*{-7pt}
    \caption{
        \textbf{ROC: In-Domain\\$~$}
    }
\end{subfigure}
\begin{subfigure}[l]{0.24\linewidth}
    \hspace*{-10pt}
    \includegraphics[width=\linewidth]{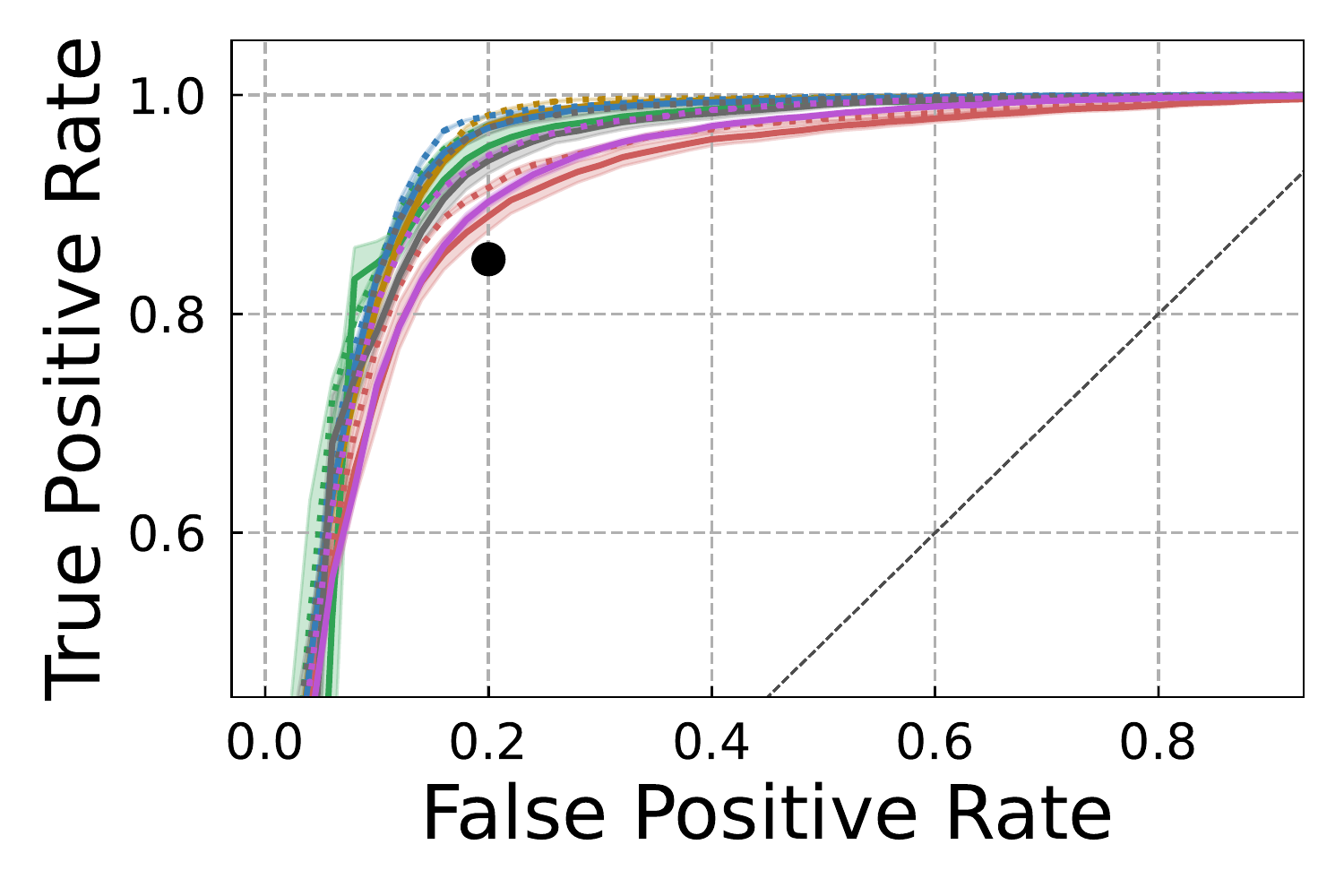}
    \vspace*{-7pt}
    \caption{
        \textbf{ROC: Country Shift\\$~$}
    }
\end{subfigure}
\begin{subfigure}[l]{0.24\linewidth}
    \hspace*{-10pt}
    \includegraphics[width=\linewidth]{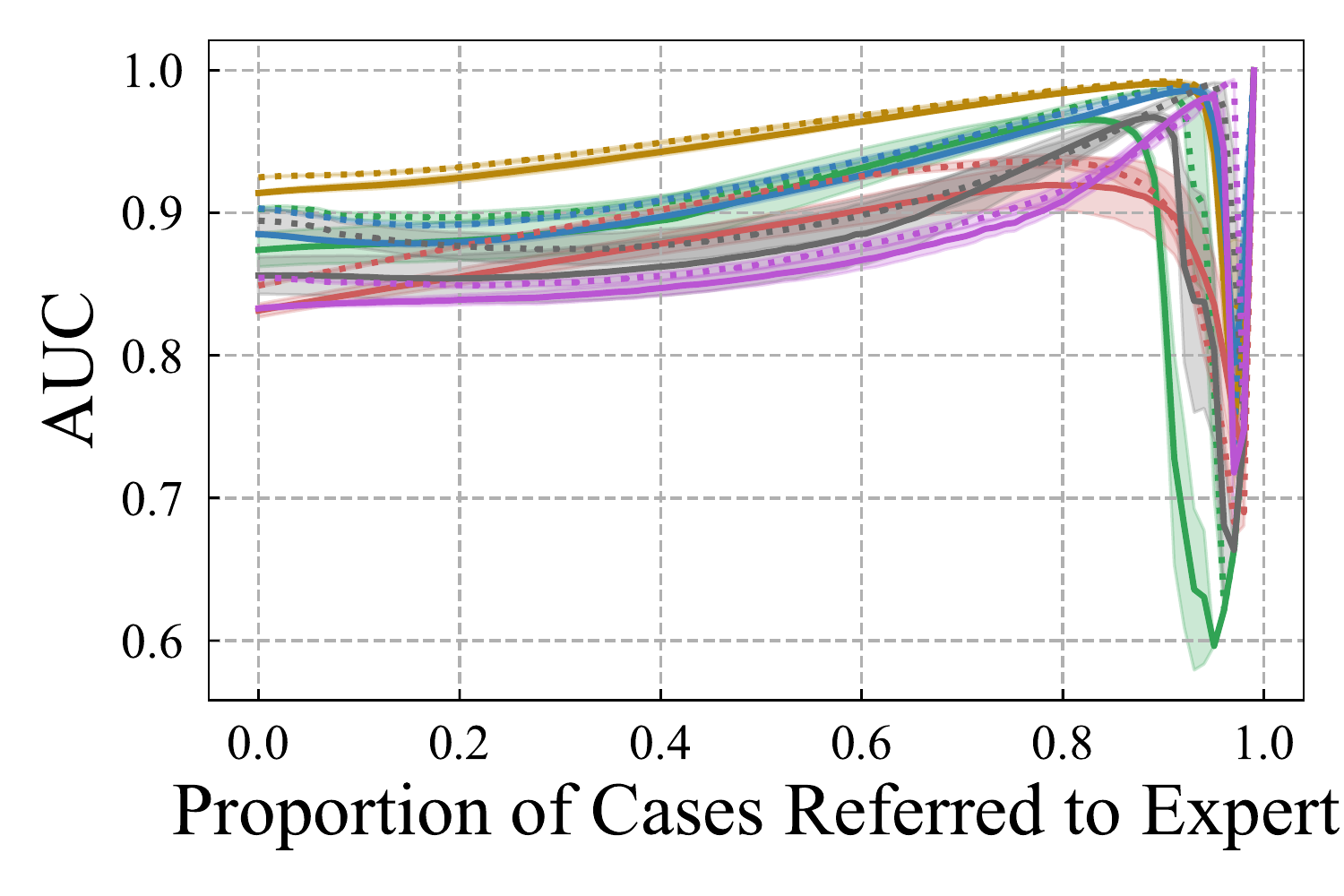}
    \vspace*{-7pt}
    \caption{
        \centering\textbf{Selective Prediction AUC: In-Domain}
    }
\end{subfigure}
\begin{subfigure}[r]{0.24\linewidth}
    \hspace*{-10pt}
    \includegraphics[width=\linewidth]{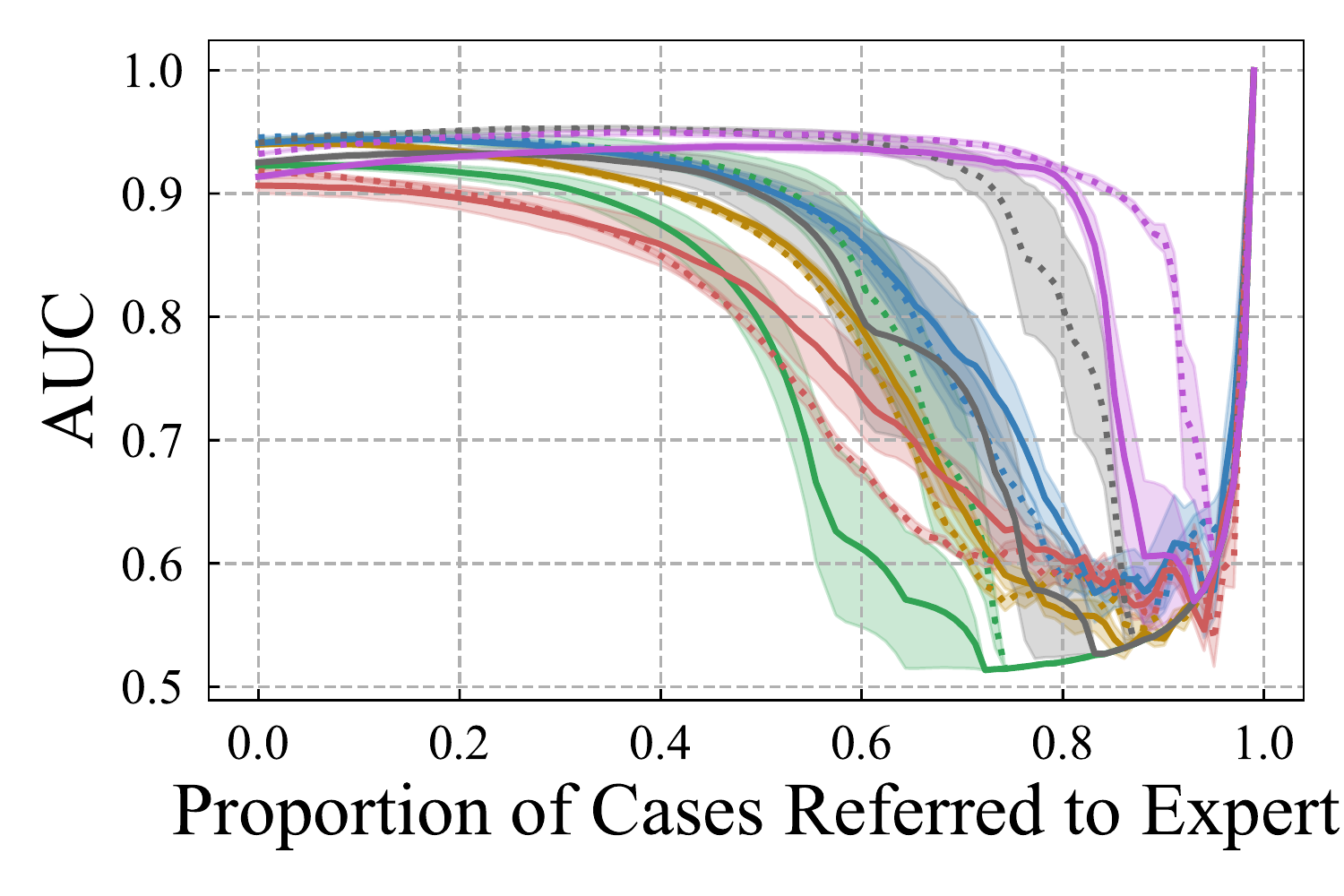}
    \vspace*{-7pt}
    \caption{
        \centering\textbf{Selective Prediction AUC: Country Shift}
    }
\end{subfigure}
\vspace*{-5pt}  %
\caption{
    \textbf{Country Shift.} 
    We jointly assess model predictive performance and uncertainty quantification on both in-domain and distributionally shifted data.
    \textbf{Left:} The \textit{receiver operating characteristic curve} (ROC) for in-population diagnosis on the (\textbf{a}) \citet{kaggle_2015} test set and for (\textbf{b}) changing medical equipment and patient populations on the \citet{APTOS_2019} test set. 
    The dot in 
    \textbf{black}
    denotes the NHS-recommended 85\% sensitivity and 80\% specificity ratios \cite{widdowson2016management}.
    \textbf{Right:} \emph{selective prediction} on AUC in the (\textbf{c}) \citet{kaggle_2015} and the (\textbf{d}) \citet{APTOS_2019} settings.
    Shading denotes standard error computed over six random seeds.
    See \Cref{subsubsec:country_shift}.
}
\label{fig:country_shift}
\vspace*{-5pt}
\end{figure}

\vspace*{-2pt}
\subsection{Severity Shift}
\label{subsubsec:severity_shift}

On the \textit{Severity Shift} task (\Cref{fig:severity_shift}, \Cref{tab:metrics_severity}), models are trained on EyePACS images that show signs of at most moderate diabetic retinopathy.
We assess their ability to generalize to images showing signs of severe or proliferative retinopathy.
Surprisingly, we find that models generalize well from cases with no worse than moderate diabetic retinopathy (in-domain) (\Cref{fig:severity_shift}(\textbf{a})) to severe cases (\Cref{fig:severity_shift}(\textbf{b})), improving their AUC under the distribution shift.

\textbf{Methods Generalize Reasonably Well Under Severity Shift.}$~~~$
Reliable predictive uncertainty estimates correlate with predictive error, and therefore we would expect a model's performance (e.g., measured in terms of accuracy or AUC) to increase as more examples on which the model exhibits high uncertainty are referred to an expert.
On both the in-domain and \textit{Severity Shift} evaluation sets (Figures~\ref{fig:severity_shift}(\textbf{c}) and (\textbf{d})), models demonstrate reasonable uncertainty in that accuracy monotonically increases as $\tau$ increases.
This highlights two ways that practitioners may use selective prediction to prepare models for a real-world deployment in the presence of potential distribution shifts.
First, given a performance target (e.g., $\geq 95\%$ accuracy) the referral curve can be used to determine the minimum $\tau$ achieving this target, estimating a medical experts' workload.
Second, for a maximum acceptable referral rate (e.g., a clinic has medical experts to handle referral of $\tau \leq 20\%$ of patients) the referral curve can be used to determine the optimal $\tau$ value and the corresponding performance.
For monotonically increasing referral curves, the optimal $\tau$ is uniquely the maximum acceptable referral rate.

\textbf{Taking into Account Epistemic Uncertainty Can Improve Reliability.}$~~~$
On the \textit{Severity Shift} task (\Cref{fig:severity_shift}(\textbf{d})) many models achieve near-perfect accuracy well before all examples have been referred.
For example, \mcd, which incorporates both epistemic and aleatoric uncertainty (cf. \Cref{sec:uncertainty_estimation}), achieves 100\% predictive accuracy near the 50\% referral rate---nearly 20\% lower than the referral rate at which a deterministic neural network (\map), which only represents aleatoric uncertainty, achieves this level of accuracy.
Other variational inference methods underperform \map, underscoring the importance of continued work on approximate inference in \bnns.

\textbf{Predictive Uncertainty Histograms Identify Harmful Uncertainty Quantification.}$~~~$
In~\Cref{fig:label_bin_sev_shifted_single} (\Cref{sec:app_single_model_pred_uncert_hist}), we find that \map, \textsc{rank}-1, and \mfvi generate worse uncertainty estimates than other methods on the shifted data (labels 3 and 4);
many of their incorrect predictions are assigned low predictive uncertainty (i.e., the red distribution is concentrated near 0).
These include false negatives with low uncertainty which are particularly dangerous in automated diagnosis settings (cf. \Cref{fig:diagnosis}), as a medical expert would not be able to catch the model's failure to recognize the condition.

\begin{table*}[htb!]
\vspace{-2mm}
\centering
\caption{
    \textbf{Country Shift.}
    Prediction and uncertainty quality of baseline methods in terms of the area under the receiver operating characteristic curve (AUC) and classification accuracy, as a function of the proportion of data referred to a medical expert.
    All methods are tuned on in-domain validation AUC, and ensembles have $K = 3$ constituent models (true for all subsequent tables unless specified otherwise).
    On in-domain data, \mcd performs best across all thresholds. 
    On distributionally shifted data, no method consistently performs best.
}
\vspace*{-5pt}
\resizebox{1.0\linewidth}{!}{%
\begin{tabular}{@{\extracolsep{2pt}}lcccccc@{}}
\midrule
\midrule
& \multicolumn{2}{c}{No Referral} & \multicolumn{2}{c}{$50\%$ Data Referred} & \multicolumn{2}{c}{$70\%$ Data Referred} \\
\cline{2-3}
\cline{4-5}
\cline{6-7}\\
\textbf{Method}         &
\textbf{AUC (\%) $\uparrow$}            &
\textbf{Accuracy (\%) $\uparrow$}       &
\textbf{AUC (\%) $\uparrow$}            &
\textbf{Accuracy (\%) $\uparrow$}       &
\textbf{AUC (\%) $\uparrow$}            &
\textbf{Accuracy $\uparrow$}       \\
\midrule
\multicolumn{7}{c}{EyePACS Dataset (In-Domain)}\\
\midrule
\map (Deterministic)	& $87.4\pms{1.3}$ & $88.6\pms{0.7}$ & $91.1\pms{1.8}$ & $95.9\pms{0.4}$ & $94.9\pms{1.1}$ & $96.5\pms{0.3}$ \\
\mfvi	& $83.3\pms{0.2}$ & $85.7\pms{0.1}$ & $85.5\pms{0.7}$ & $94.5\pms{0.1}$ & $88.2\pms{0.7}$ & $95.9\pms{0.1}$ \\
\textsc{radial}-\mfvi	& $83.2\pms{0.5}$ & $74.2\pms{5.0}$ & $88.9\pms{0.9}$ & $81.8\pms{6.0}$ & $91.2\pms{1.3}$ & $83.8\pms{5.5}$ \\
\fsvi	& $88.5\pms{0.1}$ & $89.8\pms{0.0}$ & $91.0\pms{0.4}$ & $96.4\pms{0.0}$ & $94.3\pms{0.3}$ & $97.2\pms{0.1}$ \\
\mcd	& $91.4\pms{0.2}$ & $90.9\pms{0.1}$ & $95.3\pms{0.2}$ & $97.4\pms{0.1}$ & $97.4\pms{0.1}$ & $98.1\pms{0.0}$ \\
\textsc{rank}-1	& $85.6\pms{1.4}$ & $87.7\pms{0.8}$ & $87.1\pms{2.3}$ & $95.3\pms{0.5}$ & $90.9\pms{2.0}$ & $96.4\pms{0.4}$ \\
\textsc{deep ensemble}	& $90.3\pms{0.2}$ & $90.3\pms{0.3}$ & $91.7\pms{0.6}$ & $97.2\pms{0.0}$ & $95.0\pms{0.5}$ & $97.9\pms{0.0}$ \\
\mfvi \textsc{ensemble}	& $85.4\pms{0.0}$ & $87.8\pms{0.0}$ & $86.3\pms{0.4}$ & $95.4\pms{0.0}$ & $89.2\pms{0.4}$ & $96.7\pms{0.1}$ \\
\textsc{radial}-\mfvi \textsc{ensemble}	& $84.9\pms{0.1}$ & $74.2\pms{1.5}$ & $91.4\pms{0.2}$ & $83.4\pms{1.7}$ & $93.3\pms{0.3}$ & $85.9\pms{1.6}$ \\
\fsvi \textsc{ensemble}	& $90.3\pms{0.1}$ & $90.6\pms{0.0}$ & $92.1\pms{0.2}$ & $97.1\pms{0.0}$ & $95.2\pms{0.2}$ & $97.8\pms{0.1}$ \\
\mcd \textsc{ensemble}	& $\mathbf{92.5\pms{0.0}}$ & $\mathbf{91.6\pms{0.0}}$ & $\mathbf{95.8\pms{0.1}}$ & $\mathbf{97.8\pms{0.0}}$ & $\mathbf{97.7\pms{0.1}}$ & $\mathbf{98.4\pms{0.0}}$ \\
\textsc{rank}-1 \textsc{ensemble}	& $89.5\pms{0.8}$ & $89.3\pms{0.4}$ & $88.5\pms{1.3}$ & $96.9\pms{0.3}$ & $91.6\pms{1.2}$ & $97.6\pms{0.3}$ \\

\midrule
\multicolumn{7}{c}{APTOS 2019 Dataset (Population Shift)}\\
\midrule

\map (Deterministic)	& $92.2\pms{0.2}$ & $86.2\pms{0.6}$ & $80.1\pms{3.6}$ & $87.6\pms{1.5}$ & $55.4\pms{4.3}$ & $85.4\pms{1.2}$ \\
\mfvi	& $91.4\pms{0.2}$ & $84.1\pms{0.3}$ & $93.8\pms{0.4}$ & $92.1\pms{0.5}$ & $93.0\pms{0.6}$ & $92.7\pms{0.5}$ \\
\textsc{radial}-\mfvi	& $90.7\pms{0.7}$ & $71.8\pms{4.6}$ & $82.0\pms{2.5}$ & $81.5\pms{2.7}$ & $66.4\pms{2.1}$ & $85.9\pms{1.0}$ \\
\fsvi	& $94.1\pms{0.1}$ & $87.6\pms{0.5}$ & $90.6\pms{0.9}$ & $90.7\pms{0.7}$ & $77.2\pms{4.6}$ & $89.8\pms{0.3}$ \\
\mcd	& $94.0\pms{0.2}$ & $86.8\pms{0.2}$ & $87.4\pms{0.3}$ & $88.1\pms{0.2}$ & $65.3\pms{1.7}$ & $88.2\pms{0.4}$ \\
\textsc{rank}-1	& $92.5\pms{0.3}$ & $86.2\pms{0.5}$ & $90.1\pms{2.5}$ & $91.4\pms{1.1}$ & $75.1\pms{7.8}$ & $89.5\pms{1.5}$ \\
\textsc{deep ensemble}	& $94.2\pms{0.2}$ & $87.5\pms{0.1}$ & $91.2\pms{1.9}$ & $92.4\pms{0.9}$ & $67.4\pms{7.3}$ & $90.1\pms{1.2}$ \\
\mfvi \textsc{ensemble}	& $93.2\pms{0.1}$ & $87.0\pms{0.2}$ & $\mathbf{94.9\pms{0.3}}$ & $\mathbf{93.7\pms{0.3}}$ & $\mathbf{94.2\pms{0.3}}$ & $\mathbf{94.0\pms{0.3}}$ \\
\textsc{radial}-\mfvi \textsc{ensemble}	& $91.8\pms{0.2}$ & $69.0\pms{1.9}$ & $78.6\pms{0.6}$ & $79.8\pms{0.9}$ & $60.9\pms{0.3}$ & $86.7\pms{0.2}$ \\
\fsvi \textsc{ensemble}	& $\mathbf{94.6\pms{0.1}}$ & $\mathbf{88.9\pms{0.2}}$ & $90.7\pms{0.5}$ & $91.1\pms{0.6}$ & $74.1\pms{3.4}$ & $89.8\pms{0.2}$ \\
\mcd \textsc{ensemble}	& $94.1\pms{0.1}$ & $87.6\pms{0.1}$ & $86.8\pms{0.2}$ & $88.0\pms{0.2}$ & $62.3\pms{0.4}$ & $87.7\pms{0.2}$ \\
\textsc{rank}-1 \textsc{ensemble}	& $94.1\pms{0.2}$ & $88.3\pms{0.2}$ & $\mathbf{94.9\pms{0.4}}$ & $93.5\pms{0.3}$ & $92.4\pms{1.5}$ & $93.8\pms{0.3}$ \\

\midrule
\midrule
\end{tabular}
}
\label{tab:metrics_country}
\vspace*{-15pt}
\end{table*}

\vspace*{-2pt}
\subsection{Country Shift}
\label{subsubsec:country_shift}

In the \textit{Country Shift} task (\Cref{fig:country_shift}, \Cref{tab:metrics_country}), we consider the performance of models trained on the US EyePACS~\citep{kaggle_2015} dataset and evaluated under distributional shift, on the Indian APTOS dataset~\citep{APTOS_2019}. 
The left two plots of~\Cref{fig:country_shift} present the ROC curves of methods evaluated on the in-domain (\textbf{a}) and \textit{Country Shift} (\textbf{b}) evaluation datasets.
The black dot in Figures~\ref{fig:country_shift}(\textbf{a}) and (\textbf{b}) denotes the minimum sensitivity--specificity threshold 
for the deployment of automated diabetic retinopathy diagnosis systems set by the British National Health Service (NHS)~\citep{widdowson2016management}.
On the in-domain test dataset, only the \mcd variants meet the NHS standard; on the APTOS dataset, essentially all methods surpass the standard.\footnote{We investigate this in~\Cref{subsec:app_class_balance_aptos} and find that class proportions do not account for the improved predictive performance on APTOS, implying other contributing factors such as demographics or camera type.}
Hence, practitioners using only the ROC curve and its AUC (cf.~\Cref{tab:metrics_country}) might conclude that their model generalizes 
under the distribution shift although the ROC curve provides no information on the application of uncertainty estimates to real-world scenarios (cf.~\Cref{fig:diagnosis}).

\textbf{Selective Prediction Can Indicate Failures in Uncertainty Estimation.}$~~~$
Unlike the ROC curve, the selective prediction metric conveys how a model would perform in an automated diagnosis pipeline in which the reliability of models' uncertainty estimates directly impacts performance (cf.~\Cref{fig:diagnosis}). 
Recall that if a model generates reliable predictive uncertainty estimates, the AUC should increase as more patients with uncertain predictions are referred for expert review.
This mechanism is illustrated well by the application of \mfvi to the \textit{Country Shift} task (\Cref{fig:country_shift}\textbf{(d)} and \Cref{tab:metrics_country}), since the AUC improves from an initial $91.4\%$ up to $93.8\%$ when referring 50\% of the patients, but then deteriorates as the model is forced to refer patients on which it is both certain and correct.
In contrast, other models' AUCs trend downwards; using uncertainty to refer patients actively hurts model performance on this shifted dataset.

\textbf{Different Prediction Tasks Yield Different Method Rankings.}$~~~$
In~\Cref{fig:country_shift}(\textbf{c}), variational inference methods, including \mcd, \fsvi, and \textsc{deep ensemble}, outperform \map inference.
This highlights that rankings are task-dependent, and underscores the importance of generic evaluation frameworks to enable rapid benchmarking on many tasks.

\vspace*{-5pt}
\section{Conclusions}
\label{sec:discussion_conclusion}
\vspace*{-5pt}

The deployment of modern machine learning models in safety-critical real-world settings necessitates trust in the reliability of the models' predictions.

To encourage the development of Bayesian deep learning methods that are capable of generating reliable uncertainty estimates about their predictions, we introduced the \retina Benchmark, a set of safety-critical real-world clinical prediction tasks which highlight various shortcomings of existing uncertainty quantification methods.
We demonstrate that by taking into account the quality of predictive uncertainty estimates, selective prediction can help identify whether methods might fail when deployed as part of an automated diagnosis pipeline (cf.~\Cref{fig:diagnosis}), whereas standard metrics such as ROC curves cannot.

While no single set of benchmarking tasks is a panacea, we hope that the tasks and evaluation methods presented in \retina will significantly lower the barrier for assessing the reliability of Bayesian deep learning methods on safety-critical real-world prediction tasks.
\vspace*{-10pt}

\clearpage

\begin{ack}
We thank Google Research for providing computational and storage resources.
We thank Intel Labs for their computational support.
We thank Ranganath Krishnan for his contributions to the \textsc{rank}-1 implementation by porting a CIFAR-10 training script, and for sharing his expertise on variational inference in Bayesian neural networks.
We thank Sebastian Farquhar for his contributions to the \textsc{radial}-\mfvi implementation, including a TensorFlow Probability distribution, code review, and feedback on tied means, $\ell_2$-regularization, variance reduction, and hyperparameters.
We thank Jorge Cuadros, OD, PhD (CEO of EyePACS) and Jan Brauner, MD (University of Oxford) for lending their domain expertise in task design.
We thank all other contributors to the \textit{Uncertainty Baselines} project~\citep{nado2021uncertainty} (into which this benchmark is integrated): Mark Collier, Josip Djolonga, Marton Havasi, Rodolphe Jenatton, Jeremiah Liu, Zelda Mariet, Jeremy Nixon, Shreyas Padhy, Jie Ren, Faris Sbahi, Yeming Wen, Florian Wenzel, Kevin Murphy, D. Sculley, Balaji Lakshminarayanan, and Jasper Snoek.
NB and TGJR acknowledge funding from the Rhodes Trust.
TGJR also acknowledges funding from Qualcomm and the Engineering and Physical Sciences Research Council (EPSRC).
\end{ack}

\bibliographystyle{plainnat}
\bibliography{references}

\clearpage

\begin{appendices}

\crefalias{section}{appsec}
\crefalias{subsection}{appsec}
\crefalias{subsubsection}{appsec}

\setcounter{equation}{0}
\renewcommand{\theequation}{\thesection.\arabic{equation}}

\onecolumn

\section*{\LARGE Supplementary Material}
\label{sec:appendix}

\section*{Table of Contents}
\vspace*{-10pt}
\startcontents[sections]
\printcontents[sections]{l}{1}{\setcounter{tocdepth}{2}}

\section{Implementation, Training, and Evaluation Details}

\subsection{\textsc{Retina} Benchmark Software Design Principles}
\label{subsec:swe_design}

Reproducibility in machine learning is often hampered by the wide variety of experimental artifacts made available in papers.
Perhaps the most common approach is a GitHub dump of experimental code lacking documentation and testing.
This common practice fails to enforce a rigorous standard across works:
for example, experiment protocol on cross-validation, access to distributionally shifted validation data, and various tweaks in optimization such as learning rate annealing.

The \retina Benchmark is implemented in the open-sourced \textit{Uncertainty Baselines}~\citep{nado2021uncertainty} repository. 
All models implemented in this repository conform to explicit design principles intended to facilitate easy extension and reproduction of dataset loading utilities, metrics, and evaluation.

\paragraph{Extensibility.}
Each model baseline (e.g., \map, \mcd, \fsvi) is implemented in its own self-contained experiment pipeline.
This minimizes external dependencies, and therefore provides researchers and practitioners an immediate starting point for experimenting with a particular model.
For example, \citet{tran2022plex} extended the RETINA codebase to include experiment pipelines for state-of-the-art Vision Transformers \citep{dosovitskiy2021an} pretrained on the ImageNet-21K \citep{imagenet} dataset.
Datasets are implemented as lightweight wrappers around TensorFlow Datasets~\citep{TFDS}.
Users that wish to extend our benchmark with new datasets (e.g., clinical practitioners that wish to apply our methods on their own diabetic retinopathy tasks) can follow our custom implementation of the APTOS \cite{APTOS_2019} data loader, which constructs the dataset from raw images and a CSV containing metadata, and applies the preprocessing used by the winner of the EyePACS Kaggle competition \cite{kaggle_2015}. Dataset implementation can be found here.\footnote{\href{https://github.com/google/uncertainty-baselines/tree/main/uncertainty_baselines/datasets}{\scriptsize\texttt{https://github.com/google/uncertainty-baselines/tree/main/uncertainty\_baselines/datasets}}}

\paragraph{Framework Agnosticity.}
\retina is framework-agnostic. 
For example, \fsvi is implemented in JAX, a variant of MC Dropout is in PyTorch~\citep{NEURIPS2019_9015} (though we use in this work a TensorFlow variant to simplify TPU tuning), and other models in raw TensorFlow~\citep{tensorflow2015-whitepaper}.
This interoperability means that users can easily incorporate our datasets and evaluation utilities, including an arrangement of robustness and uncertainty metrics such as \textit{selective prediction}, \textit{out-of-distribution detection}, and \textit{expected calibration error}.

\paragraph{Reproducibility.}
All models include testing, and all results are reported over multiple seeds. 
For each method (e.g., \mcd or \mfvi), downstream task (\textit{Country} and \textit{Severity Shift}), and tuning assumption (whether or not distributionally shifted validation data is available for tuning), we sweep over at least 32 hyperparameter configurations.
Instead of using a domain-specific and limiting tuning framework for this, we simply provide hyperparameters through Python flags, and implement for convenience of the user the ability to specify automatic logging to TensorBoard and Weights \& Biases, an increasingly popular deep learning experiment management service \citep{wandb}. 

\subsection{Class Imbalance Adjustment}
We compensate for the class imbalance discussed in~\Cref{sec:tasks} by reweighing the cross-entropy portion of each objective function, placing more weight on the minority class based on the relative class frequencies in each mini-batch of $M$ samples, $p(k)_{\text{mini-batch}}$~\citep{leibig2017leveraging}:
\begin{align}
\SwapAboveDisplaySkip
    \mathcal{L} = - \frac{1}{KM} \sum_{i=1}^{M} \frac{\mathcal{L}_{\text{cross-entropy}}(i)}{p(k)_{\text{mini-batch}}},
\end{align}
where $k$ is the class of sample $i$.
We also tried using constant class weights, but found that this resulted in lower overall performance.

\subsection{Mean-Field Variational Inference Implementation}

We employ a set of standard optimizations to improve training stability for the \mfvi and \textsc{radial}-\mfvi methods. 
We fix the mean of the prior to that of the variational posterior, which causes the KL term to only penalize the standard deviation of the weight posterior, and not its mean.
We use flipout for lower-variance gradients in convolutional layers and the final dense layer~\citep{wen2018flipout}, and KL annealing using a cyclical schedule, following~\citep{bowman2015generating}.
Finally, for \textsc{radial}-\mfvi, the prior's standard deviation is by default set to the He initializer standard deviation $\sqrt{2 / \text{fan\_in}}$~\citep{neal1995bayesian}.

\subsection{Uncertainty Estimation and Related Work} 
\label{subsec:app_uncert_est_details}

The Monte Carlo estimator used to computed the total uncertainty is biased but consistent and commonly used in practice~\citep{blundell2015mfvi,depeweg2018decomposition,gal2016dropout}.
A model's aleatoric uncertainty, $\E [ \mathcal{H}( p(\by_\ast \vbar f(\bx_\ast ; \btheta) ) ) ]$ is estimated analogously, and the epistemic uncertainty can then be computed as the difference between the total and the aleatoric predictive uncertainty estimates.

Some other works consider uncertainty estimation in medical imaging.
\citet{WANG201934} uses test-time augmentation for uncertainty estimation, but captures only aleatoric uncertainty.
\cite{nair2020exploring, roy2018inherent} considers uncertainty estimation with a Monte Carlo dropout model but does not isolate how their various measures of uncertainty correspond to epistemic or aleatoric uncertainty. 
None of the above works contribute and open-source tasks designed to emulate real-world distribution shifts, nor do they implement and benchmark a significant number of baseline uncertainty quantification models considering both aleatoric and epistemic uncertainty.

\subsection{Receiver Operating Characteristic Curves}
\label{subsec:app_roc_details}
The ROC curve (e.g., see~\Cref{fig:country_shift}(\textbf{a}) and (\textbf{b})) illustrates the diagnostic ability of a binary classification system as a function of the discrimination threshold.
The curve is created by plotting the true positive rate (that is, the sensitivity) against the false positive rate (that is, $1 - \text{specificity}$).
The quality of the ROC curve can be summarized by the area under the curve, which ranges from $0.5$ (chance level) to $1.0$ (perfect classification).

\subsection{Selective Prediction}
\label{subsec:app_sel_pred_details}
For the purposes of selective prediction, a model with optimal uncertainty estimates on a given dataset would have uncertainty perfectly correlate rank-wise with the model error. 
For example, the image on which the model has the highest error should be assigned the highest uncertainty, the image with the second highest error should be assigned the second highest uncertainty, and so on.
On the other hand, the worst possible uncertainty estimates are random, which would be uninformative to referral.

Finally, we explain in more detail the dip observed at the right side of the referral curves using AUC as the base metric (e.g.,~\Cref{fig:country_shift}\textbf{(c)} and \textbf{(d)}).
At relatively high referral rates $\tau$, models begin to refer examples on which they are both confident and correct.
This results in the referral curve decreasing.
At the highest $\tau$ values (the last few examples), for many models, nearly all remaining predictions are correct with high certainty, and the AUC increases.

\subsection{Hyperparameter Tuning}
\label{subsec:tuning_eval}

We provide full tuning details so that users of \retina will be able to reproduce our results.

All tuning scripts across all methods, tasks (\textit{Country} and \textit{Severity Shift}), and tuning procedures (on in-domain validation AUC and area under the selective prediction accuracy curve using the joint validation dataset, described in~\Cref{sec:joint_tuning}) are documented in the Uncertainty Baselines repository.\footnote{\href{https://github.com/google/uncertainty-baselines/tree/main/baselines/diabetic_retinopathy_detection}{\scriptsize\texttt{https://github.com/google/uncertainty-baselines/tree/main/baselines/diabetic\_retinopathy\_detection}}}

We tuned each model with a quasi-random search on several hyperparameters including learning rate, momentum, $\ell_2$ regularization, and method-specific variables including dropout rate and variational posterior initializations. 
We used a minimum of 32 trials per model.
Because of the large size of the input data and significant expense of multiple Monte Carlo samples at training time for some of the variational methods (in particular, \mfvi, \textsc{rank}-1, and \textsc{radial}-\mfvi), we were unable to achieve a large batch size with multiple Monte Carlo samples at training time.
With a single Monte Carlo sample at training time, we were able to fit more reasonable batch sizes ($\geq 64$) and found this to significantly improve convergence and performance on validation metrics.
We attribute this to the batch size increase and the usage of variance reduction techniques such as flipout layers~\citep{wen2018flipout}, which mitigate the impact of only using a single Monte Carlo sample at training time. 

We considered model selection for each of the models on each of the two tasks (\textit{Country} and \textit{Severity Shift}) using two different validation metrics: in-domain validation AUC, and area under the accuracy referral curve constructed using both in-domain and distributionally shifted validation data.
We describe the reasoning behind the latter metric in~\Cref{sec:joint_tuning}.
We used this validation performance to select the best hyperparameter setting and retrained a configuration for each combination of model, task, and validation tuning metric for $6$ random seeds.
We evaluated single models by averaging performance over those seeds, and evaluated ensembles by randomly sampling ensembles of size $3$ without replacement from the $6$ available models, and averaging over $6$ such ensemble constructions.
As described in~\Cref{subsec:eval_protocol}, for evaluation, we use five Monte Carlo samples per model to estimate predictive means (e.g., the \mcd \textsc{ensemble} with $K = 3$ ensemble members uses a total of $S=15$ Monte Carlo samples).

\paragraph{Compute Resources.} 
The majority of methods were tuned on TPU v2-8 nodes. 
\mfvi had particularly high memory requirements which required the use of TPU v3-8 nodes to achieve a reasonable batch size and stable training.
Evaluation was performed on NVIDIA A100 GPUs with 40 GB memory, though GPUs with standard sizes (e.g., >6 GB) will be sufficient to run evaluation and inference with the models in the benchmark, e.g., using the model checkpoints.
Approximately 100 TPU days and 20 GPU days were used collectively across the initial hyperparameter tuning, fine-tuning with selected configurations, and evaluation across the various tasks.
Though a significant cost, we hope that our open-sourcing of all code along with hyperparameter sweep details and checkpoints will significantly decrease future consumption of researchers interested in designing deep models for diabetic retinopathy, along with Bayesian deep learning researchers using our configurations to inform their hyperparameter tuning, or our generally applicable evaluation utilities.

\clearpage

\subsection{EyePACS and APTOS Input Data Examples}
\label{subsec:app_input_data_examples}

\begin{figure}[h!]
\begin{minipage}{\linewidth}
\centering
\begin{subfigure}[c]{0.7\linewidth}
\vspace*{-15pt}
    \includegraphics[width=\linewidth]{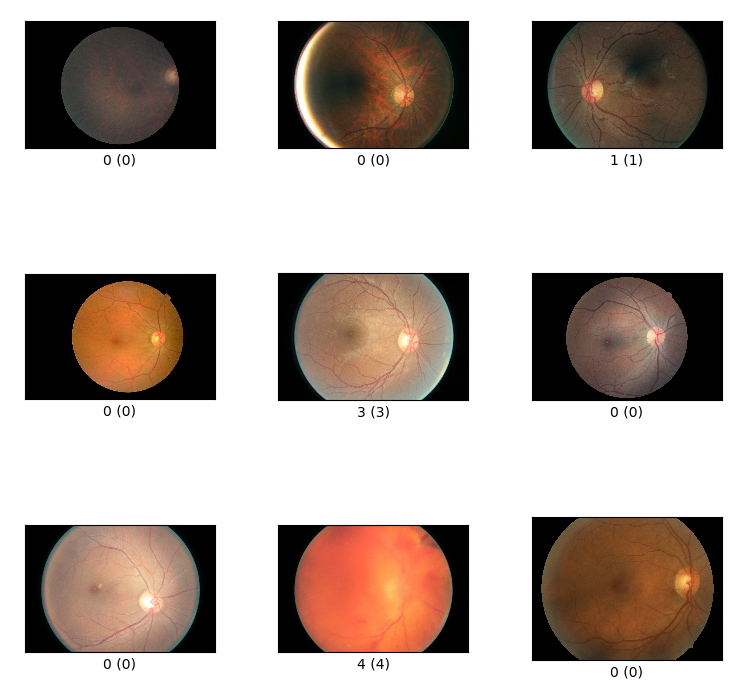}
  \centering
  \caption{
  	Original samples from the EyePACS Diabetic Retinopathy dataset \cite{kaggle_2015}.
  	}
\end{subfigure}

\end{minipage}
\centering
\begin{subfigure}[c]{0.7\linewidth}
    \includegraphics[width=\linewidth]{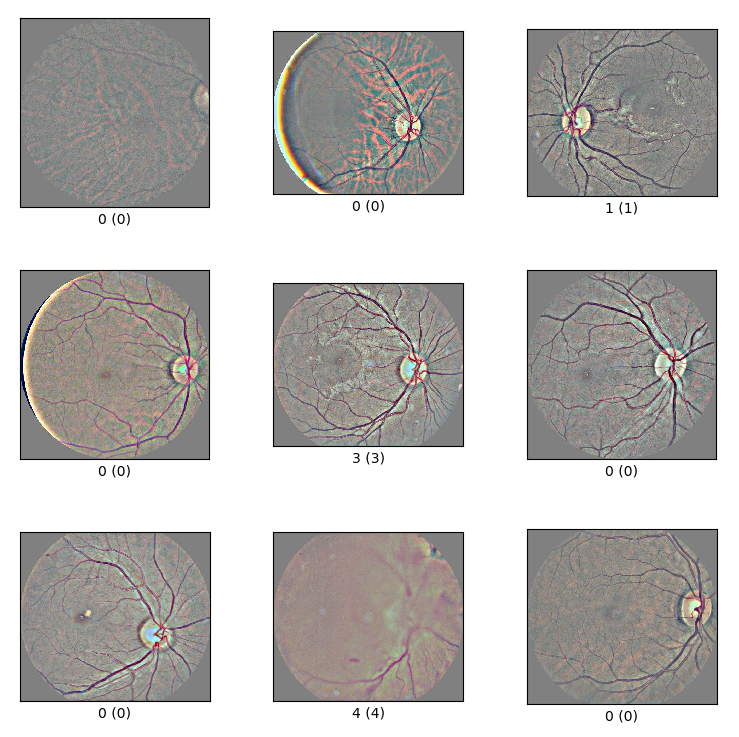}
\vspace*{-15pt}
  \centering
  \caption{
  	Processed and augmented samples from the EyePACS Diabetic Retinopathy dataset, following the procedure of the Kaggle competition winner \cite{kaggle_2015}.
  	}
\end{subfigure}
\begin{minipage}{\linewidth}
\centering

\vspace{5pt}
\centering
\caption{
    Illustrative examples of retina images in the original EyePACS dataset (top) and after preprocessing (bottom).}
\label{fig:processed}
\end{minipage}
\vspace*{-10pt}
\end{figure}

\begin{figure}[h!]
\centering
  \includegraphics[height=0.35\linewidth]{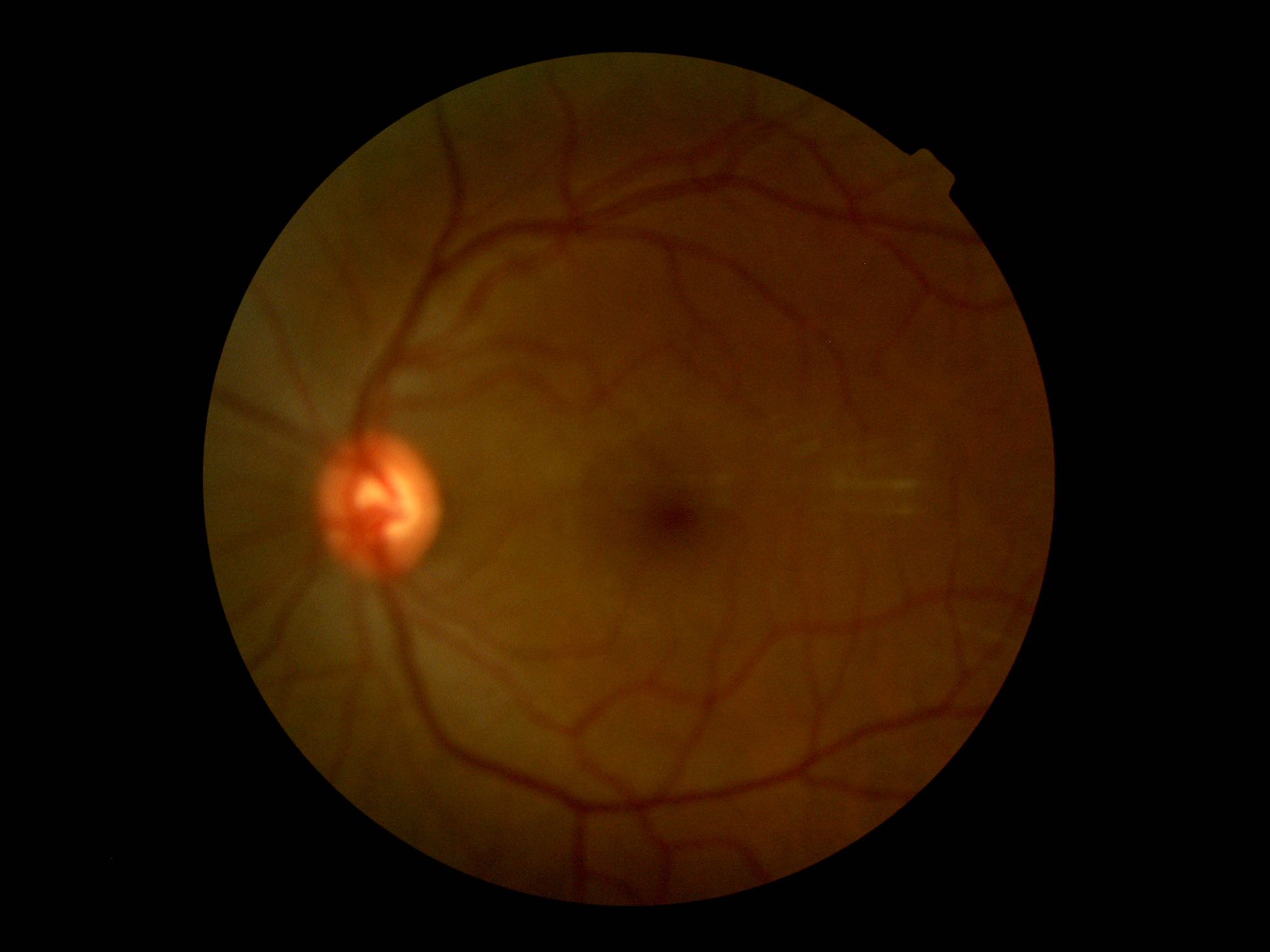}
  \includegraphics[height=0.35\linewidth]{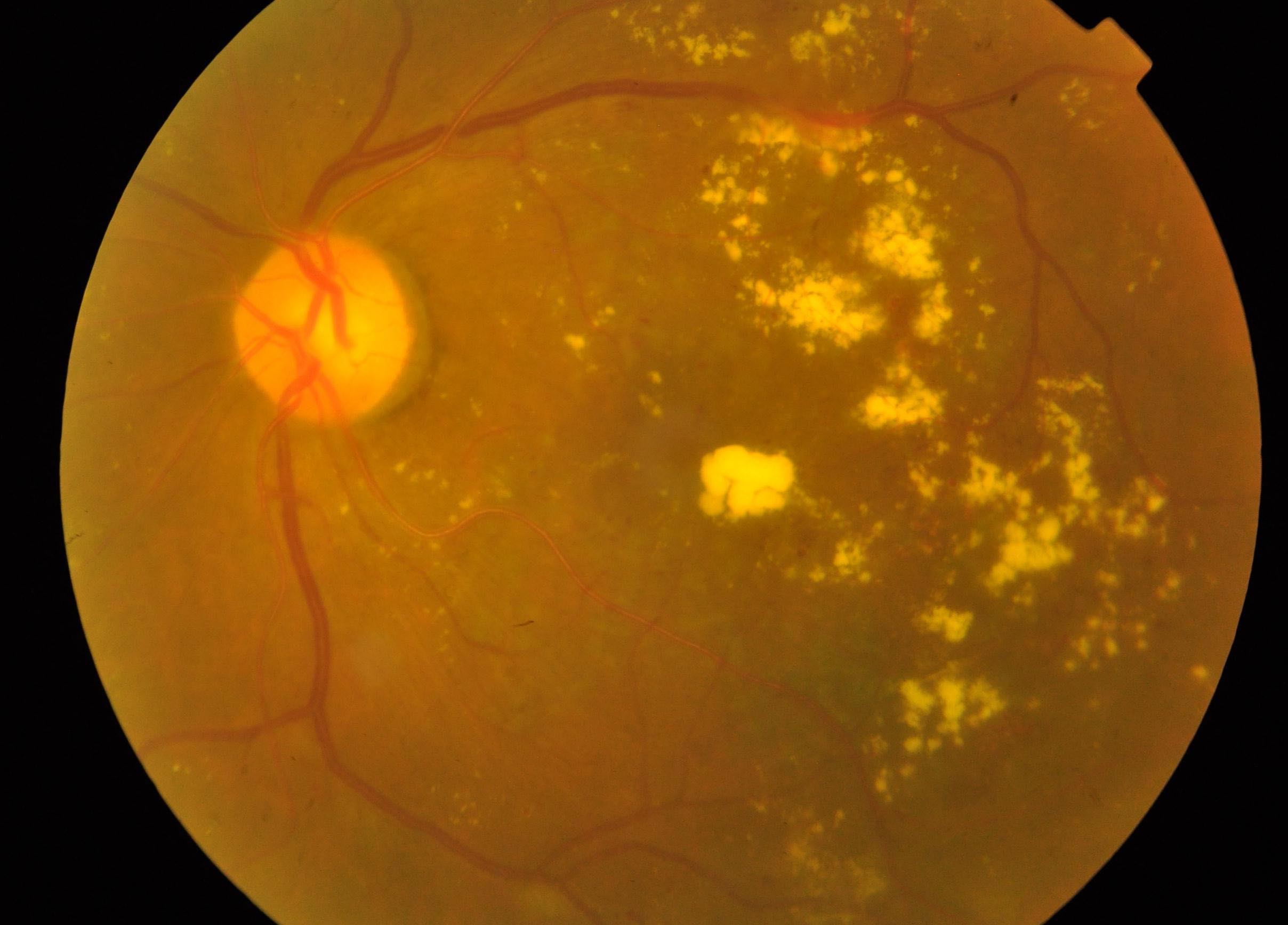}
  \includegraphics[height=0.317\linewidth]{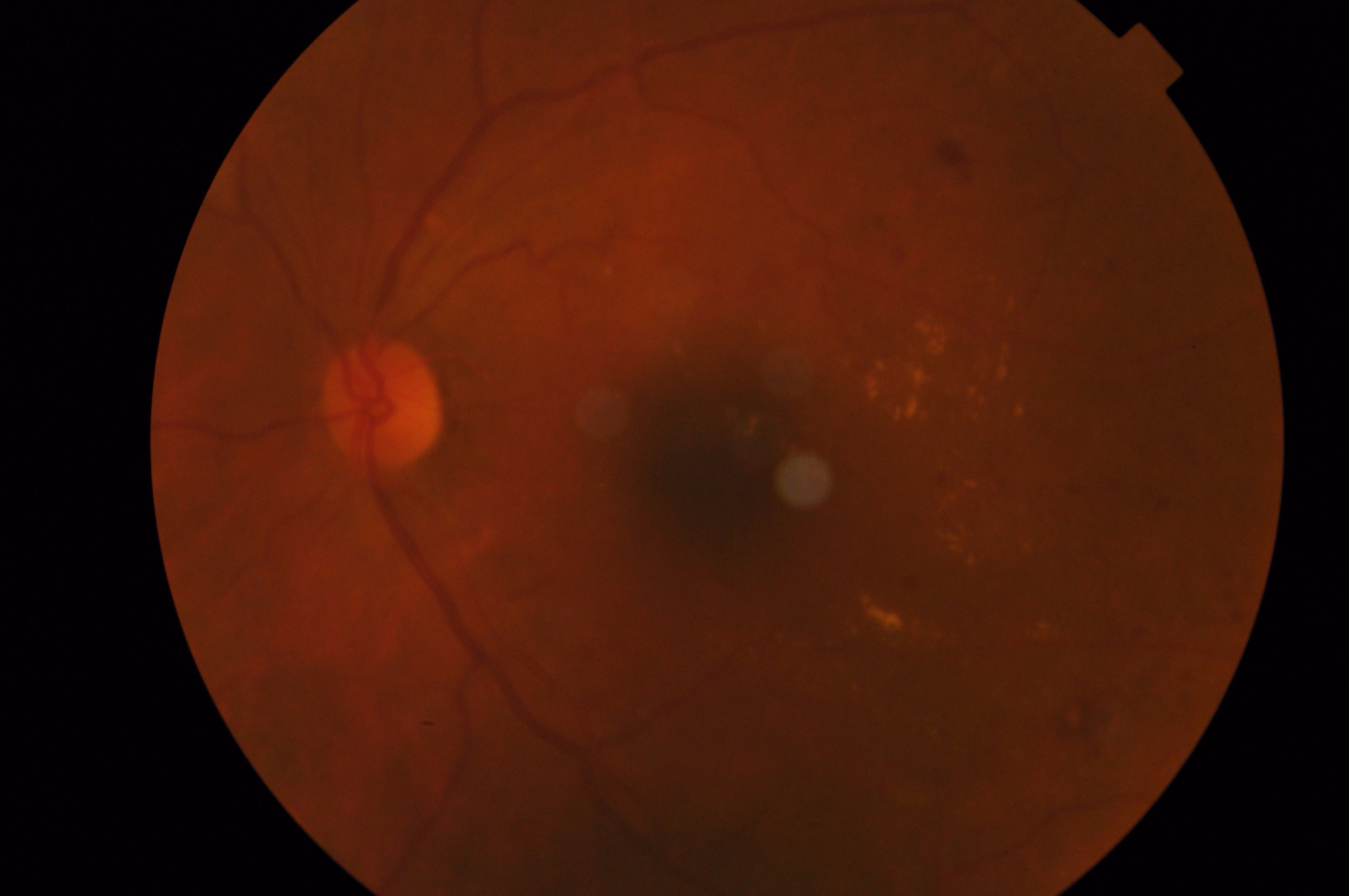}
  \includegraphics[height=0.317\linewidth]{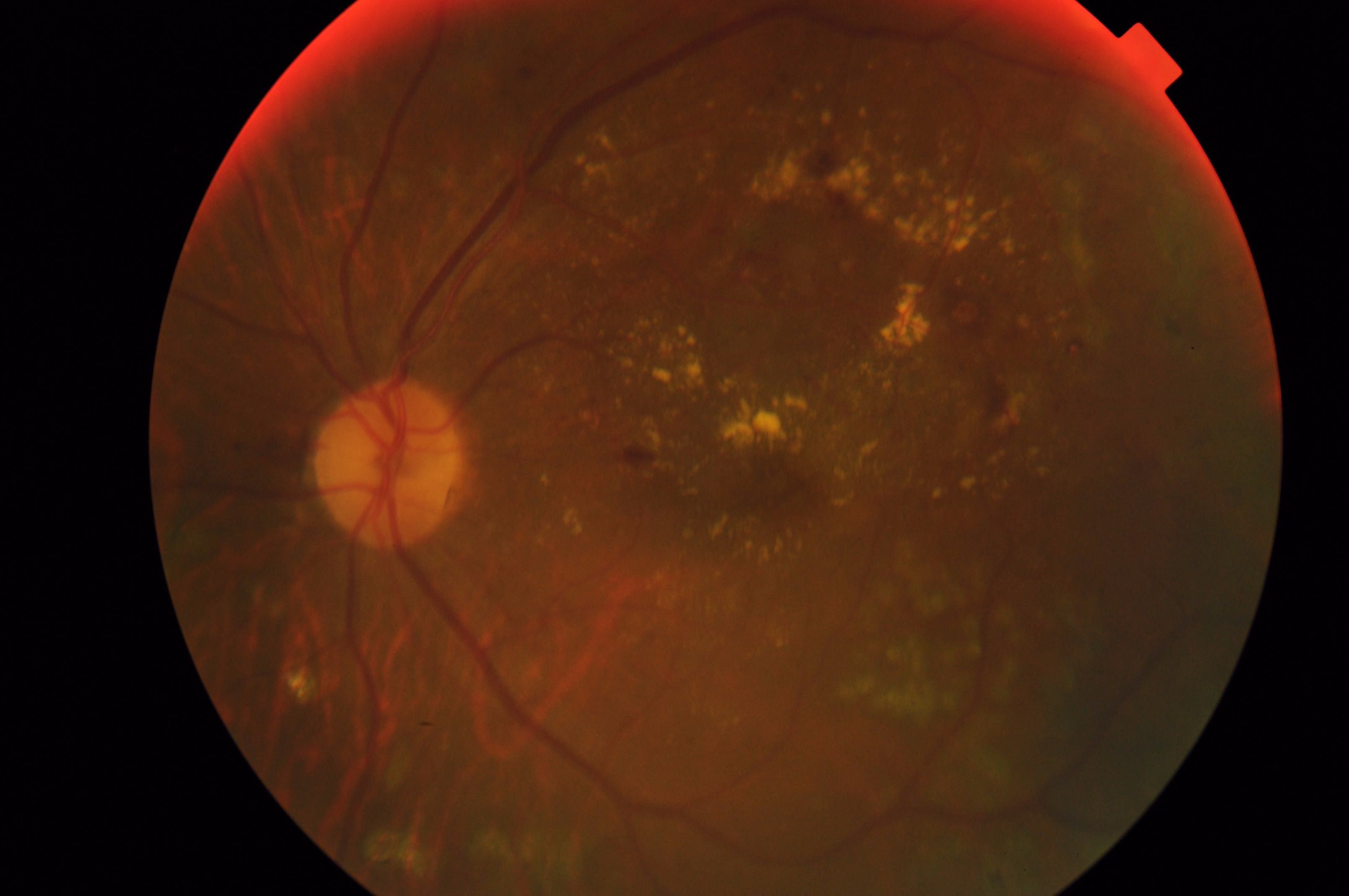}
  \includegraphics[height=0.352\linewidth]{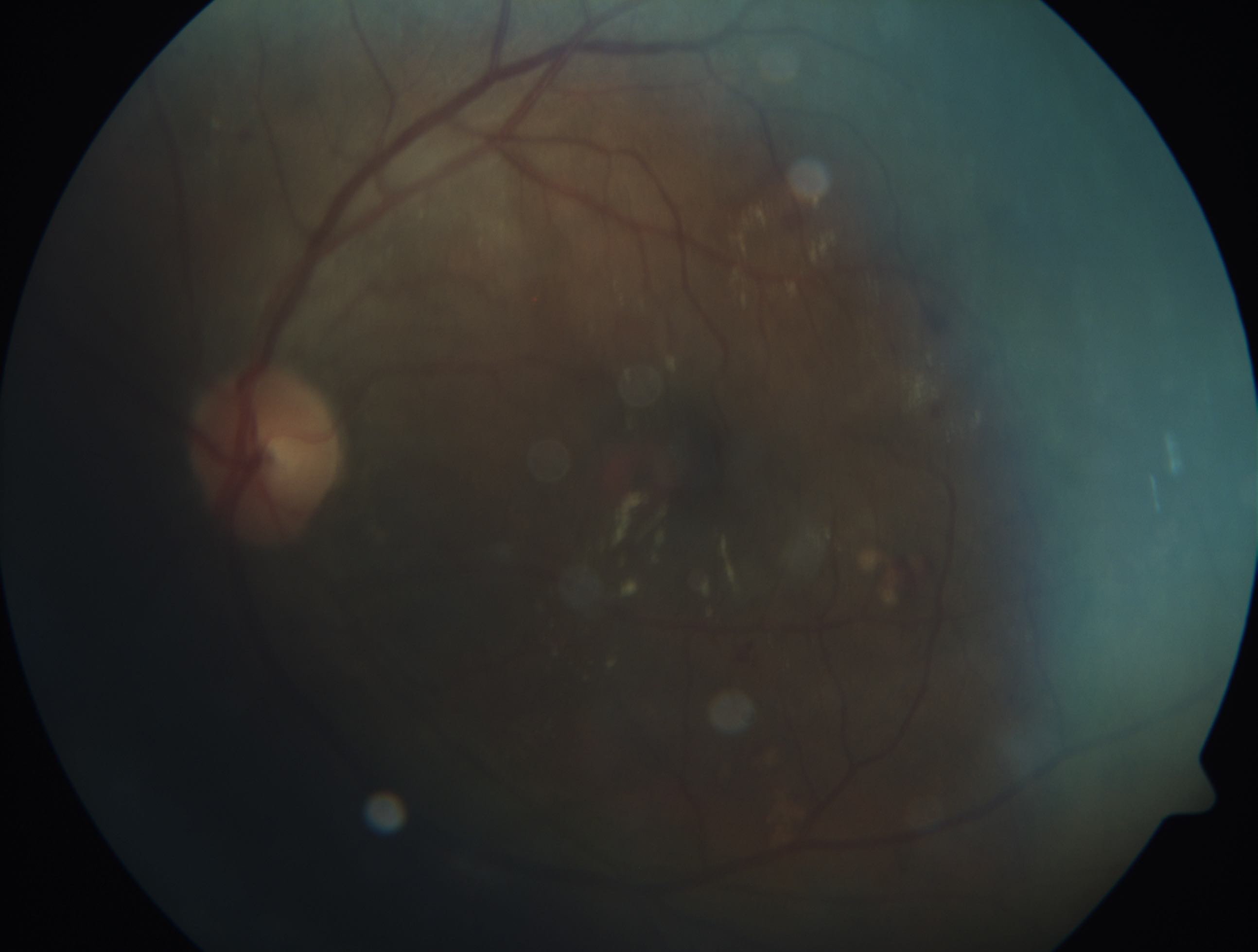}
  \includegraphics[height=0.352\linewidth]{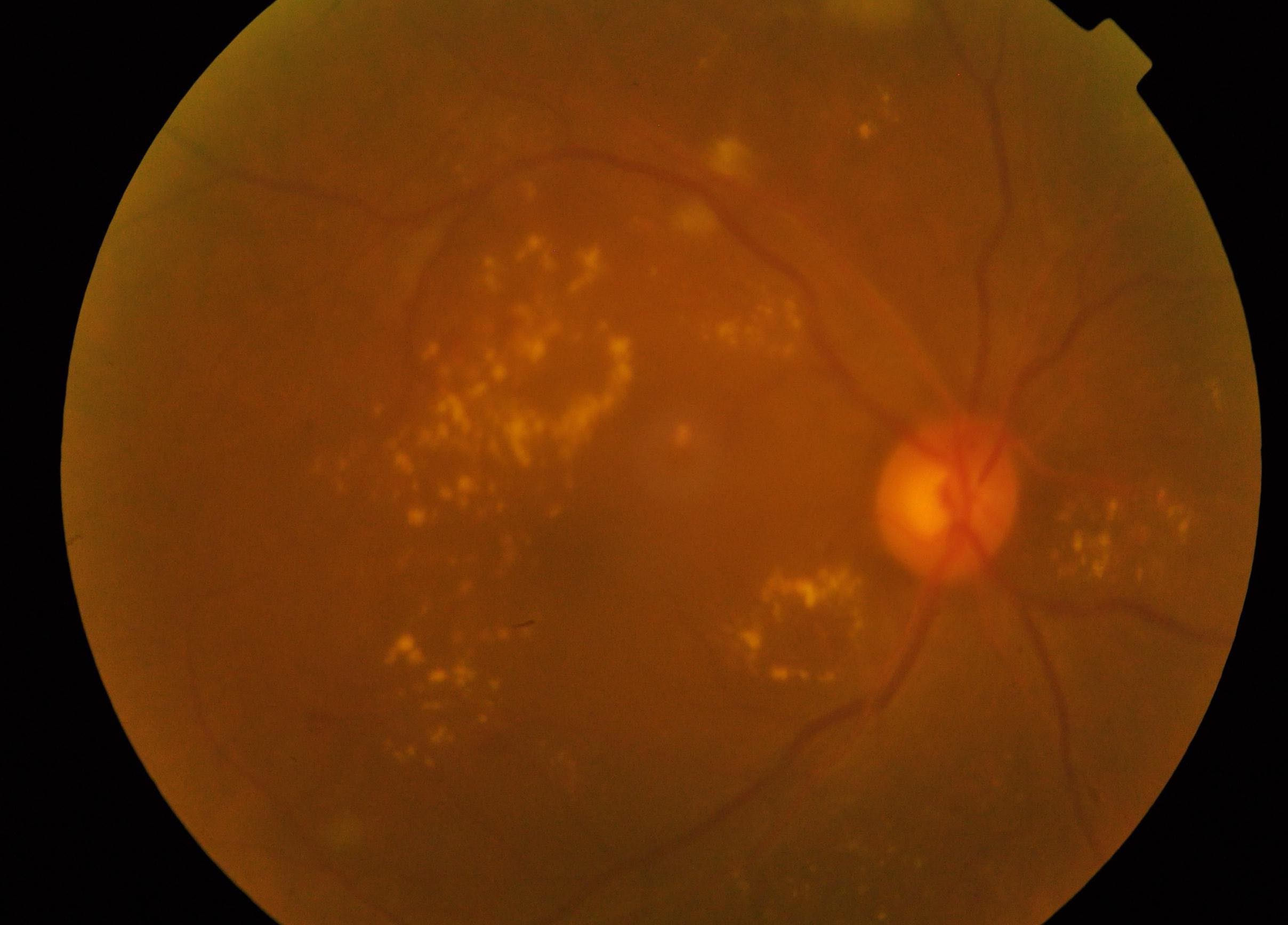}
\centering
\caption{
    Illustrative examples of retina images in the APTOS dataset. The images are collected using different measurement devices than the EyePACS dataset.
    Note the artifacts present in the images including blur, low background lighting, and effects around the edges of the retina.
    }
\label{fig:aptos_images}
\vspace*{-10pt}
\end{figure}

\clearpage

\section{Further Empirical Results}

\subsection{Predictive Uncertainty Histograms}
\label{sec:app_single_model_pred_uncert_hist}

In the figures below, predictive uncertainty (cf.~\Cref{sec:uncertainty_estimation}) is displayed as a normalized density for correct (blue) and incorrect (red) predictions.
All histograms are normalized and are displayed with the same range on the $x$- and $y$-axis.
Some bars of the histograms are cut off because the plots are zoomed-in along the $y$-axis to improve legibility.
See~\Cref{subsec:pred_uncert_hist} for a description of predictive uncertainty histograms as a model diagnostic tool, including a discussion of the expected behavior of reliable models.
See~\Cref{sec:benchmark} for a discussion of the results for single models on the shifted datasets.

\begin{figure}[h!]
\vspace*{-10pt}
\centering
  \includegraphics[width=0.84\linewidth]{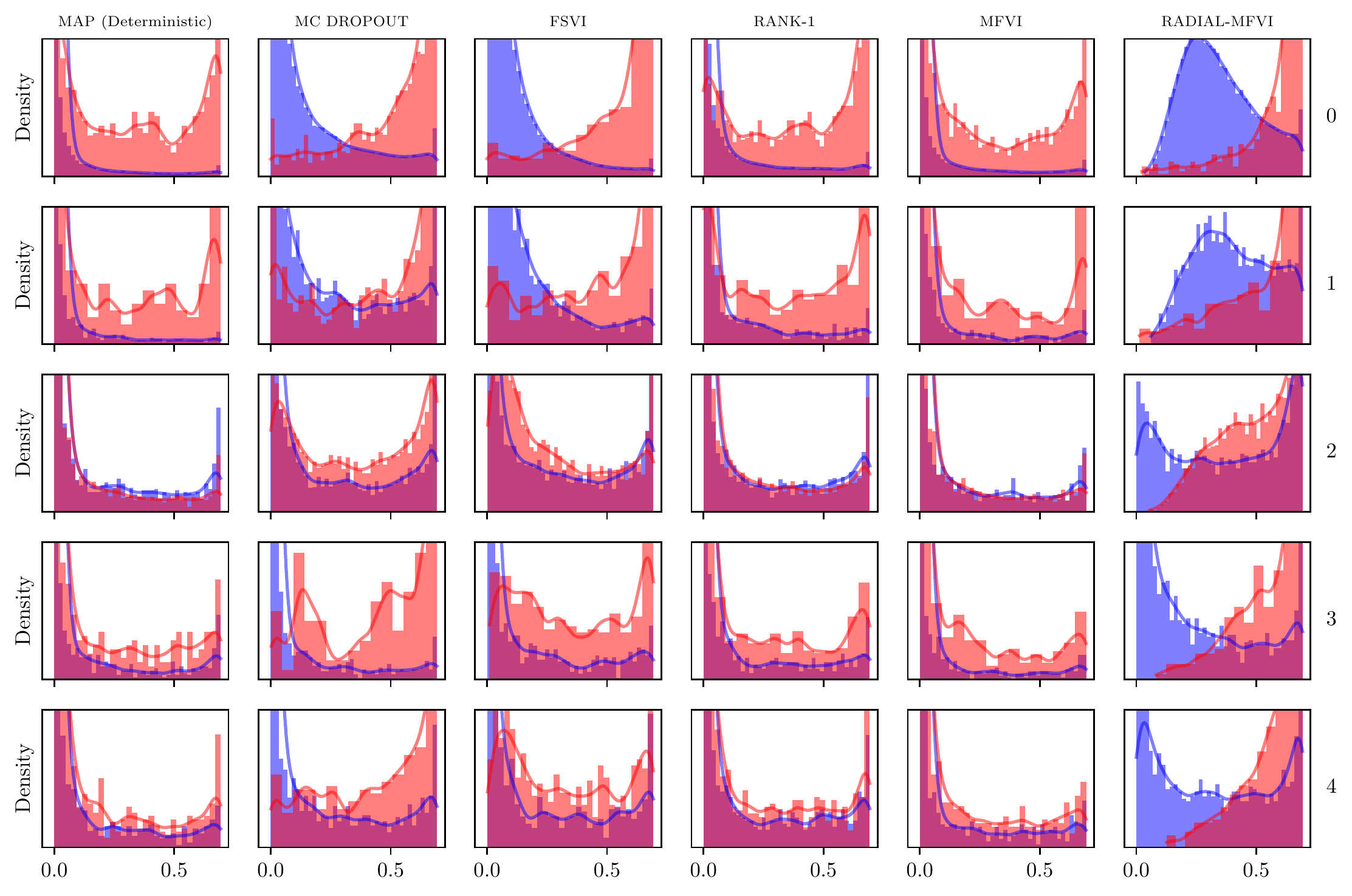}
\caption{
    \textbf{Clinical Label Binning -- Severity Shift, Single Models.} We analyze predictive uncertainty for each underlying clinical severity label (rows, label on right) and each uncertainty quantification method (columns).
    Here, we consider both the in-domain and distributionally shifted Severity Shift evaluation datasets, and single models ($K = 1$).
    Predictive uncertainty, as measured by total uncertainty (cf.~\Cref{sec:uncertainty_estimation}), is displayed as a normalized density for correct (blue) and incorrect (red)  predictions.
    }
\label{fig:label_bin_sev_shifted_single}
\vspace*{-10pt}
\end{figure}

\begin{figure}[h!]
\vspace*{-10pt}
\centering
  \includegraphics[width=0.84\linewidth]{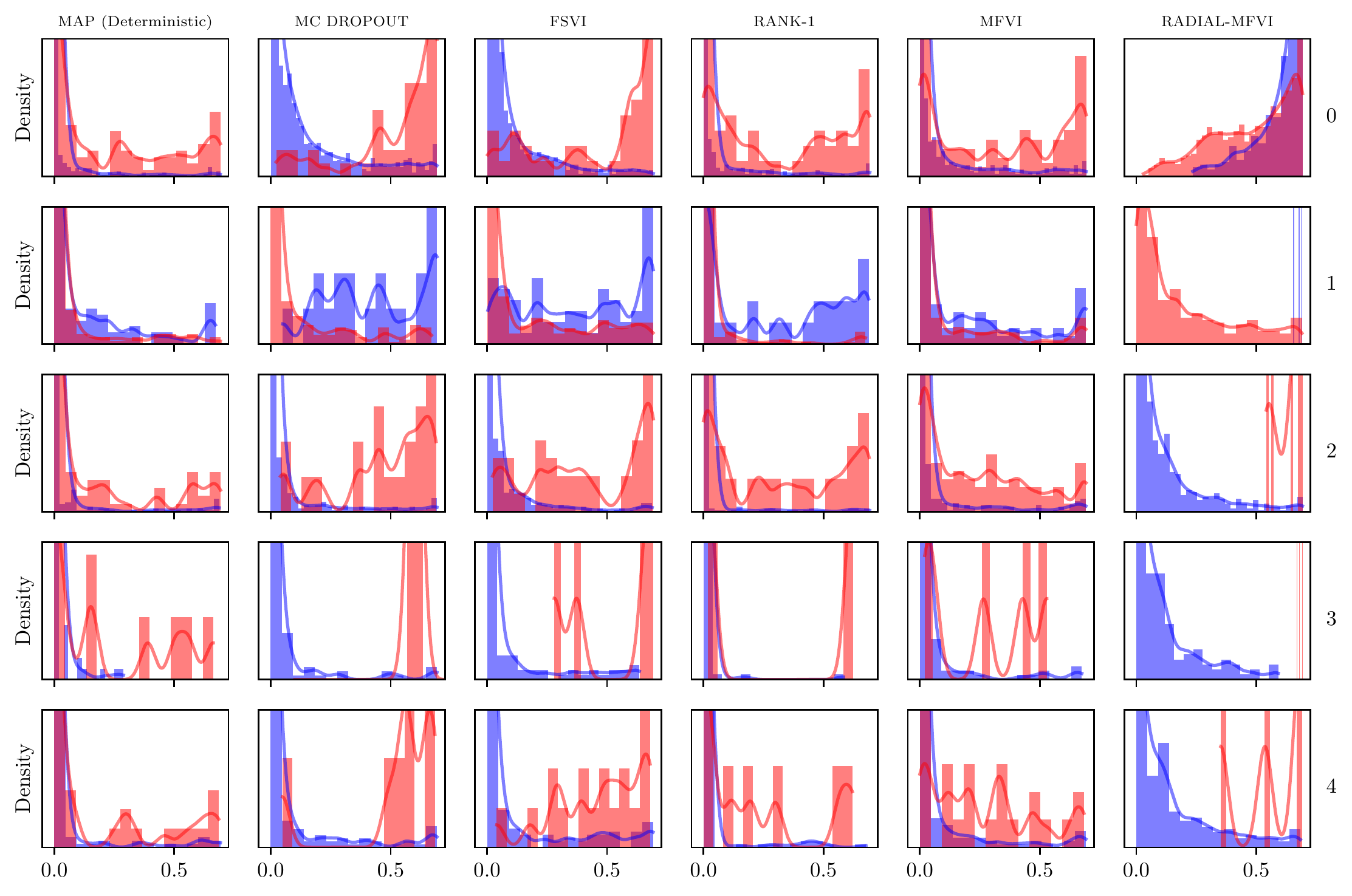}
\caption{
    \textbf{Clinical Label Binning -- Country Shift (Shifted), Single Models.} We analyze predictive uncertainty for each underlying clinical severity label (rows, label on right) and each uncertainty quantification method (columns).
    Here, we consider the distributionally shifted Country Shift evaluation dataset (APTOS), and single models ($K = 1$).
    Predictive uncertainty, as measured by total uncertainty (cf.~\Cref{sec:uncertainty_estimation}), is displayed as a normalized density for correct (blue) and incorrect (red)  predictions.
    }
\label{fig:label_bin_country_shifted_single}
\vspace*{-60pt}
\end{figure}

\FloatBarrier

\clearpage

\begin{figure}[h!]
\centering
  \includegraphics[width=0.84\linewidth]{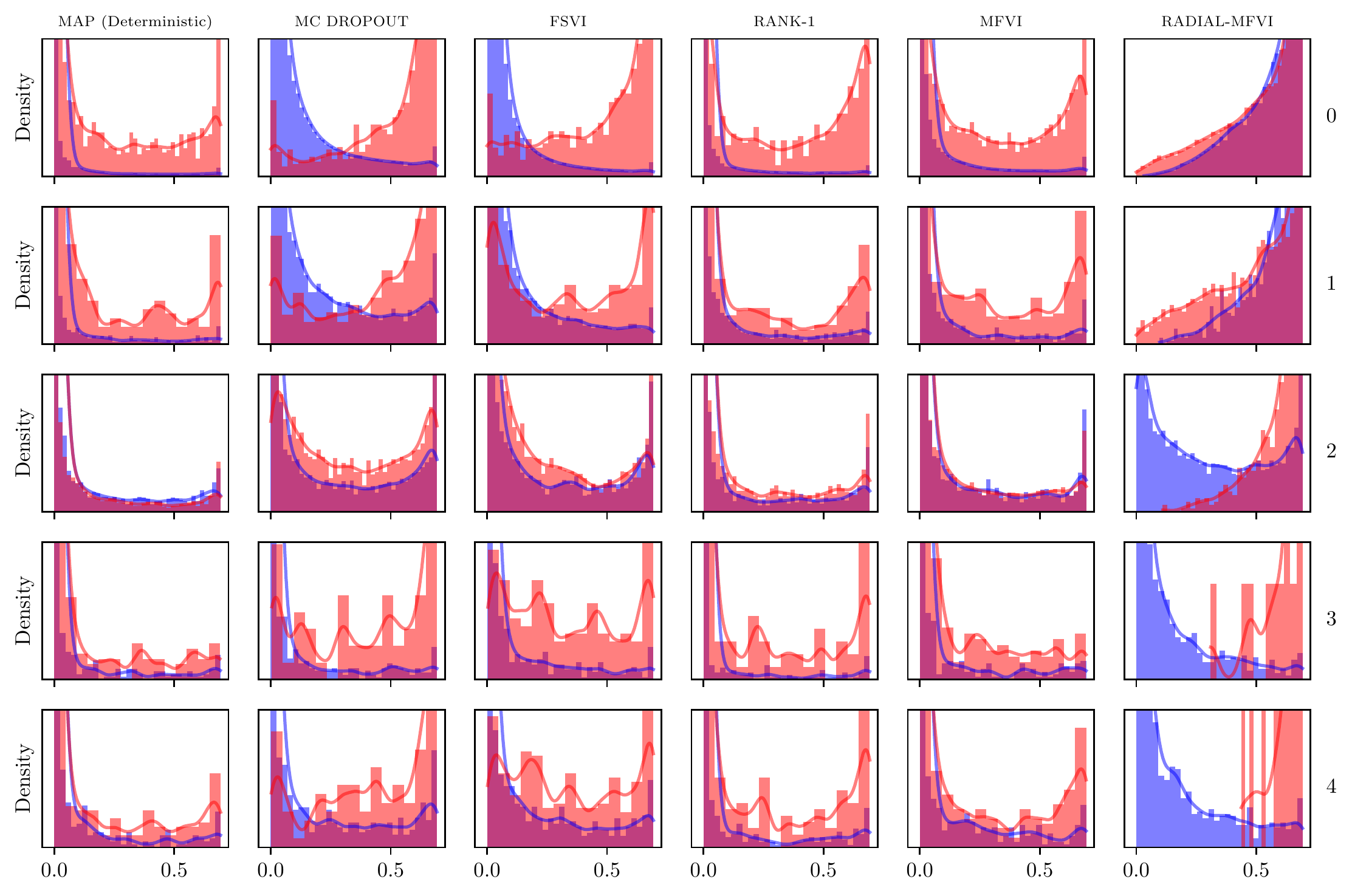}
\centering
\caption{
    \textbf{Clinical Label Binning -- Country Shift (In-Domain), Single Models.} We analyze predictive uncertainty for each ground-truth clinical label (rows) and each uncertainty quantification method (columns).
    Here, we consider the in-domain Country Shift evaluation dataset, and single models ($K = 1$).
    Predictive uncertainty, as measured by total uncertainty (cf. \Cref{sec:uncertainty_estimation}), is displayed as a normalized density for correct (blue) and incorrect (red)  predictions.
    }
\label{fig:label_bin_country_in_domain_single}
\vspace*{-10pt}
\end{figure}

\begin{figure}[h!]
\centering
  \includegraphics[width=0.84\linewidth]{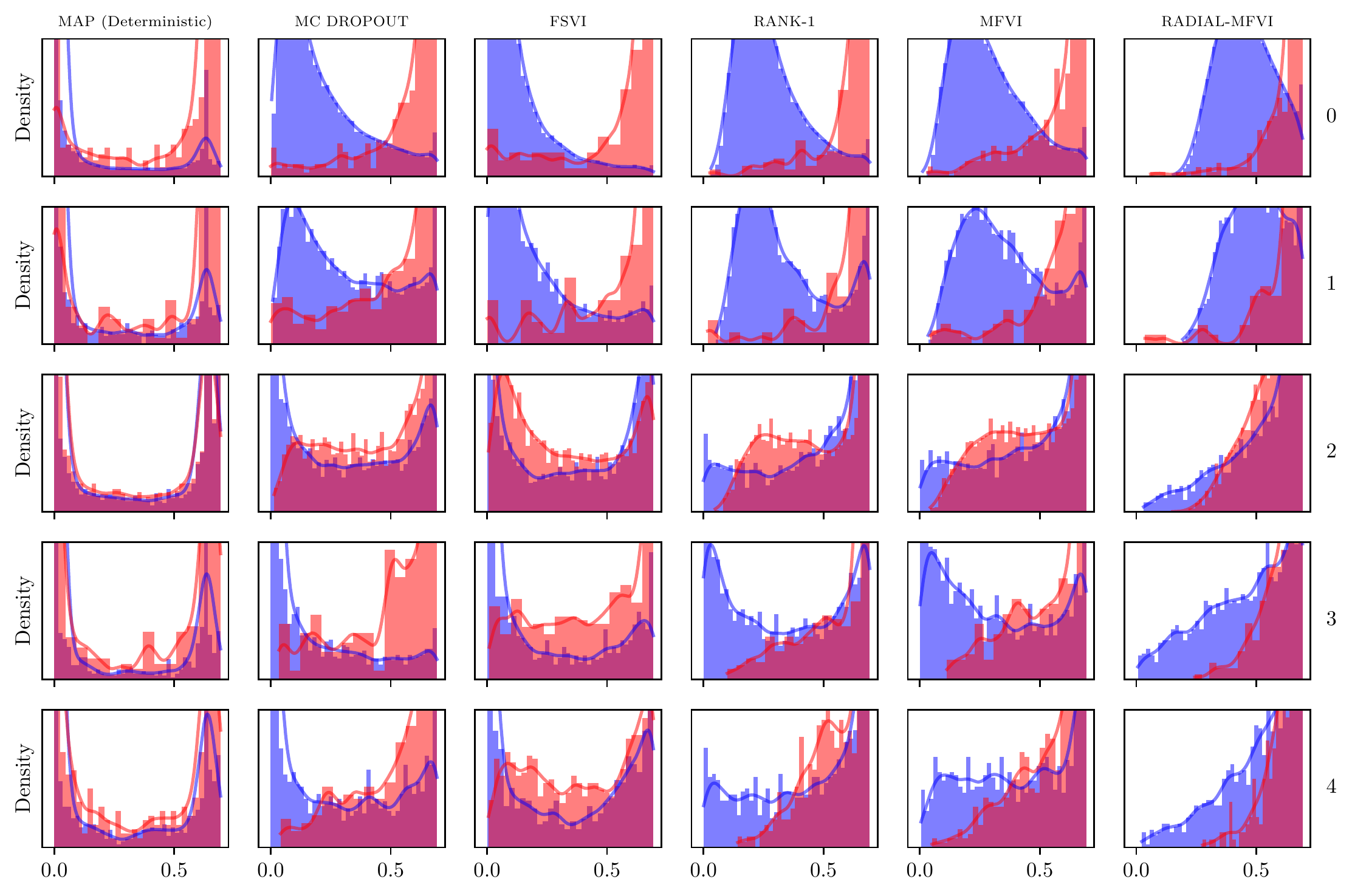}
\centering
\caption{
    \textbf{Clinical Label Binning -- Severity Shift, Ensembles.} We analyze predictive uncertainty for each ground-truth clinical label (rows, label on right) and each uncertainty quantification method (columns).
    Here, we consider both the in-domain and distributionally shifted Severity Shift evaluation datasets, and ensembles ($K = 3$).
    Predictive uncertainty, as measured by total uncertainty (cf. \Cref{sec:uncertainty_estimation}), is displayed as a normalized density for correct (blue) and incorrect (red)  predictions.
    }
\label{fig:label_bin_sev_shifted_ensemble}
\vspace*{10pt}
\end{figure}

\begin{figure}[h!]
\centering
  \includegraphics[width=0.84\linewidth]{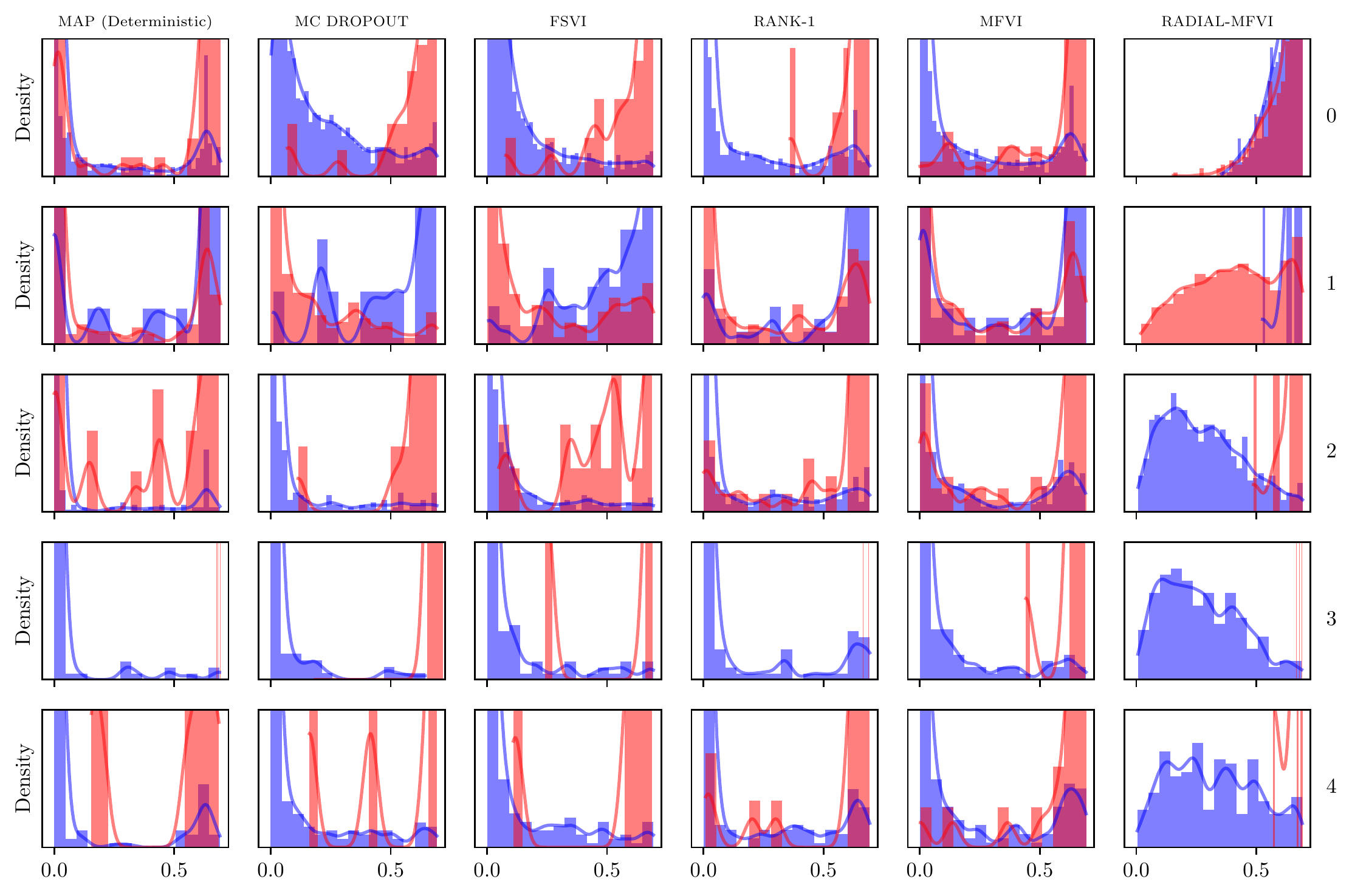}
\centering
\caption{
    \textbf{Clinical Label Binning -- Country Shift (Shifted), Ensembles.} We analyze predictive uncertainty for each ground-truth clinical label (rows, label on right) and each uncertainty quantification method (columns).
    Here, we consider the distributionally shifted Country Shift evaluation dataset (APTOS), and ensembles ($K = 3$).
    Predictive uncertainty, as measured by total uncertainty (cf. \Cref{sec:uncertainty_estimation}), is displayed as a normalized density for correct (blue) and incorrect (red)  predictions.
    }
\label{fig:label_bin_country_shifted_ensemble}
\vspace*{-10pt}
\end{figure}

\begin{figure}[h!]
\centering
  \includegraphics[width=0.84\linewidth]{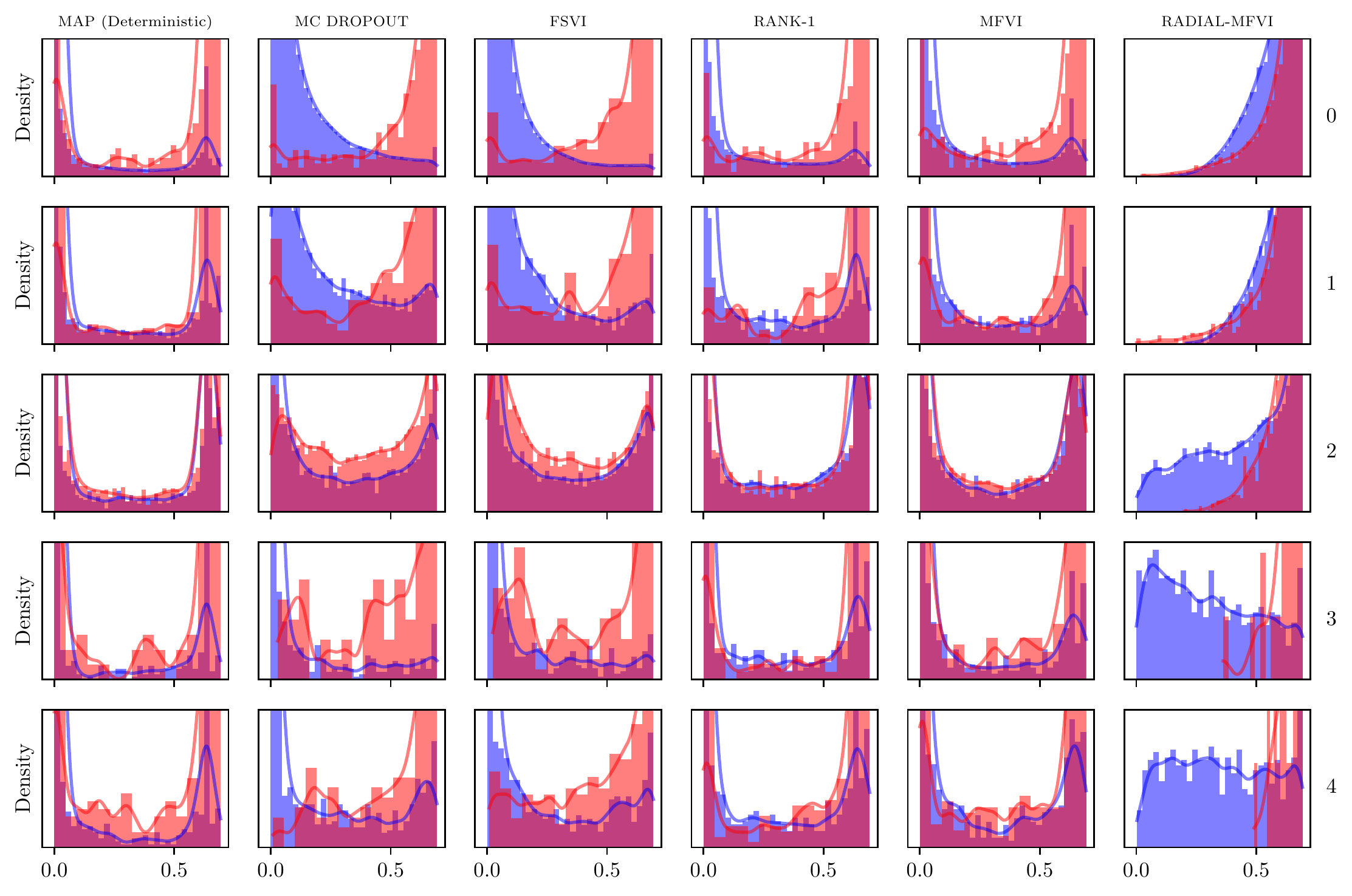}
\centering
\caption{
    \textbf{Clinical Label Binning -- Country Shift (In-Domain), Ensembles.} We analyze predictive uncertainty for each ground-truth clinical label (rows, label on right) and each uncertainty quantification method (columns).
    Here, we consider the in-domain Country Shift evaluation dataset (APTOS), and ensembles ($K = 3$).
    Predictive uncertainty, as measured by total uncertainty (cf. \Cref{sec:uncertainty_estimation}), is displayed as a normalized density for correct (blue) and incorrect (red)  predictions.
    }
\label{fig:label_bin_country_in_domain_ensemble}
\vspace*{-10pt}
\end{figure}

\clearpage

\subsection{Tuning without Distributionally Shifted Data: Country Shift Accuracy.}
\begin{figure}[h!]
\vspace{10pt}
\hspace{5pt}
\centering
\begin{subfigure}[l]{0.55\linewidth}
    \hspace{-20pt}\includegraphics[width=\linewidth]{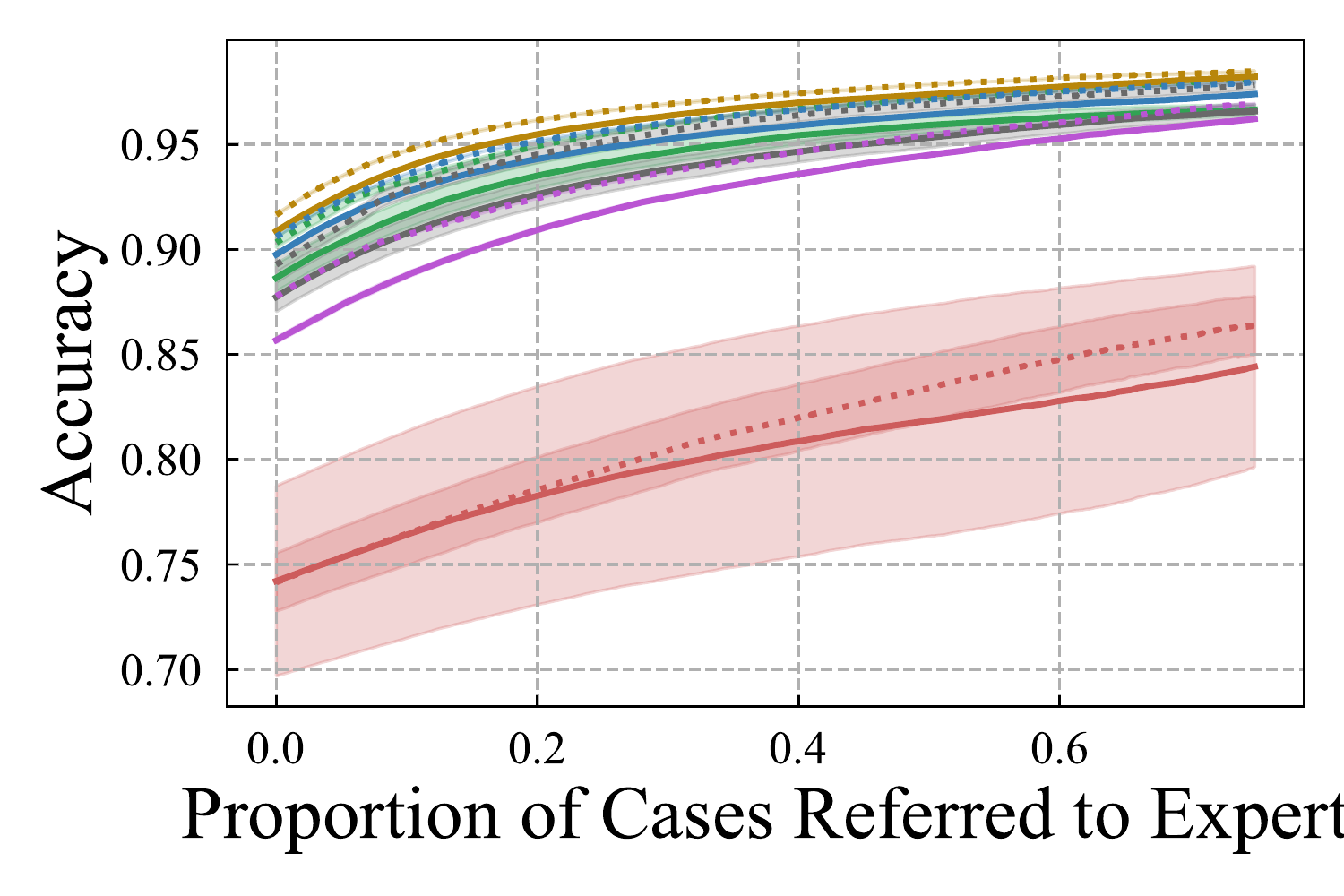}
    \caption{
        \textbf{Selective Prediction Accuracy: In-Domain}
    }
\end{subfigure}
\begin{subfigure}[r]{0.55\linewidth}
  \hspace{-20pt}\includegraphics[width=\linewidth]{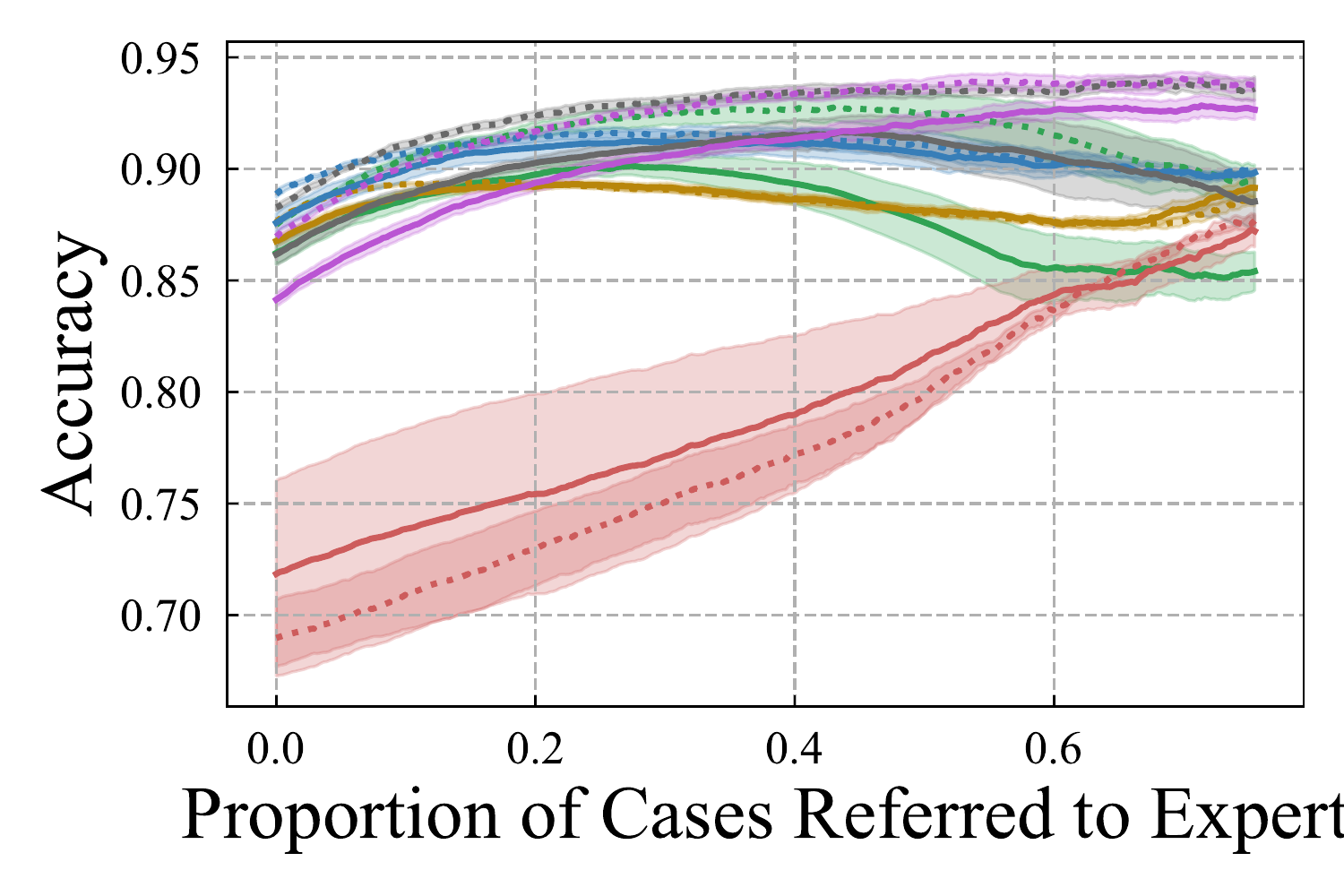}
    \caption{
        \textbf{Selective Prediction Accuracy: Country Shift}
    }
\end{subfigure}
\begin{subfigure}[r]{0.55\linewidth}
\hspace{-20pt}\includegraphics[width=\linewidth]{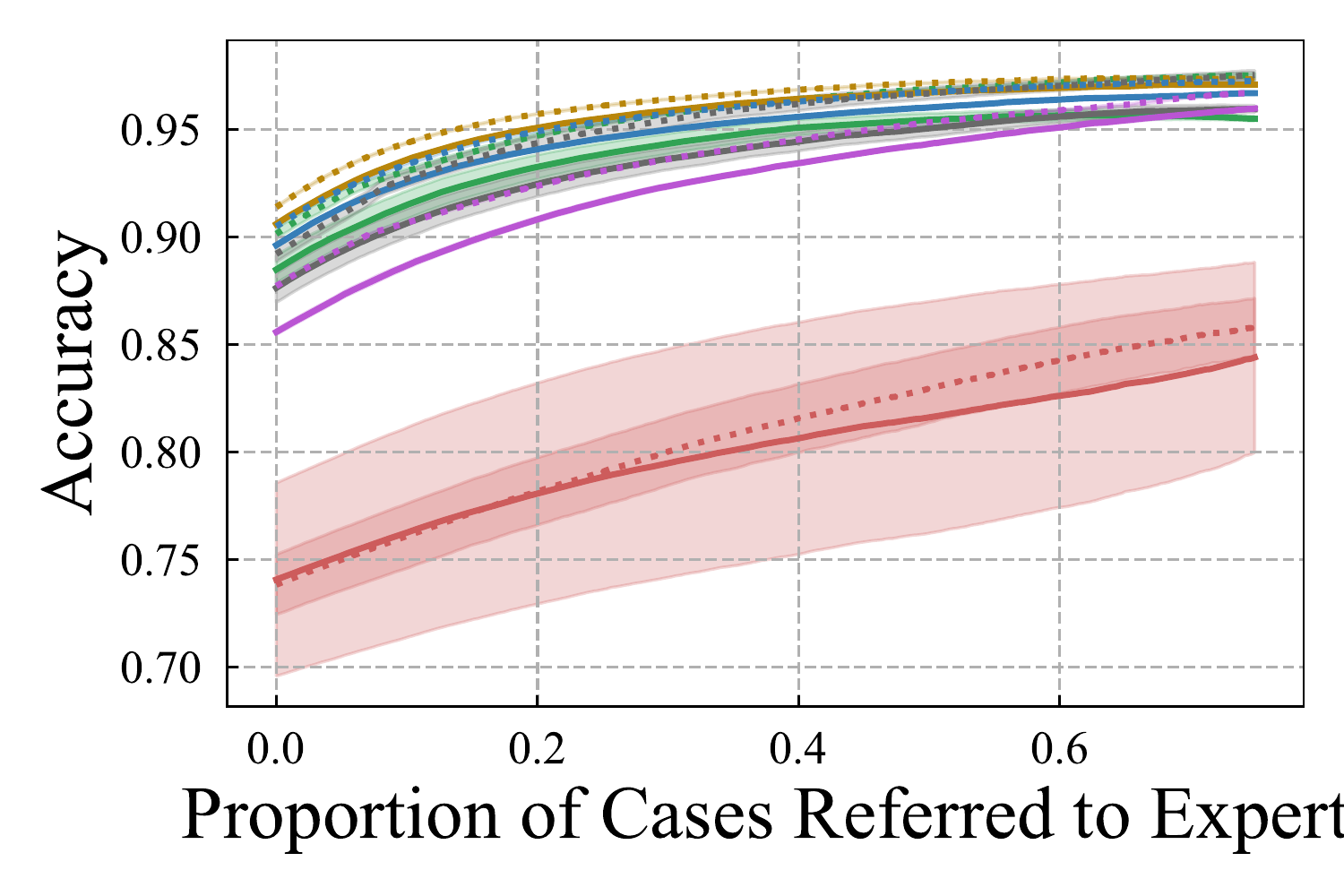}
\caption{
    \textbf{Selective Prediction Accuracy: Joint}
}
\end{subfigure}
\begin{subfigure}{\linewidth}
    \includegraphics[width=\linewidth]{fig/metrics/legend.pdf}
\end{subfigure}
\hspace*{-20pt}
\\
\vspace{10pt}

\centering
\caption{
    \textbf{Selective Prediction: Country Shift (Accuracy).} We use the binary accuracy for in-domain diagnosis on the~\citet{kaggle_2015} test set (\textbf{a}), for changing medical equipment and patient populations on the shifted~\citet{APTOS_2019} evaluation set (\textbf{b}), and on a joint dataset composed of both the in-domain and APTOS datasets (\textbf{c}).
    Shading denotes one standard error. 
}
\label{fig:country_shift_accuracy}
\end{figure}

We provide referral curves on accuracy for \textit{Country Shift} with in-domain validation tuning in~\Cref{fig:country_shift_accuracy}.

\clearpage

\subsection{Tuning in the Presence of Distributionally Shifted Data}
\label{sec:joint_tuning}

In prior work in Bayesian deep learning, little emphasis has been placed on the standardization of a training and evaluation protocol; in particular, the assumption of whether a model has access to distributionally shifted validation data for hyperparameter tuning is often changed on an ad-hoc basis across studies.

This is a significant assumption, and researchers in Bayesian deep learning should be expected to outwardly declare their tuning procedure---in particular access to distributionally shifted data---as is done in works such as Prior Networks~\citep{malinin-pn-2018, malinin-rkl-2019}.
This will permit researchers and practitioners to more fairly compare the performance of methods based on results reported in their respective papers.

\begin{figure}[h!]
\vspace{10pt}
\centering
\begin{subfigure}[l]{0.48\columnwidth}
  \includegraphics[width=\linewidth]{fig/metrics/metrics_new/retention-aptos-ood_test-indomain-Accuracy-no_oracle.pdf}
  \caption{
        \textbf{Accuracy: In-Domain Tuning}
    }
\end{subfigure}
\begin{subfigure}[r]{0.48\columnwidth}
  \includegraphics[width=\linewidth]{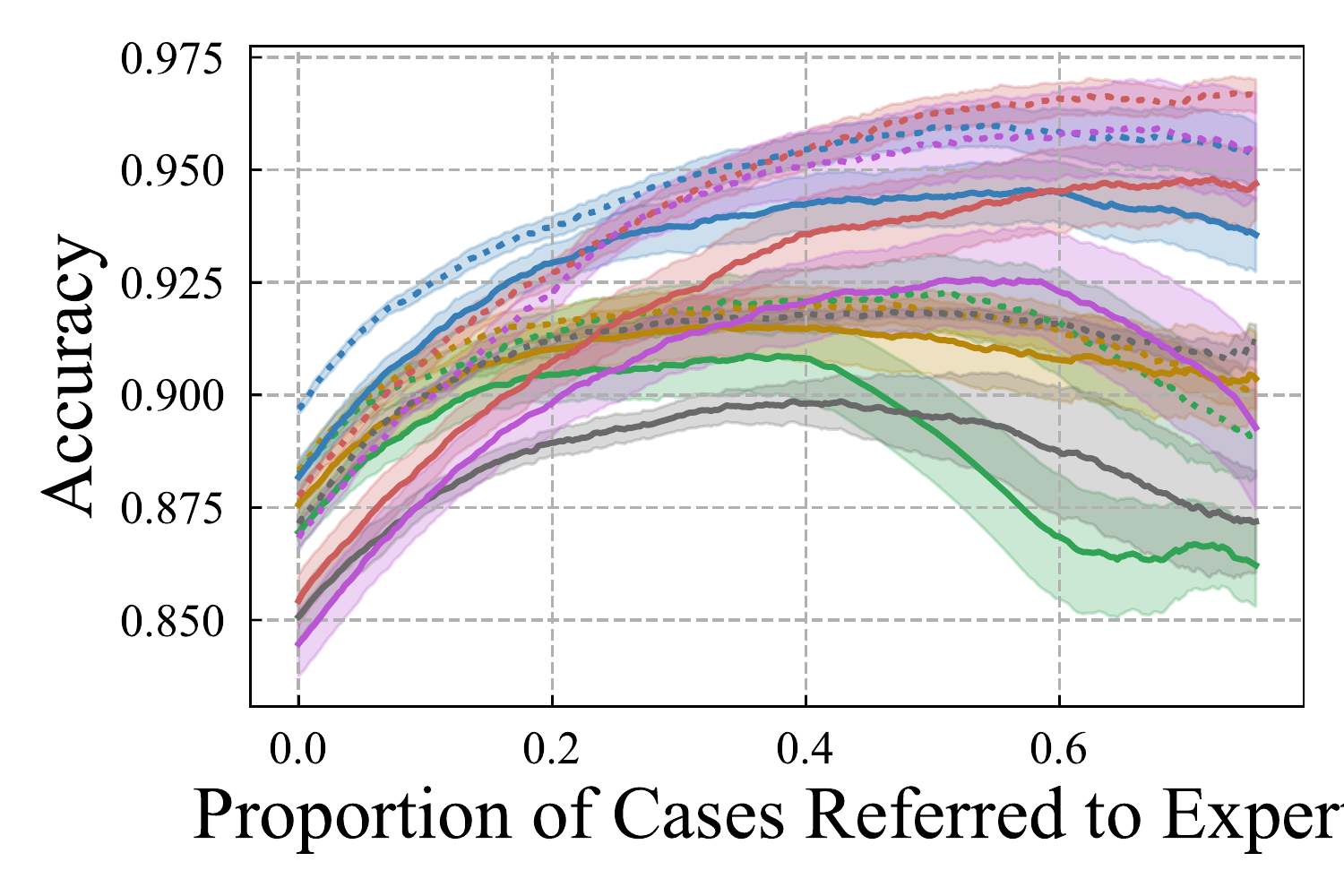}
  \caption{
        \textbf{Accuracy: Joint Tuning}
    }
\end{subfigure}
\begin{subfigure}{\linewidth}
    \includegraphics[width=\linewidth]{fig/metrics/legend.pdf}
\end{subfigure}
\centering
\caption{
    \textbf{Hyperparameter Tuning on Distributionally Shifted Data.}
    Accuracy referral curve on the distributionally shifted APTOS dataset in the Country Shift task.
    \textbf{Left:} Performance of various methods when using the in-domain validation AUC for hyperparameter tuning.
    \textbf{Right:} The same methods when using the proposed balanced referral metric evaluated over a combination of in-domain and distributionally shifted validation data. 
    Even without permitting a model to explicitly train on distributionally shifted data, the model selection process results in significantly improved predictive performance and quality of uncertainty estimates, as demonstrated by curves for respective methods shifted upwards, and steeper slopes in each curve as the first $\approx50\%$ of cases are referred to an expert, respectively.
}
\label{fig:shifted_validation_tuning_comparison}
\end{figure}

We investigate what impact this assumption---access to distributionally shifted validation data---has on downstream performance across all our tasks, and on held-out in-domain, distributionally shifted, and joint (in-domain combined with distributionally shifted) evaluation datasets. We find that it has a significant impact on metrics commonly used to assess robustness and uncertainty quantification, including area under referral curves (\Cref{fig:shifted_validation_tuning_comparison}) and expected calibration error.

\paragraph{Joint Validation Metric.}

To consider the performance of our baseline models under this assumption, we construct a metric that conveys both in-domain and distributionally shifted performance. 
In particular, we construct an accuracy referral curve on a combined set of in-domain and distributionally shifted validation examples.
Because the in-domain validation dataset is significantly larger than the distributionally shifted dataset for both of the tasks, we upsample the shifted dataset to avoid the signal from the in-domain examples overwhelming that from the shifted examples.
We construct an \textit{upsampled shifted dataset} by first duplicating the shifted validation dataset as many times as possible without exceeding the size of the in-domain validation dataset, and then randomly sampling examples from the shifted validation dataset without replacement until the upsampled shifted dataset contains the same number of examples as the in-domain validation dataset.
We construct the ``balanced'' joint validation dataset as the union of the in-domain validation and upsampled shifted datasets.
We construct a ``balanced'' accuracy referral curve using this balanced joint validation dataset, sweeping over $\tau$ to obtain all possible partitions of the dataset into ``referral'' and ``non-referral''.
We then tune on the area under this curve.

\clearpage

\subsection{Complete Tabular Results}
\label{subsec:more_experiments}

We report additional tabular results for standard predictive performance and robustness (expected calibration error), referral metrics, and out-of-distribution detection across the \emph{Severity} and \emph{Country} \emph{Shift} tasks, considering hyperparameter tuning on either in-domain validation AUC or the joint validation metric (cf.~\Cref{sec:joint_tuning}), in Tables 2-11.

\begin{table*}[b!]
\centering
\caption{
    \textbf{Severity Shift.}
    Prediction and uncertainty quality of baseline methods in terms of area under the receiver operating characteristic curve (AUC) and classification accuracy, as a function of the proportion of data referred to a medical expert for further review.
}
\vspace{-3pt}
\resizebox{1.0\linewidth}{!}{%
\begin{tabular}{@{\extracolsep{2pt}}lcccccc@{}}
\midrule
\midrule
& \multicolumn{2}{c}{No Referral} & \multicolumn{2}{c}{$50\%$ Data Referred} & \multicolumn{2}{c}{$70\%$ Data Referred} \\
\cline{2-3}
\cline{4-5}
\cline{6-7}\\
\textbf{Method}         &
\textbf{AUC (\%) $\uparrow$}            &
\textbf{Accuracy (\%) $\uparrow$}       &
\textbf{AUC (\%) $\uparrow$}            &
\textbf{Accuracy (\%) $\uparrow$}       &
\textbf{AUC (\%) $\uparrow$}            &
\textbf{Accuracy $\uparrow$}       \\
\midrule
\multicolumn{7}{c}{In-Domain (No, Mild, or Moderate DR, Clinical Labels \{0,1,2\})}\\
\midrule
\map (Deterministic)	& $82.0\pms{1.3}$ & $87.9\pms{0.5}$ & $83.1\pms{2.4}$ & $95.2\pms{0.4}$ & $88.4\pms{2.5}$ & $96.0\pms{0.3}$ \\
\textsc{deep ensemble}	& $85.1\pms{0.9}$ & $89.3\pms{0.3}$ & $82.0\pms{1.1}$ & $96.3\pms{0.3}$ & $85.3\pms{1.2}$ & $97.3\pms{0.2}$ \\
\mcd	& $89.2\pms{0.3}$ & $90.5\pms{0.1}$ & $92.8\pms{0.7}$ & $97.2\pms{0.0}$ & $95.4\pms{0.5}$ & $97.8\pms{0.0}$ \\
\mcd \textsc{ensemble}	& $\mathbf{90.6\pms{0.0}}$ & $\mathbf{91.4\pms{0.1}}$ & $\mathbf{93.1\pms{0.3}}$ & $\mathbf{97.8\pms{0.0}}$ & $\mathbf{95.7\pms{0.2}}$ & $\mathbf{98.2\pms{0.1}}$ \\
\fsvi	& $83.2\pms{0.4}$ & $89.5\pms{0.2}$ & $81.2\pms{1.1}$ & $95.6\pms{0.1}$ & $86.4\pms{0.9}$ & $96.4\pms{0.2}$ \\
\fsvi \textsc{ensemble}	& $86.2\pms{0.1}$ & $90.0\pms{0.1}$ & $81.2\pms{0.4}$ & $96.4\pms{0.0}$ & $86.1\pms{0.4}$ & $97.3\pms{0.0}$ \\
\textsc{radial}-\mfvi	& $76.9\pms{2.0}$ & $86.7\pms{0.5}$ & $69.0\pms{5.2}$ & $93.5\pms{0.6}$ & $70.1\pms{6.2}$ & $94.6\pms{0.6}$ \\
\textsc{radial}-\mfvi \textsc{ensemble}	& $81.3\pms{1.6}$ & $87.4\pms{0.4}$ & $66.3\pms{3.0}$ & $95.1\pms{0.5}$ & $66.2\pms{3.9}$ & $96.1\pms{0.5}$ \\
\textsc{rank}-1	& $81.6\pms{2.0}$ & $88.3\pms{0.7}$ & $79.4\pms{3.7}$ & $95.1\pms{0.6}$ & $82.9\pms{3.7}$ & $96.0\pms{0.5}$ \\
\textsc{rank}-1 \textsc{ensemble}	& $85.1\pms{1.4}$ & $89.3\pms{0.5}$ & $75.6\pms{1.3}$ & $96.1\pms{0.4}$ & $79.1\pms{1.7}$ & $96.9\pms{0.3}$ \\
\mfvi	& $81.3\pms{1.7}$ & $87.8\pms{0.7}$ & $79.5\pms{3.1}$ & $95.0\pms{0.5}$ & $82.6\pms{3.4}$ & $95.9\pms{0.4}$ \\
\mfvi \textsc{ensemble}	& $85.2\pms{0.8}$ & $89.4\pms{0.4}$ & $77.7\pms{1.1}$ & $96.1\pms{0.2}$ & $80.3\pms{1.3}$ & $96.8\pms{0.2}$ \\

\midrule
\multicolumn{7}{c}{Severity Shift (Severe or Proliferate DR, Clinical Labels \{3, 4\})}\\
\midrule

\map (Deterministic)	& $-$ & $74.4\pms{2.5}$ & $-$ & $93.2\pms{3.3}$ & $-$ & $98.6\pms{1.4}$ \\
\textsc{deep ensemble}	& $-$ & $74.5\pms{1.6}$ & $-$ & $89.8\pms{1.3}$ & $-$ & $97.0\pms{0.9}$ \\
\mcd	& $-$ & $86.4\pms{1.6}$ & $-$ & $\mathbf{99.5\pms{0.2}}$ & $-$ & $\mathbf{100.0\pms{0.0}}$ \\
\mcd \textsc{ensemble}	& $-$ & $\mathbf{87.4\pms{0.3}}$ & $-$ & $99.4\pms{0.1}$ & $-$ & $\mathbf{100.0\pms{0.0}}$ \\
\fsvi	& $-$ & $68.6\pms{1.2}$ & $-$ & $88.5\pms{1.3}$ & $-$ & $99.6\pms{0.3}$ \\
\fsvi \textsc{ensemble}	& $-$ & $69.3\pms{0.3}$ & $-$ & $86.3\pms{0.6}$ & $-$ & $99.4\pms{0.2}$ \\
\textsc{radial}-\mfvi	& $-$ & $52.0\pms{9.9}$ & $-$ & $59.3\pms{13.9}$ & $-$ & $63.9\pms{14.3}$ \\
\textsc{radial}-\mfvi \textsc{ensemble}	& $-$ & $54.4\pms{6.1}$ & $-$ & $58.0\pms{9.8}$ & $-$ & $60.6\pms{10.7}$ \\
\textsc{rank}-1	& $-$ & $67.5\pms{4.5}$ & $-$ & $82.6\pms{5.5}$ & $-$ & $92.7\pms{2.9}$ \\
\textsc{rank}-1 \textsc{ensemble}	& $-$ & $69.7\pms{2.4}$ & $-$ & $81.6\pms{2.1}$ & $-$ & $92.0\pms{1.7}$ \\
\mfvi	& $-$ & $71.5\pms{3.0}$ & $-$ & $86.7\pms{4.0}$ & $-$ & $94.1\pms{2.6}$ \\
\mfvi \textsc{ensemble}	& $-$ & $73.5\pms{1.6}$ & $-$ & $87.4\pms{0.9}$ & $-$ & $94.2\pms{0.8}$ \\

\midrule
\midrule
\end{tabular}
}
\label{tab:metrics_severity}
\end{table*}

\begin{table*}[!tb]
\centering
\caption{
    \textbf{OOD Detection Metrics.}
     We assess model uncertainty quantification across both shift tasks by using predictive entropy to detect out-of-distribution data.
}
\vspace{-3pt}
\resizebox{1.0\linewidth}{!}{%
\begin{tabular}{@{\extracolsep{2pt}}lccccccccccccccc@{}}
\midrule
& \multicolumn{2}{c}{Country Shift} & \multicolumn{2}{c}{Severity Shift} \\
\cline{2-3}
\cline{4-5}\\
\textbf{Method}     &
\textbf{AUROC (\%) $\uparrow$}         &
\textbf{AUPRC (\%) $\uparrow$}            &
\textbf{AUROC (\%) $\uparrow$}         &
\textbf{AUPRC (\%) $\uparrow$}      \\
\midrule

\map (Deterministic)  & $37.6\pms{1.7}$ & $5.2\pms{0.2}$ & $44.0\pms{3.5}$ & $9.3\pms{0.8}$ \\
\textsc{deep ensemble}  & $41.7\pms{1.2}$ & $5.6\pms{0.2}$ & $56.8\pms{1.2}$ & $12.4\pms{0.4}$ \\
\mcd  & $37.6\pms{0.9}$ & $5.1\pms{0.1}$ & $34.9\pms{1.4}$ & $7.1\pms{0.5}$ \\
\mcd \textsc{ensemble}  & $39.5\pms{0.3}$ & $5.3\pms{0.0}$ & $38.3\pms{1.2}$ & $7.7\pms{0.3}$ \\
\fsvi  & $42.2\pms{0.9}$ & $5.7\pms{0.1}$ & $49.0\pms{1.0}$ & $11.6\pms{0.4}$ \\
\fsvi \textsc{ensemble}  & $43.8\pms{0.6}$ & $5.9\pms{0.1}$ & $54.5\pms{0.5}$ & $14.5\pms{0.3}$ \\
\textsc{radial}-\mfvi  & $39.2\pms{2.7}$ & $5.3\pms{0.3}$ & $66.8\pms{6.2}$ & $19.9\pms{3.4}$ \\
\textsc{radial}-\mfvi \textsc{ensemble}  & $36.5\pms{0.8}$ & $4.9\pms{0.1}$ & $\mathbf{79.7\pms{3.4}}$ & $\mathbf{28.0\pms{2.9}}$ \\
\textsc{rank}-1  & $44.3\pms{2.4}$ & $6.0\pms{0.3}$ & $54.5\pms{4.4}$ & $12.8\pms{1.5}$ \\
\textsc{rank}-1 \textsc{ensemble}  & $48.9\pms{1.3}$ & $6.4\pms{0.2}$ & $65.6\pms{0.9}$ & $17.4\pms{0.7}$ \\
\mfvi  & $51.2\pms{0.8}$ & $6.7\pms{0.1}$ & $51.3\pms{3.6}$ & $10.4\pms{0.9}$ \\
\mfvi \textsc{ensemble}  & $\mathbf{52.4\pms{0.4}}$ & $\mathbf{6.9\pms{0.1}}$ & $60.4\pms{1.0}$ & $13.5\pms{0.6}$ \\

\midrule
\midrule
\end{tabular}
}
\label{tab:metrics_ood}
\end{table*}

\begin{table*}[b!]
\centering
\caption{
     \textbf{Standard Metrics, Country Shift.}
     We assess model predictive performance via standard metrics, and evaluate uncertainty quantification using expected calibration error on in-domain, shifted, and joint datasets (composed of the in-domain and shifted dataset, with no explicit balancing).
}
\vspace{-3pt}
\resizebox{1.0\linewidth}{!}{%
\begin{tabular}{@{\extracolsep{2pt}}lccccccccc@{}}
\midrule
\midrule
& \multicolumn{3}{c}{NLL $\downarrow$} & \multicolumn{3}{c}{Accuracy (\%) $\uparrow$} & \multicolumn{3}{c}{AUPRC (\%) $\uparrow$}  \\
\cline{2-4}
\cline{5-7}
\cline{8-10}\\
\textbf{Method}     &
\textbf{In-Domain}         &
\textbf{Shifted}            &
\textbf{Joint}       &
\textbf{In-Domain}         &
\textbf{Shifted}            &
\textbf{Joint}       &
\textbf{In-Domain}         &
\textbf{Shifted}            &
\textbf{Joint}         \\
\midrule
\midrule
\map (Deterministic)	& $1.27\pms{0.08}$ & $2.68\pms{0.18}$ & $1.36\pms{0.07}$ & $88.6\pms{0.7}$ & $86.2\pms{0.5}$ & $88.5\pms{0.6}$ & $75.2\pms{2.2}$ & $89.7\pms{0.3}$ & $77.2\pms{1.9}$ \\
\textsc{deep ensemble}	& $0.60\pms{0.00}$ & $1.60\pms{0.16}$ & $0.67\pms{0.01}$ & $90.3\pms{0.3}$ & $87.5\pms{0.1}$ & $90.1\pms{0.2}$ & $79.9\pms{0.5}$ & $\mathbf{91.1\pms{0.1}}$ & $81.0\pms{0.4}$ \\
\mcd	& $0.29\pms{0.00}$ & $1.07\pms{0.03}$ & $0.34\pms{0.00}$ & $90.9\pms{0.1}$ & $86.8\pms{0.2}$ & $90.6\pms{0.1}$ & $82.6\pms{0.2}$ & $88.8\pms{0.5}$ & $82.9\pms{0.2}$ \\
\mcd \textsc{ensemble}	& $\mathbf{0.25\pms{0.00}}$ & $0.92\pms{0.02}$ & $\mathbf{0.29\pms{0.00}}$ & $\mathbf{91.6\pms{0.0}}$ & $87.6\pms{0.1}$ & $\mathbf{91.4\pms{0.0}}$ & $\mathbf{84.4\pms{0.0}}$ & $88.3\pms{0.3}$ & $\mathbf{84.3\pms{0.1}}$ \\
\fsvi	& $0.35\pms{0.01}$ & $0.72\pms{0.05}$ & $0.38\pms{0.01}$ & $89.8\pms{0.0}$ & $87.6\pms{0.4}$ & $89.6\pms{0.0}$ & $77.7\pms{0.1}$ & $88.3\pms{0.5}$ & $78.9\pms{0.0}$ \\
\fsvi \textsc{ensemble}	& $0.28\pms{0.01}$ & $\mathbf{0.58\pms{0.01}}$ & $0.30\pms{0.01}$ & $90.6\pms{0.0}$ & $\mathbf{88.9\pms{0.1}}$ & $90.5\pms{0.0}$ & $80.7\pms{0.1}$ & $88.9\pms{0.2}$ & $81.3\pms{0.0}$ \\
\textsc{radial}-\mfvi	& $0.56\pms{0.07}$ & $0.70\pms{0.09}$ & $0.57\pms{0.07}$ & $74.2\pms{4.5}$ & $71.8\pms{4.2}$ & $74.1\pms{4.5}$ & $66.0\pms{0.9}$ & $84.8\pms{0.8}$ & $69.0\pms{0.8}$ \\
\textsc{radial}-\mfvi \textsc{ensemble}	& $0.55\pms{0.02}$ & $0.65\pms{0.03}$ & $0.56\pms{0.02}$ & $74.2\pms{1.4}$ & $69.0\pms{1.7}$ & $73.8\pms{1.4}$ & $68.9\pms{0.4}$ & $86.1\pms{0.1}$ & $71.6\pms{0.3}$ \\
\textsc{rank}-1	& $0.99\pms{0.07}$ & $1.85\pms{0.20}$ & $1.05\pms{0.05}$ & $87.7\pms{0.7}$ & $86.2\pms{0.5}$ & $87.6\pms{0.7}$ & $71.6\pms{2.5}$ & $88.8\pms{0.5}$ & $74.1\pms{2.1}$ \\
\textsc{rank}-1 \textsc{ensemble}	& $0.49\pms{0.04}$ & $0.96\pms{0.06}$ & $0.52\pms{0.03}$ & $89.3\pms{0.4}$ & $88.3\pms{0.1}$ & $89.2\pms{0.4}$ & $78.0\pms{1.3}$ & $89.6\pms{0.3}$ & $79.3\pms{1.1}$ \\
\mfvi	& $0.91\pms{0.02}$ & $1.26\pms{0.07}$ & $0.93\pms{0.02}$ & $85.7\pms{0.1}$ & $84.1\pms{0.3}$ & $85.6\pms{0.1}$ & $66.7\pms{0.3}$ & $85.9\pms{0.2}$ & $69.7\pms{0.3}$ \\
\mfvi \textsc{ensemble}	& $0.53\pms{0.00}$ & $0.72\pms{0.03}$ & $0.54\pms{0.00}$ & $87.8\pms{0.0}$ & $87.0\pms{0.2}$ & $87.7\pms{0.0}$ & $71.2\pms{0.1}$ & $87.4\pms{0.1}$ & $73.7\pms{0.1}$ \\

\midrule
& \multicolumn{3}{c}{AUROC (\%) $\uparrow$} & \multicolumn{3}{c}{ECE $\downarrow$} \\
\midrule

\map (Deterministic)	& $87.4\pms{1.2}$ & $92.2\pms{0.2}$ & $88.3\pms{1.1}$ & $0.10\pms{0.01}$ & $0.13\pms{0.00}$ & $0.10\pms{0.01}$ \\
\textsc{deep ensemble}	& $90.3\pms{0.2}$ & $94.2\pms{0.2}$ & $90.9\pms{0.2}$ & $0.06\pms{0.00}$ & $0.08\pms{0.00}$ & $0.06\pms{0.00}$ \\
\mcd	& $91.4\pms{0.1}$ & $94.0\pms{0.2}$ & $91.9\pms{0.1}$ & $0.03\pms{0.00}$ & $0.09\pms{0.00}$ & $0.03\pms{0.00}$ \\
\mcd \textsc{ensemble}	& $\mathbf{92.5\pms{0.0}}$ & $94.1\pms{0.1}$ & $\mathbf{92.9\pms{0.0}}$ & $\mathbf{0.02\pms{0.00}}$ & $0.09\pms{0.00}$ & $\mathbf{0.02\pms{0.00}}$ \\
\fsvi	& $88.5\pms{0.1}$ & $94.1\pms{0.1}$ & $89.4\pms{0.0}$ & $0.05\pms{0.01}$ & $0.08\pms{0.00}$ & $0.06\pms{0.01}$ \\
\fsvi \textsc{ensemble}	& $90.3\pms{0.1}$ & $\mathbf{94.6\pms{0.1}}$ & $90.9\pms{0.0}$ & $0.03\pms{0.00}$ & $0.07\pms{0.00}$ & $0.03\pms{0.00}$ \\
\textsc{radial}-\mfvi	& $83.2\pms{0.5}$ & $90.7\pms{0.6}$ & $84.3\pms{0.4}$ & $0.09\pms{0.03}$ & $0.14\pms{0.04}$ & $0.09\pms{0.03}$ \\
\textsc{radial}-\mfvi \textsc{ensemble}	& $84.9\pms{0.1}$ & $91.8\pms{0.1}$ & $85.9\pms{0.1}$ & $0.06\pms{0.01}$ & $0.10\pms{0.02}$ & $0.05\pms{0.01}$ \\
\textsc{rank}-1	& $85.6\pms{1.3}$ & $92.5\pms{0.2}$ & $86.7\pms{1.2}$ & $0.10\pms{0.01}$ & $0.11\pms{0.00}$ & $0.10\pms{0.01}$ \\
\textsc{rank}-1 \textsc{ensemble}	& $89.5\pms{0.8}$ & $94.1\pms{0.2}$ & $90.2\pms{0.7}$ & $0.05\pms{0.00}$ & $0.06\pms{0.00}$ & $0.05\pms{0.00}$ \\
\mfvi	& $83.3\pms{0.2}$ & $91.4\pms{0.2}$ & $84.6\pms{0.2}$ & $0.11\pms{0.00}$ & $0.13\pms{0.00}$ & $0.12\pms{0.00}$ \\
\mfvi \textsc{ensemble}	& $85.4\pms{0.0}$ & $93.2\pms{0.1}$ & $86.6\pms{0.0}$ & $0.06\pms{0.00}$ & $\mathbf{0.06\pms{0.00}}$ & $0.06\pms{0.00}$ \\
\midrule
\midrule
\end{tabular}
}
\label{tab:metrics_country_standard}
\end{table*}

\begin{table*}[b!]
\vspace{-24pt}
\centering
\caption{
     \textbf{Standard Metrics, Severity Shift.}
     We assess model predictive performance and expected calibration error on in-domain, shifted, and joint datasets (composed of the in-domain and shifted dataset, with no explicit balancing).
}
\vspace{-6pt}
\resizebox{1.0\linewidth}{0.15\paperheight}{%
\begin{tabular}{@{\extracolsep{2pt}}lccccccccc@{}}
\midrule
\midrule
& \multicolumn{3}{c}{NLL $\downarrow$} & \multicolumn{3}{c}{Accuracy (\%) $\uparrow$} & \multicolumn{3}{c}{AUPRC (\%) $\uparrow$}  \\
\cline{2-4}
\cline{5-7}
\cline{8-10}\\
\textbf{Method}     &
\textbf{In-Domain}         &
\textbf{Shifted}            &
\textbf{Joint}       &
\textbf{In-Domain}         &
\textbf{Shifted}            &
\textbf{Joint}       &
\textbf{In-Domain}         &
\textbf{Shifted}            &
\textbf{Joint}         \\
\midrule
\midrule
\map (Deterministic)	& $1.27\pms{0.07}$ & $2.27\pms{0.15}$ & $1.35\pms{0.08}$ & $87.9\pms{0.5}$ & $74.4\pms{2.3}$ & $86.8\pms{0.6}$ & $60.8\pms{2.4}$ & $-$ & $75.2\pms{1.7}$ \\
\textsc{deep ensemble}	& $0.62\pms{0.02}$ & $1.03\pms{0.06}$ & $0.65\pms{0.03}$ & $89.3\pms{0.3}$ & $74.5\pms{1.5}$ & $88.1\pms{0.4}$ & $65.6\pms{1.5}$ & $-$ & $79.2\pms{1.1}$ \\
\mcd	& $0.29\pms{0.00}$ & $0.33\pms{0.02}$ & $0.29\pms{0.00}$ & $90.5\pms{0.1}$ & $86.4\pms{1.5}$ & $90.1\pms{0.1}$ & $74.8\pms{0.6}$ & $-$ & $85.1\pms{0.3}$ \\
\mcd \textsc{ensemble}	& $\mathbf{0.25\pms{0.00}}$ & $\mathbf{0.28\pms{0.00}}$ & $\mathbf{0.25\pms{0.00}}$ & $\mathbf{91.4\pms{0.1}}$ & $\mathbf{87.4\pms{0.3}}$ & $\mathbf{91.1\pms{0.1}}$ & $\mathbf{77.0\pms{0.1}}$ & $-$ & $\mathbf{86.7\pms{0.1}}$ \\
\fsvi	& $0.36\pms{0.01}$ & $0.92\pms{0.05}$ & $0.41\pms{0.02}$ & $89.5\pms{0.1}$ & $68.6\pms{1.1}$ & $87.8\pms{0.2}$ & $64.7\pms{0.8}$ & $-$ & $77.6\pms{0.5}$ \\
\fsvi \textsc{ensemble}	& $0.31\pms{0.00}$ & $0.76\pms{0.01}$ & $0.34\pms{0.00}$ & $90.0\pms{0.1}$ & $69.3\pms{0.3}$ & $88.4\pms{0.1}$ & $70.0\pms{0.2}$ & $-$ & $81.6\pms{0.1}$ \\
\textsc{radial}-\mfvi	& $0.37\pms{0.01}$ & $0.76\pms{0.12}$ & $0.40\pms{0.02}$ & $86.7\pms{0.4}$ & $52.0\pms{9.0}$ & $83.9\pms{1.1}$ & $49.1\pms{3.5}$ & $-$ & $66.9\pms{2.9}$ \\
\textsc{radial}-\mfvi \textsc{ensemble}	& $0.35\pms{0.01}$ & $0.73\pms{0.07}$ & $0.38\pms{0.01}$ & $87.4\pms{0.4}$ & $54.4\pms{5.5}$ & $84.8\pms{0.8}$ & $56.2\pms{2.5}$ & $-$ & $73.5\pms{2.0}$ \\
\textsc{rank}-1	& $0.56\pms{0.06}$ & $1.14\pms{0.15}$ & $0.61\pms{0.07}$ & $88.3\pms{0.6}$ & $67.5\pms{4.1}$ & $86.6\pms{0.9}$ & $59.4\pms{3.7}$ & $-$ & $74.1\pms{2.5}$ \\
\textsc{rank}-1 \textsc{ensemble}	& $0.29\pms{0.01}$ & $0.60\pms{0.04}$ & $0.32\pms{0.01}$ & $89.3\pms{0.4}$ & $69.7\pms{2.2}$ & $87.7\pms{0.5}$ & $66.5\pms{2.5}$ & $-$ & $80.0\pms{1.6}$ \\
\mfvi	& $0.66\pms{0.11}$ & $1.26\pms{0.21}$ & $0.71\pms{0.11}$ & $87.8\pms{0.7}$ & $71.5\pms{2.7}$ & $86.5\pms{0.8}$ & $59.0\pms{3.2}$ & $-$ & $73.7\pms{2.3}$ \\
\mfvi \textsc{ensemble}	& $0.29\pms{0.01}$ & $0.55\pms{0.02}$ & $0.31\pms{0.01}$ & $89.4\pms{0.4}$ & $73.5\pms{1.4}$ & $88.2\pms{0.4}$ & $66.4\pms{1.6}$ & $-$ & $79.7\pms{1.1}$ \\

\midrule
& \multicolumn{3}{c}{AUROC (\%) $\uparrow$} & \multicolumn{3}{c}{ECE $\downarrow$} \\
\midrule

\map (Deterministic)	& $82.0\pms{1.2}$ & $-$ & $86.3\pms{1.0}$ & $0.11\pms{0.00}$ & $0.23\pms{0.02}$ & $0.12\pms{0.01}$ \\
\textsc{deep ensemble}	& $85.1\pms{0.8}$ & $-$ & $88.9\pms{0.6}$ & $0.06\pms{0.00}$ & $0.15\pms{0.01}$ & $0.07\pms{0.00}$ \\
\mcd	& $89.2\pms{0.3}$ & $-$ & $92.0\pms{0.2}$ & $0.02\pms{0.00}$ & $0.06\pms{0.01}$ & $0.02\pms{0.00}$ \\
\mcd \textsc{ensemble}	& $\mathbf{90.6\pms{0.0}}$ & $-$ & $\mathbf{93.1\pms{0.0}}$ & $\mathbf{0.01\pms{0.00}}$ & $\mathbf{0.03\pms{0.00}}$ & $\mathbf{0.01\pms{0.00}}$ \\
\fsvi	& $83.2\pms{0.4}$ & $-$ & $86.9\pms{0.3}$ & $0.06\pms{0.00}$ & $0.23\pms{0.01}$ & $0.07\pms{0.00}$ \\
\fsvi \textsc{ensemble}	& $86.2\pms{0.1}$ & $-$ & $89.4\pms{0.0}$ & $0.04\pms{0.00}$ & $0.19\pms{0.00}$ & $0.06\pms{0.00}$ \\
\textsc{radial}-\mfvi	& $76.9\pms{1.8}$ & $-$ & $82.2\pms{1.6}$ & $0.05\pms{0.01}$ & $0.23\pms{0.07}$ & $0.04\pms{0.01}$ \\
\textsc{radial}-\mfvi \textsc{ensemble}	& $81.3\pms{1.4}$ & $-$ & $86.2\pms{1.2}$ & $0.07\pms{0.01}$ & $0.15\pms{0.04}$ & $0.06\pms{0.01}$ \\
\textsc{rank}-1	& $81.6\pms{1.8}$ & $-$ & $85.8\pms{1.4}$ & $0.06\pms{0.01}$ & $0.22\pms{0.03}$ & $0.07\pms{0.02}$ \\
\textsc{rank}-1 \textsc{ensemble}	& $85.1\pms{1.3}$ & $-$ & $89.1\pms{0.9}$ & $0.02\pms{0.00}$ & $0.12\pms{0.02}$ & $0.03\pms{0.00}$ \\
\mfvi	& $81.3\pms{1.6}$ & $-$ & $85.4\pms{1.3}$ & $0.07\pms{0.01}$ & $0.19\pms{0.03}$ & $0.08\pms{0.02}$ \\
\mfvi \textsc{ensemble}	& $85.2\pms{0.7}$ & $-$ & $88.9\pms{0.6}$ & $0.02\pms{0.00}$ & $0.10\pms{0.01}$ & $0.02\pms{0.00}$ \\
\midrule
\midrule
\end{tabular}
}
\label{tab:metrics_severity_standard}
\end{table*}

\begin{table*}[!tb]
\centering
\caption{
     \textbf{Expert Referral Metrics, Country Shift.} 
     We assess model predictive performance and uncertainty quantification in the context of expert referral.
     We construct referral curves on a variety of metrics---AUC, Accuracy, NLL and AUPRC---by sweeping over the referral thresholds $\tau$, obtaining a point for each possible partition of the dataset into ``referred'' and ``non-referred''. 
     We report the area under the referral curve for metric $X$ as ``Area Under SP-$X$ Curve''.
     All methods are tuned according to the area under the ROC curve on the in-domain dataset.
     The Balanced evaluation dataset is constructed with the procedure described in \Cref{sec:joint_tuning}.
}
\vspace{-6pt}
\resizebox{1.0\linewidth}{0.15\paperheight}{%
\begin{tabular}{@{\extracolsep{2pt}}lcccccccc@{}}
\midrule
\midrule
& \multicolumn{4}{c}{Area Under SP-AUC Curve $\uparrow$} & \multicolumn{4}{c}{Area Under SP-Accuracy Curve $\uparrow$} \\
\cline{2-5}
\cline{6-9}\\
\textbf{Method}     &
\textbf{In-Domain}         &
\textbf{Shifted}            &
\textbf{Joint}       &
\textbf{Balanced}         &
\textbf{In-Domain}         &
\textbf{Shifted}            &
\textbf{Joint}       &
\textbf{Balanced}       \\
\midrule
\midrule
\map (Deterministic)	& $89.2\pms{0.9}$ & $74.7\pms{1.7}$ & $88.8\pms{0.5}$ & $88.8\pms{0.5}$ & $94.9\pms{0.4}$ & $87.3\pms{0.8}$ & $94.0\pms{0.3}$ & $89.9\pms{0.5}$ \\
\textsc{deep ensemble}	& $91.7\pms{0.2}$ & $80.7\pms{1.5}$ & $91.8\pms{0.1}$ & $91.8\pms{0.1}$ & $96.5\pms{0.1}$ & $91.0\pms{0.7}$ & $95.9\pms{0.1}$ & $92.9\pms{0.5}$ \\
\mcd	& $94.7\pms{0.2}$ & $79.7\pms{0.3}$ & $93.9\pms{0.2}$ & $93.9\pms{0.2}$ & $96.8\pms{0.0}$ & $88.9\pms{0.2}$ & $95.9\pms{0.0}$ & $91.7\pms{0.1}$ \\
\mcd \textsc{ensemble}	& $\mathbf{95.4\pms{0.1}}$ & $79.4\pms{0.1}$ & $\mathbf{94.4\pms{0.1}}$ & $\mathbf{94.4\pms{0.1}}$ & $\mathbf{97.3\pms{0.0}}$ & $89.0\pms{0.2}$ & $\mathbf{96.4\pms{0.0}}$ & $92.0\pms{0.1}$ \\
\fsvi	& $91.6\pms{0.2}$ & $83.7\pms{1.0}$ & $92.0\pms{0.1}$ & $92.0\pms{0.1}$ & $95.9\pms{0.0}$ & $90.6\pms{0.2}$ & $95.3\pms{0.1}$ & $92.5\pms{0.2}$ \\
\fsvi \textsc{ensemble}	& $92.6\pms{0.1}$ & $83.2\pms{0.4}$ & $92.9\pms{0.1}$ & $92.9\pms{0.1}$ & $96.6\pms{0.0}$ & $90.9\pms{0.1}$ & $95.9\pms{0.0}$ & $93.0\pms{0.1}$ \\
\textsc{radial}-\mfvi	& $87.9\pms{1.0}$ & $77.7\pms{1.3}$ & $88.0\pms{1.1}$ & $88.0\pms{1.1}$ & $82.4\pms{5.0}$ & $81.9\pms{2.7}$ & $82.3\pms{4.8}$ & $81.8\pms{3.6}$ \\
\textsc{radial}-\mfvi \textsc{ensemble}	& $89.5\pms{0.3}$ & $76.2\pms{0.3}$ & $89.1\pms{0.3}$ & $89.1\pms{0.3}$ & $83.4\pms{1.4}$ & $80.8\pms{0.9}$ & $83.0\pms{1.4}$ & $81.2\pms{1.2}$ \\
\textsc{rank}-1	& $87.8\pms{1.3}$ & $81.2\pms{2.2}$ & $88.4\pms{0.9}$ & $88.4\pms{0.9}$ & $94.6\pms{0.5}$ & $89.7\pms{0.7}$ & $94.0\pms{0.3}$ & $91.4\pms{0.3}$ \\
\textsc{rank}-1 \textsc{ensemble}	& $90.3\pms{0.9}$ & $88.3\pms{1.0}$ & $91.5\pms{0.7}$ & $91.5\pms{0.7}$ & $96.2\pms{0.3}$ & $\mathbf{92.8\pms{0.2}}$ & $95.9\pms{0.3}$ & $\mathbf{94.1\pms{0.2}}$ \\
\mfvi	& $86.9\pms{0.4}$ & $88.7\pms{0.8}$ & $88.2\pms{0.3}$ & $88.2\pms{0.3}$ & $93.7\pms{0.1}$ & $90.5\pms{0.4}$ & $93.5\pms{0.1}$ & $91.9\pms{0.2}$ \\
\mfvi \textsc{ensemble}	& $88.1\pms{0.3}$ & $\mathbf{92.0\pms{0.4}}$ & $89.7\pms{0.2}$ & $89.7\pms{0.2}$ & $94.8\pms{0.0}$ & $92.4\pms{0.2}$ & $94.6\pms{0.0}$ & $93.5\pms{0.1}$ \\

\midrule
& \multicolumn{4}{c}{Area Under SP-NLL Curve $\downarrow$} & \multicolumn{4}{c}{Area Under SP-AUPRC Curve $\uparrow$} \\
\midrule

\map (Deterministic)    & $1.22\pms{0.09}$ & $4.10\pms{0.33}$ & $1.54\pms{0.04}$ & $3.09\pms{0.22}$ & $86.4\pms{2.2}$ & $92.3\pms{0.3}$ & $87.7\pms{1.6}$ & $87.7\pms{1.6}$ \\
\textsc{deep ensemble}  & $0.54\pms{0.01}$ & $2.61\pms{0.30}$ & $0.79\pms{0.04}$ & $1.87\pms{0.21}$ & $87.6\pms{0.8}$ & $\mathbf{94.0\pms{0.3}}$ & $89.1\pms{0.5}$ & $89.1\pms{0.5}$ \\
\mcd    & $0.19\pms{0.01}$ & $1.87\pms{0.08}$ & $0.36\pms{0.01}$ & $1.22\pms{0.04}$ & $92.1\pms{0.3}$ & $91.6\pms{0.6}$ & $91.5\pms{0.2}$ & $91.5\pms{0.2}$ \\
\mcd \textsc{ensemble}  & $\mathbf{0.14\pms{0.00}}$ & $1.61\pms{0.05}$ & $0.29\pms{0.01}$ & $1.04\pms{0.03}$ & $\mathbf{92.8\pms{0.1}}$ & $91.0\pms{0.4}$ & $\mathbf{91.9\pms{0.1}}$ & $\mathbf{91.9\pms{0.1}}$ \\
\fsvi   & $0.24\pms{0.01}$ & $1.13\pms{0.12}$ & $0.33\pms{0.02}$ & $0.79\pms{0.09}$ & $86.7\pms{0.4}$ & $91.0\pms{0.6}$ & $87.6\pms{0.3}$ & $87.6\pms{0.3}$ \\
\fsvi \textsc{ensemble} & $0.17\pms{0.01}$ & $0.90\pms{0.06}$ & $\mathbf{0.24\pms{0.01}}$ & $0.60\pms{0.03}$ & $87.8\pms{0.2}$ & $91.4\pms{0.3}$ & $88.6\pms{0.2}$ & $88.6\pms{0.2}$ \\
\textsc{radial}-\mfvi   & $0.50\pms{0.11}$ & $0.68\pms{0.11}$ & $0.51\pms{0.11}$ & $0.61\pms{0.11}$ & $80.6\pms{1.1}$ & $88.7\pms{0.7}$ & $82.2\pms{1.0}$ & $82.2\pms{1.0}$ \\
\textsc{radial}-\mfvi \textsc{ensemble} & $0.44\pms{0.02}$ & $\mathbf{0.59\pms{0.05}}$ & $0.46\pms{0.03}$ & $\mathbf{0.54\pms{0.04}}$ & $83.0\pms{0.4}$ & $89.3\pms{0.1}$ & $84.1\pms{0.2}$ & $84.1\pms{0.2}$ \\
\textsc{rank}-1 & $0.93\pms{0.08}$ & $2.92\pms{0.35}$ & $1.16\pms{0.03}$ & $2.22\pms{0.22}$ & $81.3\pms{3.1}$ & $92.7\pms{0.3}$ & $83.8\pms{2.4}$ & $83.8\pms{2.4}$ \\
Rank1 Ensemble  & $0.41\pms{0.05}$ & $1.58\pms{0.11}$ & $0.54\pms{0.04}$ & $1.15\pms{0.07}$ & $82.7\pms{1.8}$ & $93.5\pms{0.2}$ & $85.3\pms{1.4}$ & $85.3\pms{1.4}$ \\
\mfvi   & $0.79\pms{0.02}$ & $1.92\pms{0.13}$ & $0.89\pms{0.03}$ & $1.44\pms{0.08}$ & $77.9\pms{0.9}$ & $91.1\pms{0.1}$ & $80.6\pms{0.7}$ & $80.6\pms{0.7}$ \\
\mfvi \textsc{ensemble} & $0.47\pms{0.01}$ & $1.22\pms{0.05}$ & $0.53\pms{0.01}$ & $0.89\pms{0.03}$ & $79.3\pms{0.6}$ & $91.1\pms{0.1}$ & $82.0\pms{0.4}$ & $82.0\pms{0.4}$ \\

\midrule
\midrule
\end{tabular}
}
\label{tab:metrics_country_2}
\end{table*}

\begin{table*}[!tb]
\vspace{-24pt}
\centering
\caption{
     \textbf{Expert Referral Metrics, Severity Shift.} 
     We assess model predictive performance and uncertainty quantification in the context of expert referral.
     We construct referral curves on a variety of metrics---AUC, Accuracy, NLL and AUPRC---by sweeping over the referral thresholds $\tau$, obtaining a point for each possible partition of the dataset into ``referred" and ``non-referred". 
     We report the area under the referral curve for metric $X$ as ``Area Under SP-$X$ Curve''.
     All methods are tuned according to the area under the ROC curve on the in-domain dataset.
     The Balanced evaluation dataset is constructed with the procedure described in \Cref{sec:joint_tuning}.
}
\vspace{-6pt}
\resizebox{1.0\linewidth}{0.15\paperheight}{%
\begin{tabular}{@{\extracolsep{2pt}}lcccccccc@{}}
\midrule
\midrule
& \multicolumn{4}{c}{Area Under SP-AUC Curve $\uparrow$} & \multicolumn{4}{c}{Area Under SP-Accuracy Curve $\uparrow$} \\
\cline{2-5}
\cline{6-9}\\
\textbf{Method}     &
\textbf{In-Domain}         &
\textbf{Shifted}            &
\textbf{Joint}       &
\textbf{Balanced}         &
\textbf{In-Domain}         &
\textbf{Shifted}            &
\textbf{Joint}       &
\textbf{Balanced}       \\
\midrule
\midrule

\map (Deterministic)	& $84.9\pms{1.6}$ & $-$ & $88.3\pms{0.8}$ & $88.3\pms{0.8}$ & $94.2\pms{0.4}$ & $90.4\pms{1.9}$ & $94.2\pms{0.4}$ & $93.1\pms{1.0}$ \\
\textsc{deep ensemble}	& $85.1\pms{1.0}$ & $-$ & $90.2\pms{0.6}$ & $90.2\pms{0.6}$ & $95.7\pms{0.3}$ & $89.3\pms{1.0}$ & $95.5\pms{0.3}$ & $93.4\pms{0.5}$ \\
\mcd	& $93.2\pms{0.6}$ & $-$ & $95.2\pms{0.3}$ & $95.2\pms{0.3}$ & $96.5\pms{0.0}$ & $97.1\pms{0.6}$ & $96.7\pms{0.0}$ & $97.0\pms{0.3}$ \\
\mcd \textsc{ensemble}	& $\mathbf{93.6\pms{0.2}}$ & $-$ & $\mathbf{95.7\pms{0.1}}$ & $\mathbf{95.7\pms{0.1}}$ & $\mathbf{97.1\pms{0.1}}$ & $\mathbf{97.3\pms{0.2}}$ & $\mathbf{97.2\pms{0.0}}$ & $\mathbf{97.4\pms{0.0}}$ \\
\fsvi	& $84.4\pms{0.8}$ & $-$ & $89.5\pms{0.5}$ & $89.5\pms{0.5}$ & $95.2\pms{0.1}$ & $87.2\pms{0.9}$ & $94.8\pms{0.2}$ & $92.0\pms{0.4}$ \\
\fsvi \textsc{ensemble}	& $84.8\pms{0.3}$ & $-$ & $90.4\pms{0.1}$ & $90.4\pms{0.1}$ & $96.0\pms{0.0}$ & $86.5\pms{0.3}$ & $95.6\pms{0.0}$ & $92.5\pms{0.1}$ \\
\textsc{radial}-\mfvi	& $72.3\pms{4.8}$ & $-$ & $78.2\pms{4.8}$ & $78.2\pms{4.8}$ & $92.9\pms{0.6}$ & $61.7\pms{12.7}$ & $92.0\pms{0.9}$ & $83.8\pms{3.9}$ \\
\textsc{radial}-\mfvi \textsc{ensemble}	& $70.6\pms{3.2}$ & $-$ & $76.8\pms{3.6}$ & $76.8\pms{3.6}$ & $94.4\pms{0.5}$ & $60.3\pms{8.9}$ & $93.5\pms{0.6}$ & $85.2\pms{2.5}$ \\
\textsc{rank}-1	& $82.3\pms{2.7}$ & $-$ & $87.4\pms{1.5}$ & $87.4\pms{1.5}$ & $94.5\pms{0.5}$ & $83.9\pms{3.7}$ & $94.1\pms{0.6}$ & $90.8\pms{1.6}$ \\
\textsc{rank}-1 \textsc{ensemble}	& $80.7\pms{1.2}$ & $-$ & $88.1\pms{1.0}$ & $88.1\pms{1.0}$ & $95.6\pms{0.4}$ & $84.5\pms{1.7}$ & $95.3\pms{0.4}$ & $92.0\pms{0.8}$ \\
\mfvi	& $82.2\pms{2.5}$ & $-$ & $87.4\pms{1.4}$ & $87.4\pms{1.4}$ & $94.3\pms{0.5}$ & $86.5\pms{2.8}$ & $93.9\pms{0.6}$ & $91.2\pms{1.4}$ \\
\mfvi \textsc{ensemble}	& $81.7\pms{1.0}$ & $-$ & $88.9\pms{0.7}$ & $88.9\pms{0.7}$ & $95.6\pms{0.2}$ & $87.6\pms{0.9}$ & $95.2\pms{0.2}$ & $92.6\pms{0.4}$ \\

\midrule
& \multicolumn{4}{c}{Area Under SP-NLL Curve $\downarrow$} & \multicolumn{4}{c}{Area Under SP-AUPRC Curve $\uparrow$} \\
\midrule

\map (Deterministic)    & $1.26\pms{0.11}$ & $1.23\pms{0.18}$ & $1.19\pms{0.09}$ & $1.10\pms{0.11}$ & $73.2\pms{3.5}$ & $-$ & $85.2\pms{2.2}$ & $85.2\pms{2.2}$ \\
\textsc{deep ensemble}  & $0.57\pms{0.03}$ & $0.76\pms{0.07}$ & $0.56\pms{0.03}$ & $0.60\pms{0.05}$ & $70.2\pms{1.8}$ & $-$ & $84.5\pms{1.2}$ & $84.5\pms{1.2}$ \\
\mcd    & $0.19\pms{0.01}$ & $0.10\pms{0.01}$ & $0.17\pms{0.01}$ & $0.12\pms{0.00}$ & $86.8\pms{1.2}$ & $-$ & $93.5\pms{0.5}$ & $93.5\pms{0.5}$ \\
\mcd \textsc{ensemble}  & $\mathbf{0.14\pms{0.01}}$ & $\mathbf{0.08\pms{0.00}}$ & $\mathbf{0.13\pms{0.01}}$ & $\mathbf{0.10\pms{0.00}}$ & $\mathbf{87.4\pms{0.4}}$ & $-$ & $\mathbf{94.0\pms{0.2}}$ & $\mathbf{94.0\pms{0.2}}$ \\
\fsvi   & $0.26\pms{0.01}$ & $0.50\pms{0.05}$ & $0.26\pms{0.01}$ & $0.35\pms{0.02}$ & $70.8\pms{1.7}$ & $-$ & $84.2\pms{0.9}$ & $84.2\pms{0.9}$ \\
\fsvi \textsc{ensemble} & $0.19\pms{0.00}$ & $0.44\pms{0.01}$ & $0.20\pms{0.00}$ & $0.28\pms{0.00}$ & $69.6\pms{0.9}$ & $-$ & $84.4\pms{0.4}$ & $84.4\pms{0.4}$ \\
\textsc{radial}-\mfvi   & $0.26\pms{0.02}$ & $0.70\pms{0.21}$ & $0.27\pms{0.03}$ & $0.38\pms{0.08}$ & $43.5\pms{9.8}$ & $-$ & $59.2\pms{9.7}$ & $59.2\pms{9.7}$ \\
\textsc{radial}-\mfvi \textsc{ensemble} & $0.24\pms{0.01}$ & $0.72\pms{0.13}$ & $0.25\pms{0.01}$ & $0.37\pms{0.04}$ & $33.7\pms{7.0}$ & $-$ & $50.8\pms{8.0}$ & $50.8\pms{8.0}$ \\
\textsc{rank}-1 & $0.49\pms{0.09}$ & $0.77\pms{0.20}$ & $0.48\pms{0.08}$ & $0.56\pms{0.11}$ & $65.9\pms{5.9}$ & $-$ & $80.2\pms{3.6}$ & $80.2\pms{3.6}$ \\
Rank1 Ensemble  & $0.18\pms{0.01}$ & $0.39\pms{0.04}$ & $0.18\pms{0.01}$ & $0.24\pms{0.02}$ & $60.9\pms{2.5}$ & $-$ & $79.0\pms{1.8}$ & $79.0\pms{1.8}$ \\
\mfvi   & $0.60\pms{0.14}$ & $0.79\pms{0.17}$ & $0.58\pms{0.13}$ & $0.62\pms{0.12}$ & $66.6\pms{5.2}$ & $-$ & $81.1\pms{3.2}$ & $81.1\pms{3.2}$ \\
\mfvi \textsc{ensemble} & $0.18\pms{0.01}$ & $0.35\pms{0.03}$ & $0.19\pms{0.01}$ & $0.24\pms{0.02}$ & $63.7\pms{1.8}$ & $-$ & $81.1\pms{1.1}$ & $81.1\pms{1.1}$ \\

\midrule
\midrule
\end{tabular}
}
\label{tab:metrics_severity_2}
\end{table*}

\begin{table*}[b!]
\centering
\caption{
    \textbf{Standard Metrics, Country Shift, Tuned on Joint Dataset.}
    Here all methods are tuned according to the joint validation metric (\Cref{sec:joint_tuning}): area under the selective prediction accuracy curve constructed on the balanced joint validation dataset (composed of the in-domain and upsampled shifted validation datasets).
    Ensembles have $K = 3$ models.
    We assess model predictive performance and expected calibration error on in-domain, shifted, and joint (union of in-domain and shifted, without explicit balancing) evaluation datasets.
}
\vspace{-6pt}
\resizebox{1.0\linewidth}{0.15\paperheight}{%
\begin{tabular}{@{\extracolsep{2pt}}lccccccccc@{}}
\midrule
\midrule
& \multicolumn{3}{c}{NLL $\downarrow$} & \multicolumn{3}{c}{Accuracy (\%) $\uparrow$} & \multicolumn{3}{c}{AUPRC (\%) $\uparrow$}  \\
\cline{2-4}
\cline{5-7}
\cline{8-10}\\
\textbf{Method}     &
\textbf{In-Domain}         &
\textbf{Shifted}            &
\textbf{Joint}       &
\textbf{In-Domain}         &
\textbf{Shifted}            &
\textbf{Joint}       &
\textbf{In-Domain}         &
\textbf{Shifted}            &
\textbf{Joint}         \\
\midrule
\midrule
\map (Deterministic)	& $1.02\pms{0.09}$ & $2.41\pms{0.18}$ & $1.11\pms{0.07}$ & $89.3\pms{0.3}$ & $87.0\pms{0.3}$ & $89.2\pms{0.3}$ & $77.5\pms{1.2}$ & $90.5\pms{0.2}$ & $79.2\pms{1.0}$ \\
\textsc{deep ensemble}	& $0.54\pms{0.01}$ & $1.65\pms{0.17}$ & $0.61\pms{0.01}$ & $90.8\pms{0.0}$ & $88.3\pms{0.2}$ & $90.7\pms{0.0}$ & $81.1\pms{0.2}$ & $\mathbf{91.3\pms{0.2}}$ & $82.0\pms{0.2}$ \\
\mcd	& $0.31\pms{0.01}$ & $0.77\pms{0.08}$ & $0.34\pms{0.01}$ & $90.0\pms{0.2}$ & $87.6\pms{0.4}$ & $89.9\pms{0.2}$ & $81.1\pms{0.4}$ & $87.7\pms{0.5}$ & $82.0\pms{0.3}$ \\
\mcd \textsc{ensemble}	& $\mathbf{0.25\pms{0.00}}$ & $0.58\pms{0.04}$ & $\mathbf{0.28\pms{0.00}}$ & $\mathbf{91.2\pms{0.0}}$ & $88.3\pms{0.2}$ & $\mathbf{91.0\pms{0.0}}$ & $\mathbf{83.3\pms{0.1}}$ & $87.7\pms{0.4}$ & $\mathbf{83.7\pms{0.1}}$ \\
\fsvi	& $0.52\pms{0.05}$ & $0.67\pms{0.06}$ & $0.53\pms{0.05}$ & $88.7\pms{0.5}$ & $88.2\pms{0.4}$ & $88.7\pms{0.4}$ & $75.8\pms{0.8}$ & $88.1\pms{1.0}$ & $77.5\pms{0.6}$ \\
\fsvi \textsc{ensemble}	& $0.39\pms{0.02}$ & $0.42\pms{0.02}$ & $0.39\pms{0.02}$ & $89.2\pms{0.3}$ & $\mathbf{89.7\pms{0.1}}$ & $89.3\pms{0.3}$ & $79.5\pms{0.4}$ & $88.9\pms{0.5}$ & $80.5\pms{0.3}$ \\
\textsc{radial}-\mfvi	& $0.60\pms{0.10}$ & $0.72\pms{0.20}$ & $0.61\pms{0.11}$ & $85.9\pms{0.3}$ & $85.4\pms{0.6}$ & $85.9\pms{0.3}$ & $66.2\pms{0.8}$ & $87.9\pms{0.5}$ & $69.6\pms{0.6}$ \\
\textsc{radial}-\mfvi \textsc{ensemble}	& $0.38\pms{0.00}$ & $\mathbf{0.34\pms{0.02}}$ & $0.38\pms{0.00}$ & $87.2\pms{0.2}$ & $87.8\pms{0.1}$ & $87.2\pms{0.2}$ & $69.7\pms{0.4}$ & $89.3\pms{0.2}$ & $72.6\pms{0.3}$ \\
\textsc{rank}-1	& $0.88\pms{0.08}$ & $1.95\pms{0.27}$ & $0.95\pms{0.09}$ & $87.0\pms{0.8}$ & $85.1\pms{0.5}$ & $86.9\pms{0.7}$ & $71.3\pms{2.4}$ & $88.4\pms{0.3}$ & $73.8\pms{2.0}$ \\
\textsc{rank}-1 \textsc{ensemble}	& $0.40\pms{0.02}$ & $1.02\pms{0.11}$ & $0.44\pms{0.02}$ & $89.1\pms{0.4}$ & $87.1\pms{0.3}$ & $89.0\pms{0.4}$ & $77.2\pms{1.2}$ & $89.4\pms{0.1}$ & $78.7\pms{1.0}$ \\
\mfvi	& $1.09\pms{0.13}$ & $1.69\pms{0.26}$ & $1.12\pms{0.14}$ & $85.9\pms{0.5}$ & $84.5\pms{0.7}$ & $85.8\pms{0.5}$ & $67.1\pms{1.9}$ & $87.6\pms{0.9}$ & $70.3\pms{1.5}$ \\
\mfvi \textsc{ensemble}	& $0.46\pms{0.04}$ & $0.71\pms{0.16}$ & $0.48\pms{0.05}$ & $88.4\pms{0.2}$ & $86.8\pms{0.3}$ & $88.3\pms{0.1}$ & $73.5\pms{0.9}$ & $89.6\pms{0.6}$ & $75.7\pms{0.7}$ \\

\midrule
& \multicolumn{3}{c}{AUROC (\%) $\uparrow$} & \multicolumn{3}{c}{ECE $\downarrow$} \\
\midrule

\map (Deterministic)	& $88.6\pms{0.6}$ & $93.2\pms{0.2}$ & $89.5\pms{0.5}$ & $0.09\pms{0.00}$ & $0.12\pms{0.00}$ & $0.09\pms{0.00}$ \\
\textsc{deep ensemble}	& $90.6\pms{0.0}$ & $94.5\pms{0.2}$ & $91.3\pms{0.0}$ & $0.05\pms{0.00}$ & $0.09\pms{0.00}$ & $0.05\pms{0.00}$ \\
\mcd	& $90.7\pms{0.2}$ & $93.9\pms{0.2}$ & $91.4\pms{0.2}$ & $0.03\pms{0.00}$ & $0.08\pms{0.00}$ & $0.04\pms{0.00}$ \\
\mcd \textsc{ensemble}	& $\mathbf{91.9\pms{0.1}}$ & $94.2\pms{0.2}$ & $\mathbf{92.5\pms{0.0}}$ & $\mathbf{0.02\pms{0.00}}$ & $0.06\pms{0.00}$ & $\mathbf{0.02\pms{0.00}}$ \\
\fsvi	& $87.4\pms{0.4}$ & $94.0\pms{0.4}$ & $88.5\pms{0.4}$ & $0.08\pms{0.01}$ & $0.08\pms{0.00}$ & $0.08\pms{0.01}$ \\
\fsvi \textsc{ensemble}	& $89.6\pms{0.2}$ & $\mathbf{94.6\pms{0.2}}$ & $90.4\pms{0.2}$ & $0.06\pms{0.01}$ & $0.05\pms{0.00}$ & $0.06\pms{0.01}$ \\
\textsc{radial}-\mfvi	& $83.0\pms{0.4}$ & $92.7\pms{0.4}$ & $84.3\pms{0.3}$ & $0.09\pms{0.01}$ & $0.07\pms{0.02}$ & $0.09\pms{0.01}$ \\
\textsc{radial}-\mfvi \textsc{ensemble}	& $84.8\pms{0.2}$ & $94.1\pms{0.1}$ & $86.0\pms{0.2}$ & $0.05\pms{0.00}$ & $\mathbf{0.03\pms{0.01}}$ & $0.05\pms{0.00}$ \\
\textsc{rank}-1	& $85.4\pms{1.3}$ & $92.0\pms{0.3}$ & $86.5\pms{1.2}$ & $0.10\pms{0.01}$ & $0.12\pms{0.01}$ & $0.10\pms{0.01}$ \\
\textsc{rank}-1 \textsc{ensemble}	& $89.0\pms{0.8}$ & $94.0\pms{0.2}$ & $89.8\pms{0.7}$ & $0.05\pms{0.00}$ & $0.07\pms{0.00}$ & $0.05\pms{0.00}$ \\
\mfvi	& $83.4\pms{0.9}$ & $91.7\pms{0.6}$ & $84.7\pms{0.8}$ & $0.11\pms{0.01}$ & $0.12\pms{0.02}$ & $0.11\pms{0.01}$ \\
\mfvi \textsc{ensemble}	& $86.8\pms{0.5}$ & $94.0\pms{0.3}$ & $87.9\pms{0.5}$ & $0.05\pms{0.00}$ & $0.06\pms{0.01}$ & $0.05\pms{0.00}$ \\
\midrule
\midrule
\end{tabular}
}
\label{tab:metrics_country_joint_standard}
\end{table*}

\begin{table*}[!tb]
\vspace{-24pt}
\centering
\caption{
    \textbf{Standard Metrics, Severity Shift, Tuned on Joint Dataset.}
    Here all methods are tuned according to the joint validation metric (\Cref{sec:joint_tuning}): area under the selective prediction accuracy curve constructed on the balanced joint validation dataset (composed of the in-domain and upsampled shifted validation datasets).
    Ensembles have $K = 3$ models.
    We assess model predictive performance and expected calibration error on in-domain, shifted, and joint (union of in-domain and shifted, without explicit balancing) evaluation datasets.
    }
\vspace{-6pt}
\resizebox{1.0\linewidth}{0.15\paperheight}{%
\begin{tabular}{@{\extracolsep{2pt}}lccccccccc@{}}
\midrule
\midrule
& \multicolumn{3}{c}{NLL $\downarrow$} & \multicolumn{3}{c}{Accuracy (\%) $\uparrow$} & \multicolumn{3}{c}{AUPRC (\%) $\uparrow$}  \\
\cline{2-4}
\cline{5-7}
\cline{8-10}\\
\textbf{Method}     &
\textbf{In-Domain}         &
\textbf{Shifted}            &
\textbf{Joint}       &
\textbf{In-Domain}         &
\textbf{Shifted}            &
\textbf{Joint}       &
\textbf{In-Domain}         &
\textbf{Shifted}            &
\textbf{Joint}         \\
\midrule
\midrule
\map (Deterministic)	& $1.05\pms{0.15}$ & $1.48\pms{0.26}$ & $1.09\pms{0.15}$ & $87.6\pms{0.8}$ & $81.5\pms{1.2}$ & $87.1\pms{0.8}$ & $63.6\pms{2.5}$ & $-$ & $77.5\pms{1.7}$ \\
\textsc{deep ensemble}	& $0.39\pms{0.05}$ & $0.49\pms{0.09}$ & $0.40\pms{0.05}$ & $89.6\pms{0.4}$ & $83.1\pms{0.5}$ & $89.1\pms{0.4}$ & $68.1\pms{1.4}$ & $-$ & $81.3\pms{0.8}$ \\
\mcd	& $0.32\pms{0.02}$ & $0.31\pms{0.03}$ & $0.32\pms{0.02}$ & $89.0\pms{0.8}$ & $87.5\pms{1.1}$ & $88.9\pms{0.8}$ & $72.6\pms{2.1}$ & $-$ & $83.5\pms{1.5}$ \\
\mcd \textsc{ensemble}	& $\mathbf{0.26\pms{0.00}}$ & $\mathbf{0.24\pms{0.01}}$ & $\mathbf{0.26\pms{0.00}}$ & $\mathbf{90.9\pms{0.1}}$ & $\mathbf{89.2\pms{0.2}}$ & $\mathbf{90.8\pms{0.1}}$ & $\mathbf{76.9\pms{0.2}}$ & $-$ & $\mathbf{86.6\pms{0.1}}$ \\
\fsvi	& $0.40\pms{0.03}$ & $0.57\pms{0.03}$ & $0.41\pms{0.02}$ & $87.8\pms{0.7}$ & $79.8\pms{1.1}$ & $87.1\pms{0.7}$ & $63.3\pms{2.1}$ & $-$ & $77.1\pms{1.5}$ \\
\fsvi \textsc{ensemble}	& $0.29\pms{0.00}$ & $0.41\pms{0.01}$ & $0.30\pms{0.00}$ & $90.0\pms{0.2}$ & $81.5\pms{0.5}$ & $89.4\pms{0.2}$ & $68.7\pms{0.7}$ & $-$ & $81.4\pms{0.4}$ \\
\textsc{radial}-\mfvi	& $0.37\pms{0.01}$ & $0.76\pms{0.12}$ & $0.40\pms{0.02}$ & $86.7\pms{0.4}$ & $52.0\pms{9.0}$ & $83.9\pms{1.1}$ & $49.1\pms{3.5}$ & $-$ & $66.9\pms{2.9}$ \\
\textsc{radial}-\mfvi \textsc{ensemble}	& $0.35\pms{0.01}$ & $0.73\pms{0.07}$ & $0.38\pms{0.01}$ & $87.4\pms{0.4}$ & $54.4\pms{5.5}$ & $84.8\pms{0.8}$ & $56.2\pms{2.5}$ & $-$ & $73.5\pms{2.0}$ \\
\textsc{rank}-1	& $0.56\pms{0.06}$ & $1.14\pms{0.15}$ & $0.61\pms{0.07}$ & $88.3\pms{0.6}$ & $67.5\pms{4.1}$ & $86.6\pms{0.9}$ & $59.4\pms{3.7}$ & $-$ & $74.1\pms{2.5}$ \\
\textsc{rank}-1 \textsc{ensemble}	& $0.29\pms{0.01}$ & $0.60\pms{0.04}$ & $0.32\pms{0.01}$ & $89.3\pms{0.4}$ & $69.7\pms{2.2}$ & $87.7\pms{0.5}$ & $66.5\pms{2.5}$ & $-$ & $80.0\pms{1.6}$ \\
\mfvi	& $0.56\pms{0.08}$ & $0.75\pms{0.20}$ & $0.57\pms{0.09}$ & $83.7\pms{0.3}$ & $79.8\pms{2.3}$ & $83.4\pms{0.1}$ & $55.2\pms{0.6}$ & $-$ & $71.0\pms{0.8}$ \\
\mfvi \textsc{ensemble}	& $0.35\pms{0.00}$ & $0.37\pms{0.01}$ & $0.36\pms{0.00}$ & $86.2\pms{0.3}$ & $81.6\pms{0.7}$ & $85.8\pms{0.3}$ & $59.9\pms{0.3}$ & $-$ & $75.3\pms{0.2}$ \\

\midrule
& \multicolumn{3}{c}{AUROC (\%) $\uparrow$} & \multicolumn{3}{c}{ECE $\downarrow$} \\
\midrule

\map (Deterministic)	& $83.7\pms{1.1}$ & $-$ & $87.8\pms{0.8}$ & $0.09\pms{0.01}$ & $0.15\pms{0.03}$ & $0.09\pms{0.01}$ \\
\textsc{deep ensemble}	& $86.3\pms{0.5}$ & $-$ & $90.0\pms{0.4}$ & $0.03\pms{0.01}$ & $0.07\pms{0.01}$ & $0.03\pms{0.01}$ \\
\mcd	& $88.2\pms{1.1}$ & $-$ & $91.1\pms{0.9}$ & $0.02\pms{0.00}$ & $0.06\pms{0.01}$ & $0.02\pms{0.01}$ \\
\mcd \textsc{ensemble}	& $\mathbf{90.6\pms{0.1}}$ & $-$ & $\mathbf{93.1\pms{0.1}}$ & $0.02\pms{0.00}$ & $\mathbf{0.02\pms{0.00}}$ & $0.02\pms{0.00}$ \\
\fsvi	& $82.8\pms{1.0}$ & $-$ & $86.8\pms{0.7}$ & $0.06\pms{0.01}$ & $0.14\pms{0.01}$ & $0.06\pms{0.01}$ \\
\fsvi \textsc{ensemble}	& $86.1\pms{0.3}$ & $-$ & $89.7\pms{0.2}$ & $0.03\pms{0.00}$ & $0.08\pms{0.01}$ & $0.03\pms{0.00}$ \\
\textsc{radial}-\mfvi	& $76.9\pms{1.8}$ & $-$ & $82.2\pms{1.6}$ & $0.05\pms{0.01}$ & $0.23\pms{0.07}$ & $0.04\pms{0.01}$ \\
\textsc{radial}-\mfvi \textsc{ensemble}	& $81.3\pms{1.4}$ & $-$ & $86.2\pms{1.2}$ & $0.07\pms{0.01}$ & $0.15\pms{0.04}$ & $0.06\pms{0.01}$ \\
\textsc{rank}-1	& $81.6\pms{1.8}$ & $-$ & $85.8\pms{1.4}$ & $0.06\pms{0.01}$ & $0.22\pms{0.03}$ & $0.07\pms{0.02}$ \\
\textsc{rank}-1 \textsc{ensemble}	& $85.1\pms{1.3}$ & $-$ & $89.1\pms{0.9}$ & $\mathbf{0.02\pms{0.00}}$ & $0.12\pms{0.02}$ & $0.03\pms{0.00}$ \\
\mfvi	& $79.8\pms{0.5}$ & $-$ & $84.3\pms{0.5}$ & $0.07\pms{0.02}$ & $0.12\pms{0.03}$ & $0.07\pms{0.02}$ \\
\mfvi \textsc{ensemble}	& $82.3\pms{0.1}$ & $-$ & $86.8\pms{0.1}$ & $0.02\pms{0.00}$ & $0.05\pms{0.01}$ & $\mathbf{0.02\pms{0.00}}$ \\
\midrule
\midrule
\end{tabular}
}
\label{tab:metrics_severity_joint_standard}
\vspace{-5pt}
\end{table*}

\begin{table*}[!tb]
\centering
\caption{
     \textbf{Expert Referral Metrics, Country Shift, Tuned on Joint Dataset.} 
     We assess model predictive performance and uncertainty quantification in the context of expert referral.
     Here all methods are tuned according to the joint validation metric (\Cref{sec:joint_tuning}): area under the Selective Prediction Accuracy curve constructed on the balanced joint validation dataset (composed of the in-domain and upsampled shifted validation datasets).
     We construct referral curves on a variety of metrics---AUC, Accuracy, NLL and AUPRC---by sweeping over the referral thresholds $\tau$, obtaining a point for each possible partition of the dataset into ``referred" and ``non-referred". 
     The Balanced evaluation dataset is constructed using the procedure described in \Cref{sec:joint_tuning}.
}
\vspace{-6pt}
\resizebox{1.0\linewidth}{0.15\paperheight}{%
\begin{tabular}{@{\extracolsep{2pt}}lcccccccc@{}}
\midrule
\midrule
& \multicolumn{4}{c}{R-AUROC AUC $\uparrow$} & \multicolumn{4}{c}{R-Accuracy AUC $\uparrow$} \\
\cline{2-5}
\cline{6-9}\\
\textbf{Method}     &
\textbf{In-Domain}         &
\textbf{Shifted}            &
\textbf{Joint}       &
\textbf{Balanced}         &
\textbf{In-Domain}         &
\textbf{Shifted}            &
\textbf{Joint}       &
\textbf{Balanced}       \\
\midrule
\midrule

\map (Deterministic)	& $90.1\pms{0.9}$ & $76.0\pms{1.6}$ & $89.9\pms{0.4}$ & $89.9\pms{0.4}$ & $95.6\pms{0.2}$ & $88.2\pms{0.9}$ & $94.8\pms{0.1}$ & $90.8\pms{0.6}$ \\
\textsc{deep ensemble}	& $91.8\pms{0.3}$ & $80.7\pms{1.6}$ & $92.0\pms{0.1}$ & $92.0\pms{0.1}$ & $96.6\pms{0.0}$ & $90.7\pms{0.8}$ & $96.0\pms{0.1}$ & $92.9\pms{0.6}$ \\
\mcd	& $94.4\pms{0.5}$ & $86.1\pms{2.1}$ & $94.5\pms{0.3}$ & $94.5\pms{0.3}$ & $96.6\pms{0.1}$ & $90.8\pms{0.8}$ & $96.0\pms{0.1}$ & $93.0\pms{0.5}$ \\
\mcd \textsc{ensemble}	& $\mathbf{95.2\pms{0.1}}$ & $86.9\pms{1.0}$ & $\mathbf{95.3\pms{0.0}}$ & $\mathbf{95.3\pms{0.0}}$ & $\mathbf{97.2\pms{0.0}}$ & $91.1\pms{0.4}$ & $\mathbf{96.6\pms{0.1}}$ & $93.5\pms{0.3}$ \\
\fsvi	& $88.4\pms{1.1}$ & $90.5\pms{1.3}$ & $90.0\pms{0.7}$ & $90.0\pms{0.7}$ & $95.4\pms{0.2}$ & $92.9\pms{0.7}$ & $95.2\pms{0.1}$ & $93.9\pms{0.3}$ \\
\fsvi \textsc{ensemble}	& $88.5\pms{0.9}$ & $93.8\pms{0.9}$ & $90.3\pms{0.7}$ & $90.3\pms{0.7}$ & $96.1\pms{0.1}$ & $94.3\pms{0.5}$ & $95.9\pms{0.1}$ & $\mathbf{95.1\pms{0.2}}$ \\
\textsc{radial}-\mfvi	& $82.0\pms{2.1}$ & $91.7\pms{1.8}$ & $84.4\pms{1.8}$ & $84.4\pms{1.8}$ & $93.7\pms{0.2}$ & $92.7\pms{0.8}$ & $93.5\pms{0.2}$ & $92.9\pms{0.4}$ \\
\textsc{radial}-\mfvi \textsc{ensemble}	& $80.6\pms{1.3}$ & $\mathbf{95.3\pms{0.7}}$ & $83.7\pms{1.2}$ & $83.7\pms{1.2}$ & $94.4\pms{0.1}$ & $\mathbf{94.6\pms{0.4}}$ & $94.3\pms{0.1}$ & $94.2\pms{0.1}$ \\
\textsc{rank}-1	& $88.7\pms{0.8}$ & $79.5\pms{1.8}$ & $89.0\pms{0.5}$ & $89.0\pms{0.5}$ & $94.1\pms{0.5}$ & $88.3\pms{0.7}$ & $93.5\pms{0.4}$ & $90.4\pms{0.4}$ \\
\textsc{rank}-1 \textsc{ensemble}	& $91.7\pms{0.6}$ & $84.4\pms{0.3}$ & $92.3\pms{0.5}$ & $92.3\pms{0.5}$ & $96.0\pms{0.3}$ & $90.9\pms{0.3}$ & $95.5\pms{0.3}$ & $93.0\pms{0.2}$ \\
\mfvi	& $85.7\pms{2.1}$ & $84.3\pms{2.5}$ & $87.0\pms{1.3}$ & $87.0\pms{1.3}$ & $93.6\pms{0.4}$ & $89.9\pms{1.2}$ & $93.3\pms{0.3}$ & $91.3\pms{0.6}$ \\
\mfvi \textsc{ensemble}	& $85.0\pms{2.4}$ & $91.5\pms{1.8}$ & $88.0\pms{1.4}$ & $88.0\pms{1.4}$ & $95.2\pms{0.3}$ & $93.7\pms{0.9}$ & $95.0\pms{0.2}$ & $94.2\pms{0.5}$ \\

\midrule
& \multicolumn{4}{c}{R-NLL AUC $\downarrow$} & \multicolumn{4}{c}{R-AUPRC AUC $\uparrow$} \\
\midrule

\map (Deterministic)    & $0.91\pms{0.06}$ & $3.78\pms{0.33}$ & $1.20\pms{0.03}$ & $2.73\pms{0.20}$ & $87.2\pms{2.3}$ & $92.8\pms{0.4}$ & $88.5\pms{1.6}$ & $88.5\pms{1.6}$ \\
\textsc{deep ensemble}  & $0.48\pms{0.01}$ & $2.69\pms{0.29}$ & $0.72\pms{0.04}$ & $1.86\pms{0.21}$ & $87.9\pms{1.2}$ & $\mathbf{93.9\pms{0.4}}$ & $89.4\pms{0.7}$ & $89.4\pms{0.7}$ \\
\mcd    & $0.20\pms{0.01}$ & $1.23\pms{0.19}$ & $0.30\pms{0.02}$ & $0.81\pms{0.12}$ & $90.7\pms{1.1}$ & $90.5\pms{0.7}$ & $90.7\pms{0.7}$ & $90.7\pms{0.7}$ \\
\mcd \textsc{ensemble}  & $\mathbf{0.14\pms{0.00}}$ & $0.87\pms{0.09}$ & $\mathbf{0.21\pms{0.01}}$ & $0.57\pms{0.06}$ & $\mathbf{91.7\pms{0.4}}$ & $90.1\pms{0.6}$ & $\mathbf{91.5\pms{0.2}}$ & $\mathbf{91.5\pms{0.2}}$ \\
\fsvi   & $0.36\pms{0.03}$ & $1.06\pms{0.17}$ & $0.43\pms{0.04}$ & $0.77\pms{0.10}$ & $79.2\pms{2.7}$ & $91.0\pms{1.4}$ & $82.2\pms{1.7}$ & $82.2\pms{1.7}$ \\
\fsvi \textsc{ensemble} & $0.25\pms{0.02}$ & $0.55\pms{0.09}$ & $0.28\pms{0.01}$ & $0.42\pms{0.04}$ & $77.6\pms{2.3}$ & $91.8\pms{0.8}$ & $81.2\pms{1.7}$ & $81.2\pms{1.7}$ \\
\textsc{radial}-\mfvi   & $0.47\pms{0.12}$ & $0.87\pms{0.37}$ & $0.52\pms{0.14}$ & $0.74\pms{0.28}$ & $66.3\pms{4.9}$ & $92.1\pms{0.3}$ & $71.7\pms{4.1}$ & $71.7\pms{4.1}$ \\
\textsc{radial}-\mfvi \textsc{ensemble} & $0.26\pms{0.01}$ & $\mathbf{0.27\pms{0.03}}$ & $0.26\pms{0.01}$ & $\mathbf{0.28\pms{0.02}}$ & $61.9\pms{3.0}$ & $92.9\pms{0.2}$ & $68.7\pms{2.6}$ & $68.7\pms{2.6}$ \\
\textsc{rank}-1 & $0.88\pms{0.11}$ & $3.14\pms{0.47}$ & $1.10\pms{0.13}$ & $2.29\pms{0.35}$ & $83.8\pms{2.1}$ & $92.1\pms{0.2}$ & $85.5\pms{1.6}$ & $85.5\pms{1.6}$ \\
Rank1 Ensemble  & $0.32\pms{0.03}$ & $1.83\pms{0.26}$ & $0.46\pms{0.04}$ & $1.19\pms{0.16}$ & $86.4\pms{1.0}$ & $92.9\pms{0.1}$ & $87.7\pms{0.8}$ & $87.7\pms{0.8}$ \\
\mfvi   & $1.01\pms{0.16}$ & $2.64\pms{0.52}$ & $1.15\pms{0.18}$ & $1.97\pms{0.36}$ & $76.0\pms{5.3}$ & $92.2\pms{0.6}$ & $79.9\pms{3.6}$ & $79.9\pms{3.6}$ \\
\mfvi \textsc{ensemble} & $0.37\pms{0.06}$ & $1.08\pms{0.40}$ & $0.44\pms{0.09}$ & $0.80\pms{0.26}$ & $71.6\pms{5.4}$ & $93.8\pms{0.6}$ & $78.1\pms{3.4}$ & $78.1\pms{3.4}$ \\

\midrule
\midrule
\end{tabular}
}
\label{tab:metrics_country_joint}
\vspace{-5pt}
\end{table*}

\begin{table*}[!tb]
\vspace{-2mm}
\centering
\caption{
     \textbf{Expert Referral Metrics, Severity Shift, Tuned on Joint Dataset.} 
     We assess model predictive performance and uncertainty quantification in the context of expert referral.
     Here all methods are tuned according to the joint validation metric (\Cref{sec:joint_tuning}): area under the retention--accuracy curve constructed on the balanced joint validation dataset (composed of the in-domain and upsampled shifted validation datasets).
     We construct referral curves on a variety of metrics---AUC, Accuracy, NLL and AUPRC---by sweeping over the referral thresholds $\tau$, obtaining a point for each possible partition of the dataset into ``referred" and ``non-referred". 
     The Balanced evaluation dataset is constructed using the procedure described in \Cref{sec:joint_tuning}.
}
\vspace{-3pt}
\resizebox{1.0\linewidth}{!}{%
\begin{tabular}{@{\extracolsep{2pt}}lcccccccc@{}}
\midrule
\midrule
& \multicolumn{4}{c}{R-AUROC AUC $\uparrow$} & \multicolumn{4}{c}{R-Accuracy AUC $\uparrow$} \\
\cline{2-5}
\cline{6-9}\\
\textbf{Method}     &
\textbf{In-Domain}         &
\textbf{Shifted}            &
\textbf{Joint}       &
\textbf{Balanced}         &
\textbf{In-Domain}         &
\textbf{Shifted}            &
\textbf{Joint}       &
\textbf{Balanced}       \\
\midrule
\midrule

\map (Deterministic)	& $87.7\pms{0.8}$ & $-$ & $90.2\pms{0.5}$ & $90.2\pms{0.5}$ & $94.3\pms{0.6}$ & $94.9\pms{0.7}$ & $94.6\pms{0.5}$ & $95.1\pms{0.5}$ \\
\textsc{deep ensemble}	& $88.9\pms{0.5}$ & $-$ & $92.6\pms{0.2}$ & $92.6\pms{0.2}$ & $95.7\pms{0.2}$ & $95.6\pms{0.2}$ & $95.9\pms{0.2}$ & $96.1\pms{0.2}$ \\
\mcd	& $92.9\pms{0.7}$ & $-$ & $94.4\pms{0.4}$ & $94.4\pms{0.4}$ & $95.7\pms{0.4}$ & $97.5\pms{0.6}$ & $96.0\pms{0.4}$ & $96.9\pms{0.5}$ \\
\mcd \textsc{ensemble}	& $\mathbf{94.2\pms{0.2}}$ & $-$ & $\mathbf{95.6\pms{0.2}}$ & $\mathbf{95.6\pms{0.2}}$ & $\mathbf{96.9\pms{0.1}}$ & $\mathbf{98.1\pms{0.1}}$ & $\mathbf{97.1\pms{0.0}}$ & $\mathbf{97.7\pms{0.1}}$ \\
\fsvi	& $87.7\pms{0.8}$ & $-$ & $90.9\pms{0.4}$ & $90.9\pms{0.4}$ & $94.1\pms{0.4}$ & $94.8\pms{0.7}$ & $94.4\pms{0.4}$ & $94.7\pms{0.5}$ \\
\fsvi \textsc{ensemble}	& $89.1\pms{0.3}$ & $-$ & $92.6\pms{0.2}$ & $92.6\pms{0.2}$ & $95.9\pms{0.1}$ & $94.8\pms{0.3}$ & $96.0\pms{0.1}$ & $95.7\pms{0.1}$ \\
\textsc{radial}-\mfvi	& $72.3\pms{4.8}$ & $-$ & $78.2\pms{4.8}$ & $78.2\pms{4.8}$ & $92.9\pms{0.6}$ & $61.7\pms{12.7}$ & $92.0\pms{0.9}$ & $83.8\pms{3.9}$ \\
\textsc{radial}-\mfvi \textsc{ensemble}	& $70.6\pms{3.2}$ & $-$ & $76.8\pms{3.6}$ & $76.8\pms{3.6}$ & $94.4\pms{0.5}$ & $60.3\pms{8.9}$ & $93.5\pms{0.6}$ & $85.2\pms{2.5}$ \\
\textsc{rank}-1	& $82.3\pms{2.7}$ & $-$ & $87.4\pms{1.5}$ & $87.4\pms{1.5}$ & $94.5\pms{0.5}$ & $83.9\pms{3.7}$ & $94.1\pms{0.6}$ & $90.8\pms{1.6}$ \\
\textsc{rank}-1 \textsc{ensemble}	& $80.7\pms{1.2}$ & $-$ & $88.1\pms{1.0}$ & $88.1\pms{1.0}$ & $95.6\pms{0.4}$ & $84.5\pms{1.7}$ & $95.3\pms{0.4}$ & $92.0\pms{0.8}$ \\
\mfvi	& $86.0\pms{0.9}$ & $-$ & $90.4\pms{0.8}$ & $90.4\pms{0.8}$ & $92.6\pms{0.1}$ & $92.6\pms{1.7}$ & $92.8\pms{0.1}$ & $92.9\pms{0.9}$ \\
\mfvi \textsc{ensemble}	& $88.2\pms{0.1}$ & $-$ & $92.3\pms{0.1}$ & $92.3\pms{0.1}$ & $94.1\pms{0.1}$ & $94.9\pms{0.3}$ & $94.3\pms{0.1}$ & $94.8\pms{0.1}$ \\

\midrule
& \multicolumn{4}{c}{R-NLL AUC $\downarrow$} & \multicolumn{4}{c}{R-AUPRC AUC $\uparrow$} \\
\midrule

\map (Deterministic)    & $1.06\pms{0.20}$ & $0.62\pms{0.15}$ & $0.96\pms{0.18}$ & $0.72\pms{0.15}$ & $78.2\pms{2.2}$ & $-$ & $88.6\pms{1.2}$ & $88.6\pms{1.2}$ \\
\textsc{deep ensemble}  & $0.32\pms{0.08}$ & $0.19\pms{0.05}$ & $0.29\pms{0.07}$ & $0.22\pms{0.05}$ & $78.7\pms{1.4}$ & $-$ & $89.9\pms{0.6}$ & $89.9\pms{0.6}$ \\
\mcd    & $0.23\pms{0.02}$ & $0.09\pms{0.03}$ & $0.20\pms{0.02}$ & $0.13\pms{0.02}$ & $86.6\pms{1.7}$ & $-$ & $93.3\pms{1.0}$ & $93.3\pms{1.0}$ \\
\mcd \textsc{ensemble}  & $\mathbf{0.15\pms{0.01}}$ & $\mathbf{0.06\pms{0.00}}$ & $\mathbf{0.13\pms{0.00}}$ & $\mathbf{0.09\pms{0.00}}$ & $\mathbf{88.8\pms{0.5}}$ & $-$ & $\mathbf{94.6\pms{0.2}}$ & $\mathbf{94.6\pms{0.2}}$ \\
\fsvi   & $0.34\pms{0.03}$ & $0.20\pms{0.03}$ & $0.31\pms{0.03}$ & $0.24\pms{0.03}$ & $78.4\pms{2.1}$ & $-$ & $88.8\pms{1.1}$ & $88.8\pms{1.1}$ \\
\fsvi \textsc{ensemble} & $0.19\pms{0.00}$ & $0.15\pms{0.01}$ & $0.18\pms{0.00}$ & $0.15\pms{0.00}$ & $79.2\pms{0.9}$ & $-$ & $89.9\pms{0.4}$ & $89.9\pms{0.4}$ \\
\textsc{radial}-\mfvi   & $0.26\pms{0.02}$ & $0.70\pms{0.21}$ & $0.27\pms{0.03}$ & $0.38\pms{0.08}$ & $43.5\pms{9.8}$ & $-$ & $59.2\pms{9.7}$ & $59.2\pms{9.7}$ \\
\textsc{radial}-\mfvi \textsc{ensemble} & $0.24\pms{0.01}$ & $0.72\pms{0.13}$ & $0.25\pms{0.01}$ & $0.37\pms{0.04}$ & $33.7\pms{7.0}$ & $-$ & $50.8\pms{8.0}$ & $50.8\pms{8.0}$ \\
\textsc{rank}-1 & $0.49\pms{0.09}$ & $0.77\pms{0.20}$ & $0.48\pms{0.08}$ & $0.56\pms{0.11}$ & $65.9\pms{5.9}$ & $-$ & $80.2\pms{3.6}$ & $80.2\pms{3.6}$ \\
Rank1 Ensemble  & $0.18\pms{0.01}$ & $0.39\pms{0.04}$ & $0.18\pms{0.01}$ & $0.24\pms{0.02}$ & $60.9\pms{2.5}$ & $-$ & $79.0\pms{1.8}$ & $79.0\pms{1.8}$ \\
\mfvi   & $0.45\pms{0.09}$ & $0.46\pms{0.19}$ & $0.44\pms{0.10}$ & $0.43\pms{0.14}$ & $72.0\pms{1.4}$ & $-$ & $84.7\pms{1.1}$ & $84.7\pms{1.1}$ \\
\mfvi \textsc{ensemble} & $0.22\pms{0.00}$ & $0.13\pms{0.01}$ & $0.21\pms{0.00}$ & $0.16\pms{0.00}$ & $76.1\pms{0.1}$ & $-$ & $87.8\pms{0.1}$ & $87.8\pms{0.1}$ \\

\midrule
\midrule
\end{tabular}
}
\label{tab:metrics_severity_joint}
\vspace{-5pt}
\end{table*}

\clearpage

\begin{figure}[h!]
\vspace{10pt}
\hspace{5pt}
\centering
\begin{subfigure}[l]{0.55\linewidth}
    \hspace{-20pt}\includegraphics[width=\linewidth]{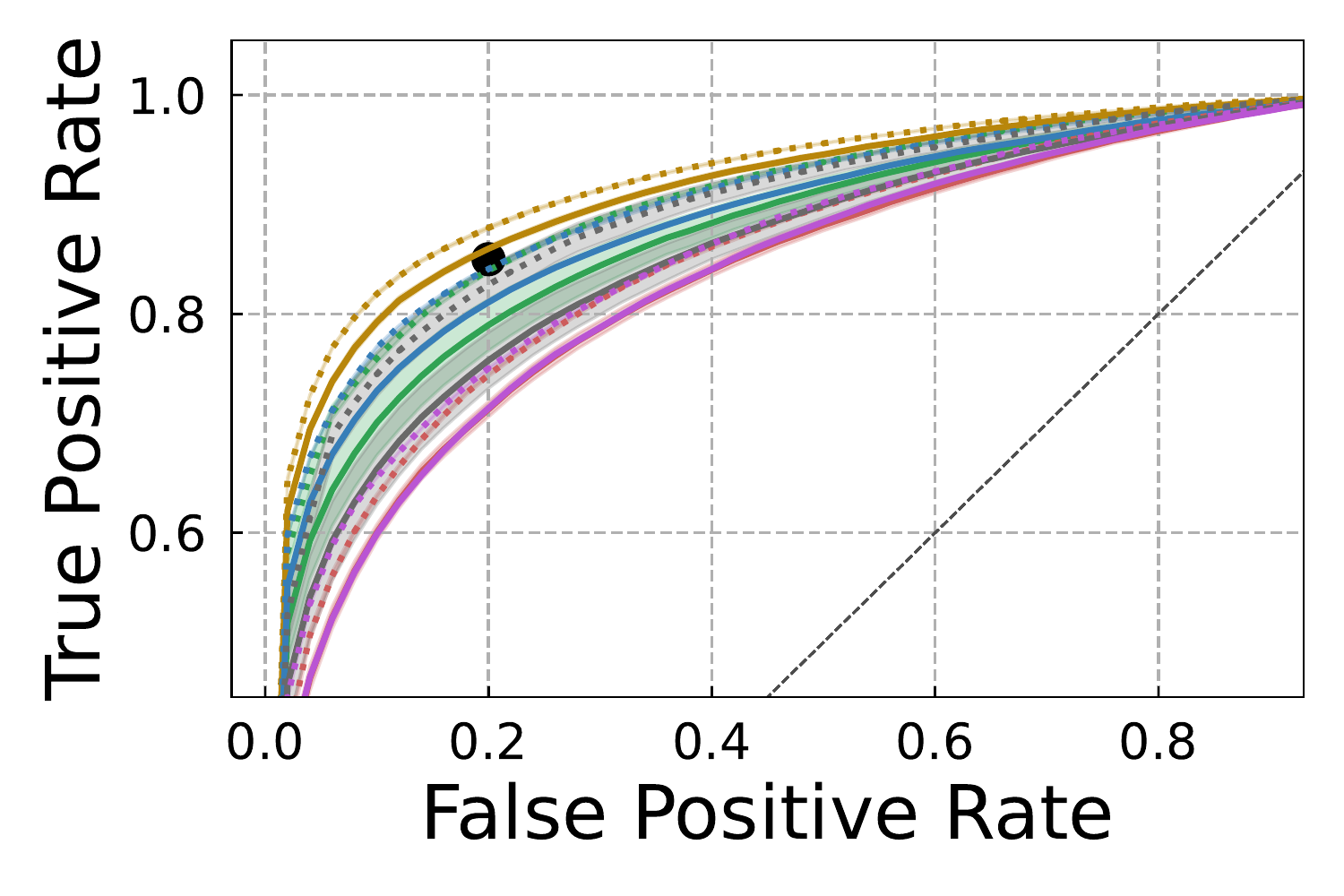}
    \caption{
        \textbf{ROC: In-Domain}
    }
\end{subfigure}
\begin{subfigure}[r]{0.55\linewidth}
  \hspace{-20pt}\includegraphics[width=\linewidth]{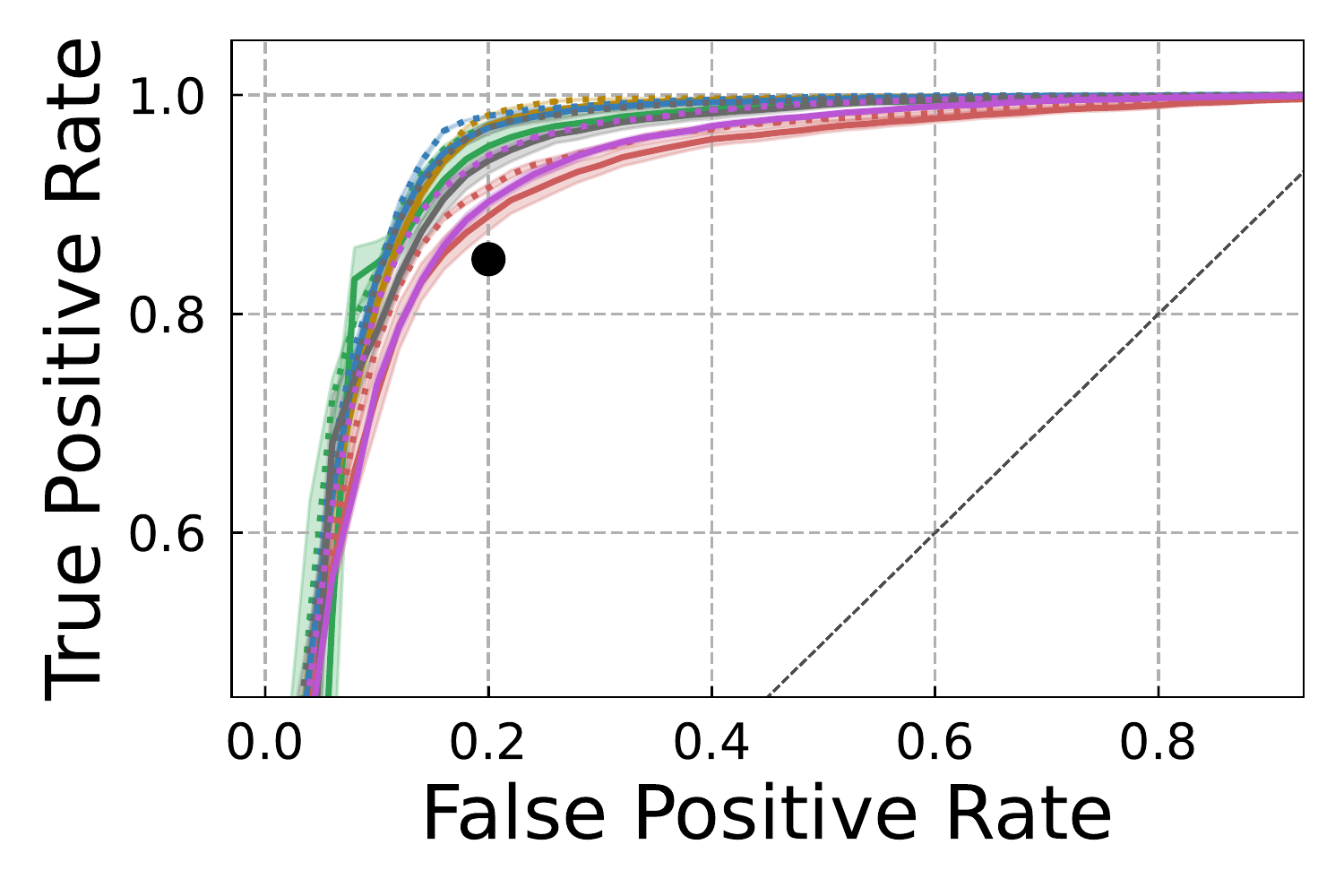}
    \caption{
        \textbf{ROC: Country Shift}
    }
\end{subfigure}
\begin{subfigure}[r]{0.55\linewidth}
\hspace{-20pt}\includegraphics[width=\linewidth]{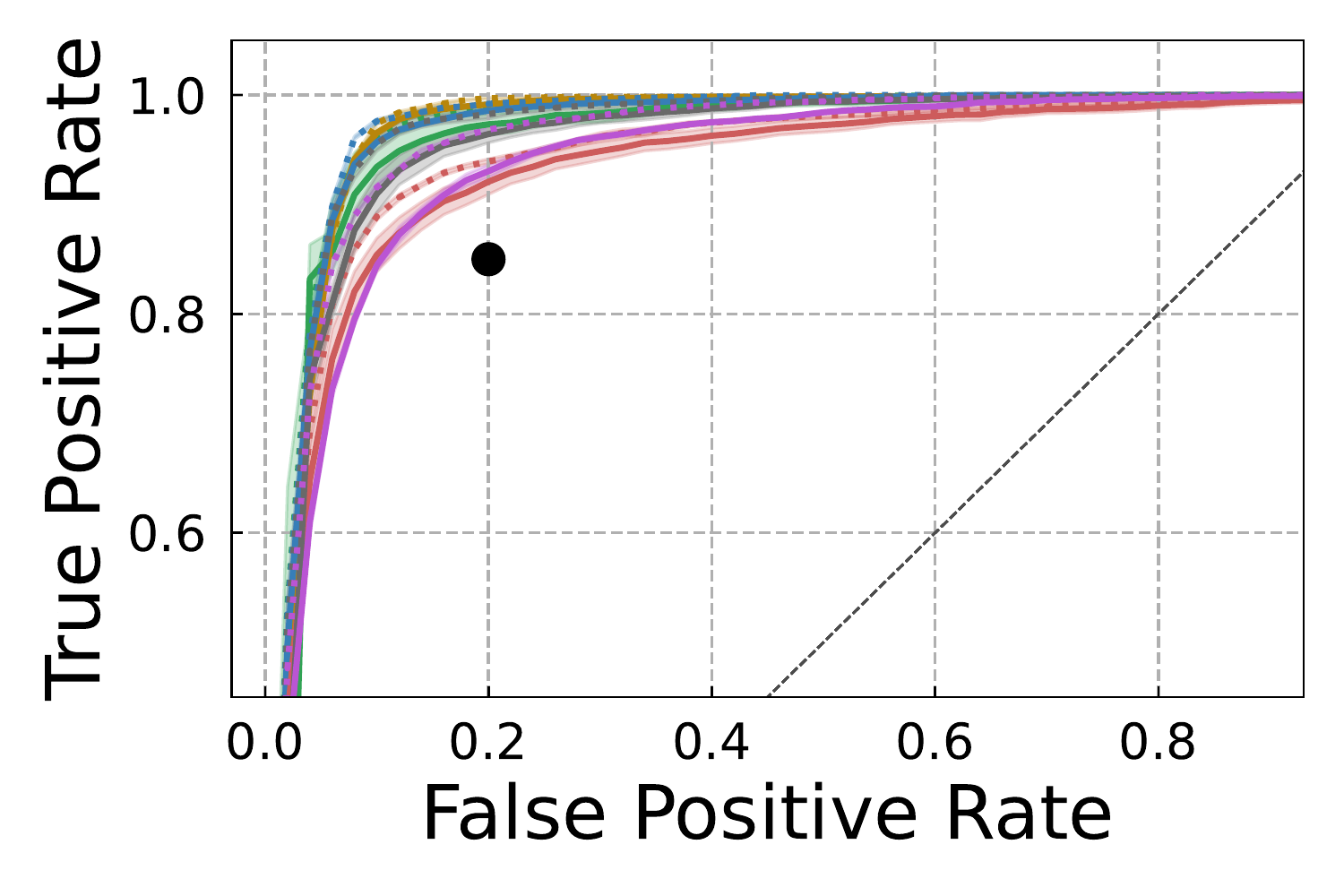}
\caption{
    \textbf{ROC: Class-Balanced Country Shift}}
\end{subfigure}
\begin{subfigure}{\linewidth}
    \includegraphics[width=\linewidth]{fig/metrics/legend.pdf}
\end{subfigure}
\hspace*{-20pt}
\\
\vspace{10pt}

\centering
\caption{
    \textbf{Class Balancing the Country Shift Dataset (ROC Curves).} We consider how balancing the proportions of the ground-truth clinical class labels---ranging from 0 (No DR) to 4 (Proliferative DR)---affects performance on the \textit{Country Shift} receiver-operating characteristic (ROC) curve.
    (\textbf{a}): ROC curve on in-domain test data.
    (\textbf{b}): ROC curve for changing medical equipment and patient populations on the shifted~\citet{APTOS_2019} test set.
    (\textbf{c}): ROC curve on the class rebalanced APTOS dataset. 
    Shading denotes one standard error. 
}
\label{fig:country_shift_class_balancing_roc}
\end{figure}

\begin{figure}[h!]
\vspace{10pt}
\hspace{5pt}
\centering
\begin{subfigure}[l]{0.55\linewidth}
    \hspace{-20pt}\includegraphics[width=\linewidth]{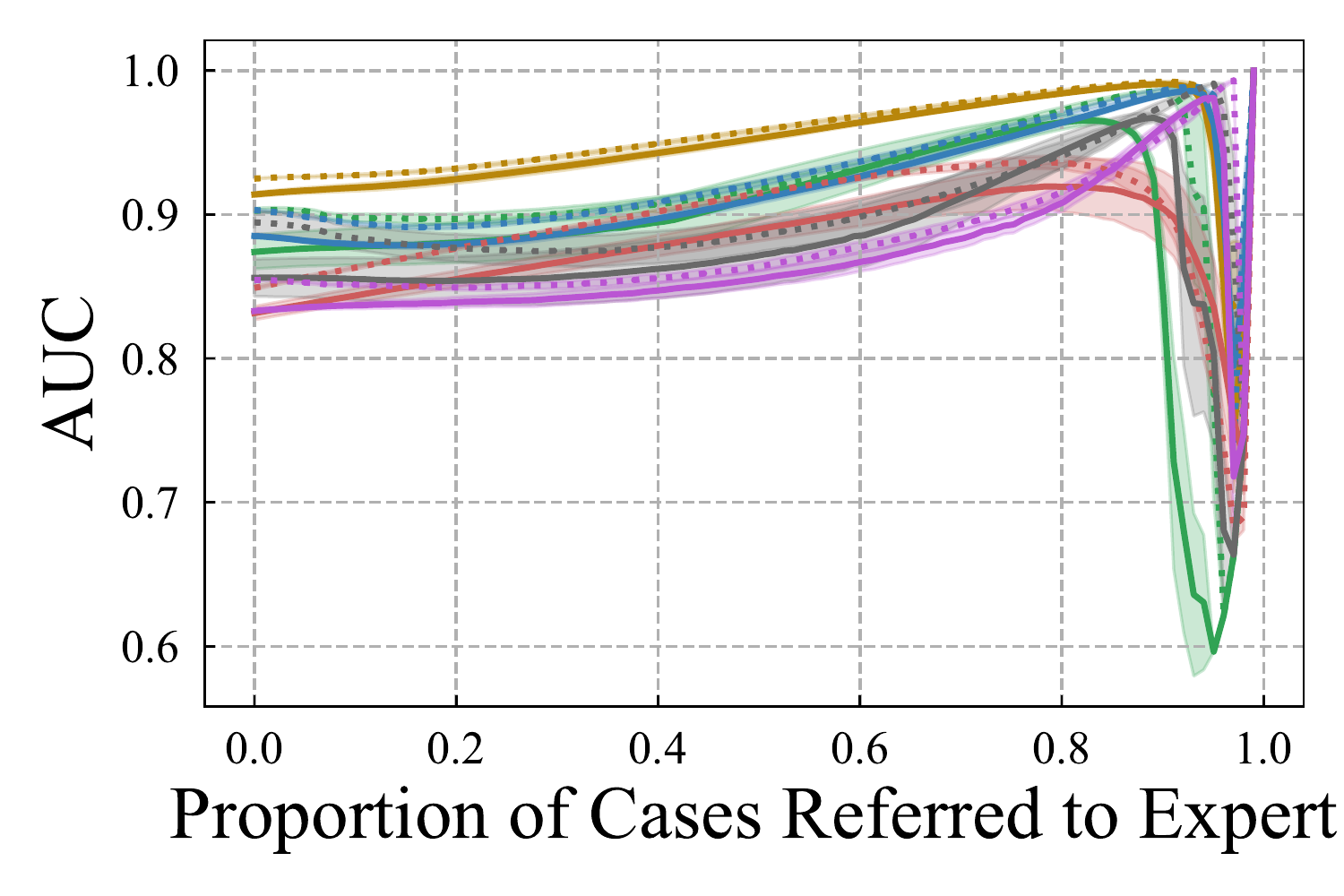}
    \caption{
        \textbf{Selective Prediction AUC: In-Domain}
    }
\end{subfigure}
\begin{subfigure}[r]{0.55\linewidth}
  \hspace{-20pt}\includegraphics[width=\linewidth]{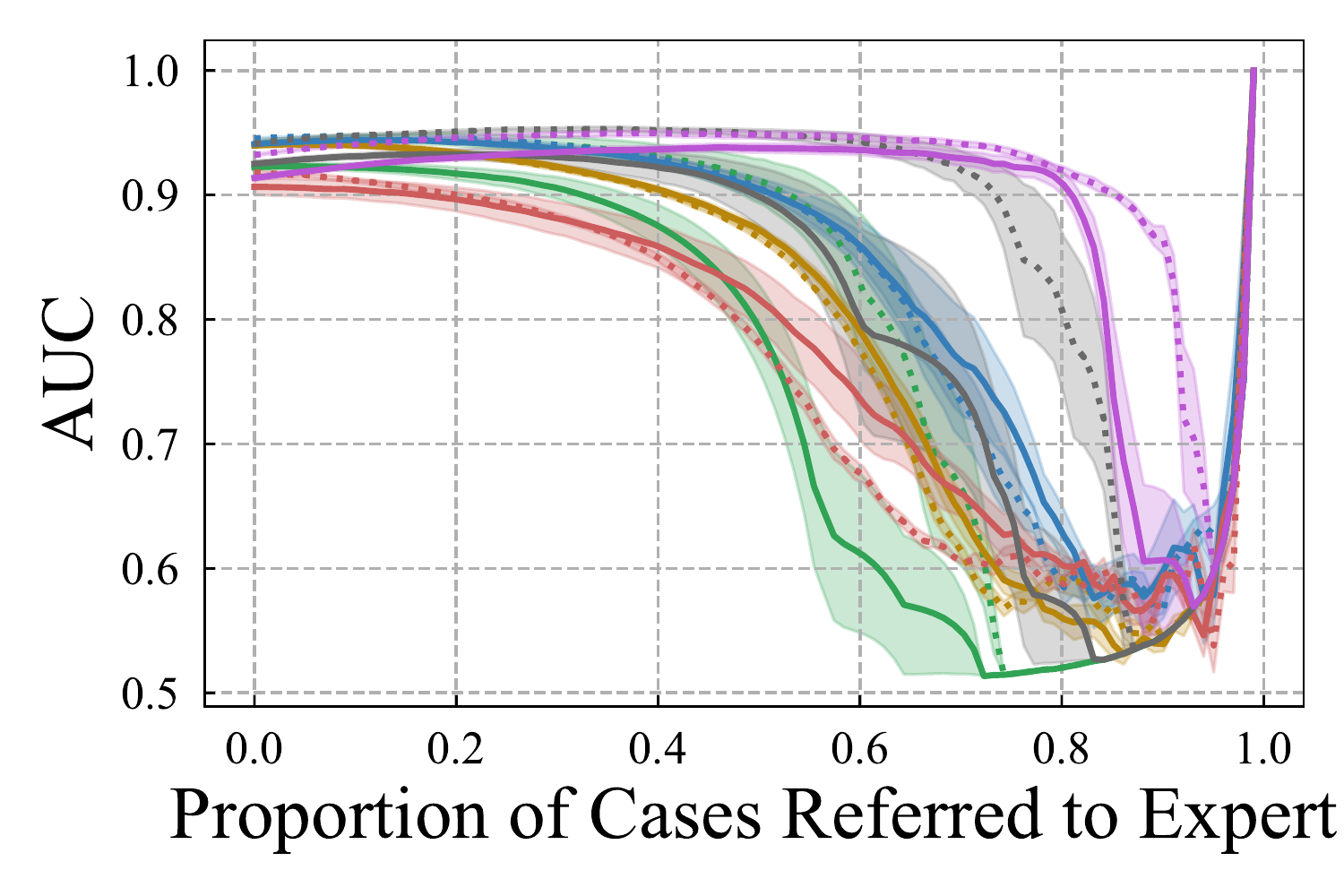}
    \caption{
        \textbf{Selective Prediction AUC: Country Shift}
    }
\end{subfigure}
\begin{subfigure}[r]{0.55\linewidth}
\hspace{-20pt}\includegraphics[width=\linewidth]{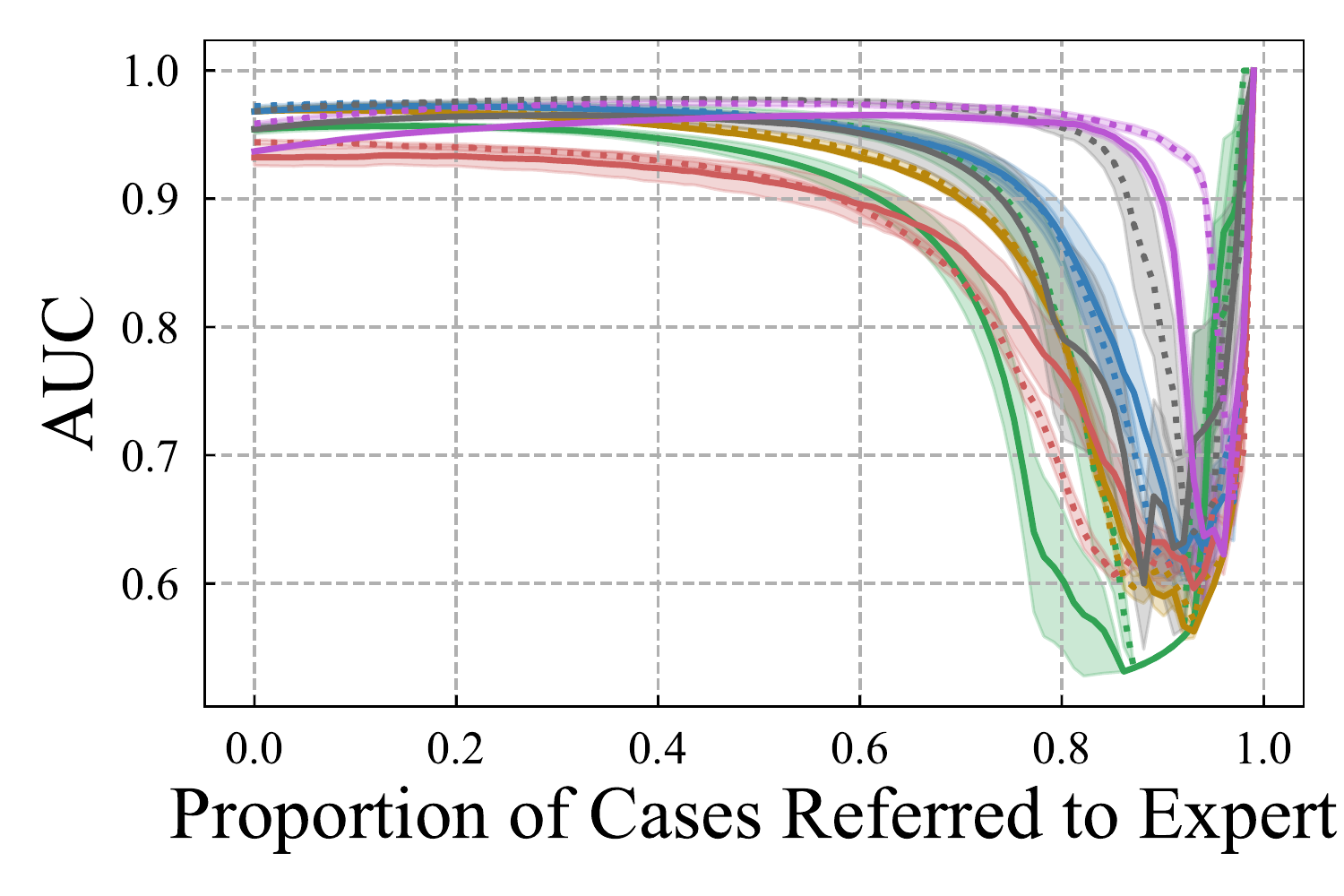}
\caption{
    \centering\textbf{Selective Prediction AUC:\hspace{100pt}
    Class-Balanced Country Shift}
}
\end{subfigure}
\begin{subfigure}{\linewidth}
    \includegraphics[width=\linewidth]{fig/metrics/legend.pdf}
\end{subfigure}
\hspace*{-20pt}
\\
\vspace{10pt}

\caption{
    \textbf{Class Balancing the Country Shift Dataset (Selective Prediction).} We consider how balancing the proportions of the ground-truth clinical class labels---ranging from 0 (No DR) to 4 (Proliferative DR)---affects performance on the \textit{Country Shift} selective prediction over AUC.
    (\textbf{a}): selective prediction AUC on in-domain test data.
    (\textbf{b}): selective prediction AUC 
    for changing medical equipment and patient populations on the shifted~\citet{APTOS_2019} test set.
    (\textbf{c}): selective prediction AUC on the class rebalanced APTOS dataset. 
    Shading denotes one standard error. 
}
\label{fig:country_shift_class_balancing_sel_pred}
\end{figure}

\subsection{Effect of Class Balancing the APTOS Dataset (\Cref{fig:country_shift_class_balancing_roc} and \ref{fig:country_shift_class_balancing_sel_pred}).}\label{subsec:app_class_balance_aptos}

We additionally investigated to what extent the change in class distribution---in terms of the ground-truth clinical labels ranging from 0 (No DR) to 4 (Proliferative DR)---contributed to the higher performance of models in AUC, and weaker performance of models in selective prediction on the APTOS dataset (the distributionally shifted dataset in the \textit{Country Shift} task) than the in-domain test dataset.

In order to normalize for the change in class distribution, we constructed a variant of the APTOS dataset with the same clinical class proportions as the in-domain EyePACS dataset. This was done by randomly sampling APTOS examples from each class, weighted by the empirical class probability of the EyePACS dataset, until reaching 10,000 samples.

In~\Cref{fig:country_shift_class_balancing_roc}, we see that the ROC curves of models on the rebalanced APTOS dataset is shifted further towards the upper left as compared to the original APTOS dataset. 
This suggests that the class proportions of the original APTOS dataset were not the reason why models obtained stronger ROC performance on APTOS than the in-domain test set---on the contrary, introducing the in-domain class proportions in the class-balanced dataset improves model performance. 

In~\Cref{fig:country_shift_class_balancing_sel_pred}, we observe that the selective prediction performance of models on this rebalanced APTOS dataset is slightly better than on the original APTOS dataset, but the ordering of models does not notably change, and performance is still significantly worse at high referral thresholds than on the in-domain data. 

This supports the claim that factors other than simply a changed class distribution, such as meaningful shifts in equipment or patient demographics, result in both stronger predictive performance at $0\%$ of data referred and poor quality of uncertainty estimates in the shifted setting.

\clearpage

\subsection{Effect of Preprocessing on Downstream Tasks}\label{subsec:app_preprocessing_exps}

Preprocessing played an important role in the EyePACS Kaggle challenge \cite{kaggle_2015}.
Here, we investigate how changes in preprocessing affect downstream predictive performance and uncertainty quantification.

In the above experiments, we used the preprocessing procedure of the Kaggle competition winner which consisted of the following steps:
\begin{itemize}
    \item[1.] Rescaling the images such that the retinas have a radius of 300 pixels,
    \item[2.] Subtracting the local average color, computed using Gaussian blur, and finally,
    \item[3.] Clipping the images to 90\% size to remove ``boundary effects''.
\end{itemize}

While (1) and (3) are (somewhat) standard techniques used to make the data more amenable for use in non-convex optimization, the standard deviation hyperparameter of the Gaussian blur kernel in (2) presupposes some amount of expert knowledge as the size of the standard deviation governs how visible certain visual artifacts are. As such, varying it has a dramatic visual effect on the preprocessed image, and likely required significant tuning.

In the preprocessing procedure, the standard deviation of the kernel is computed as $\sigma = (\texttt{target\_radius} / \texttt{blur\_constant})$, where by default, $\texttt{target\_radius}=300$ and $\texttt{blur\_constant}=30$.

\begin{figure}[h!]
\centering
  \includegraphics[width=\linewidth]{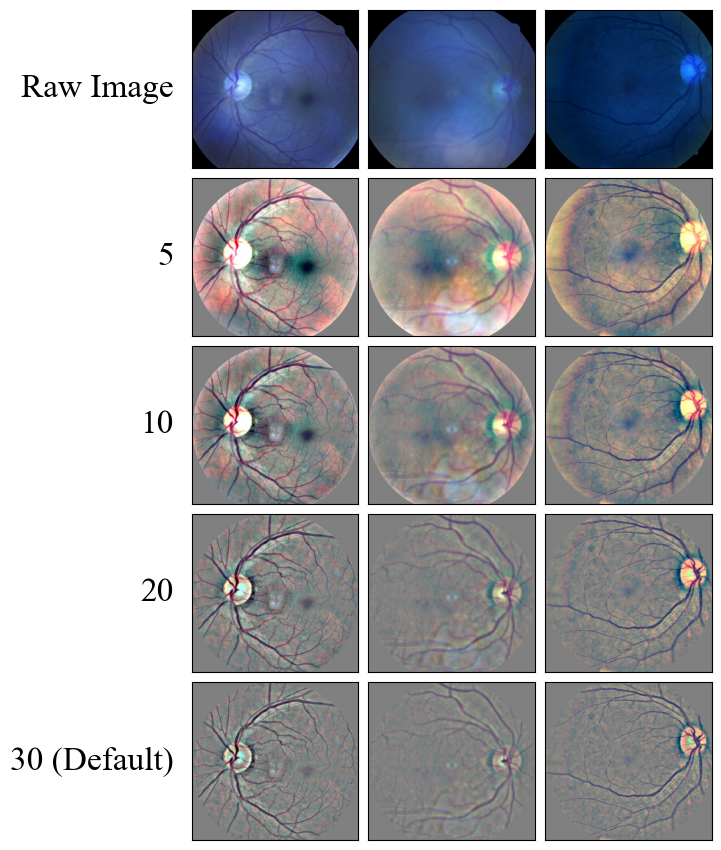}
\centering
\caption{
    \textbf{Preprocessing Examples.} Input unprocessed EyePACS images (top row), and images processed with varying \texttt{blur\_constant} (labeled on left side of grid). 
    Higher \texttt{blur\_constant} corresponds to stronger smoothing.
    }
\label{fig:preproc_example}
\vspace*{-10pt}
\end{figure}

Decreasing the \texttt{blur\_constant} results in a larger kernel standard deviation, and hence the local average color at each pixel location is computed using a larger window.
This ultimately results in the preservation of more signal as well as more noise in the input image (because lower-frequency patterns are subtracted).
See~\Cref{fig:preproc_example} for examples of unprocessed retina images along with processed images with various blur constants.

We test the downstream performance of \textsc{MAP} estimation (a deterministic model), a \textsc{Deep Ensemble}, \textsc{MC Dropout}, and an \textsc{MC Dropout Ensemble} on the Country and Severity Shift prediction tasks, varying the \texttt{blur\_constant} $\in \{5, 10, 20, 30\}$.

\begin{figure}[t!]
\centering
\begin{subfigure}[l]{0.24\linewidth}
    \hspace*{-10pt}
    \includegraphics[width=\linewidth]{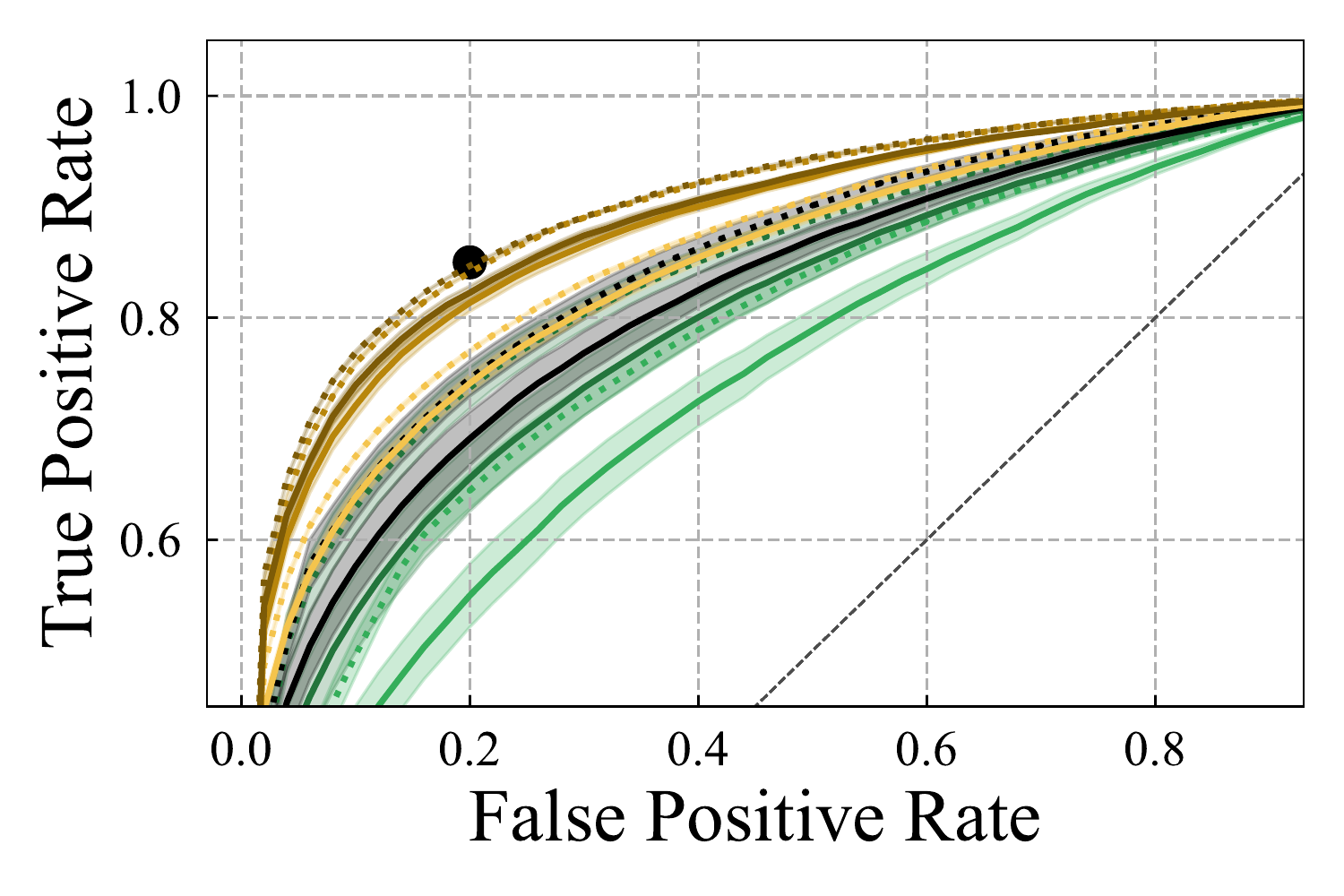}
    \vspace*{-7pt}
    \caption{
        \textbf{ROC: In-Domain\\$~$}
    }
\end{subfigure}
\begin{subfigure}[l]{0.24\linewidth}
    \hspace*{-10pt}
    \includegraphics[width=\linewidth]{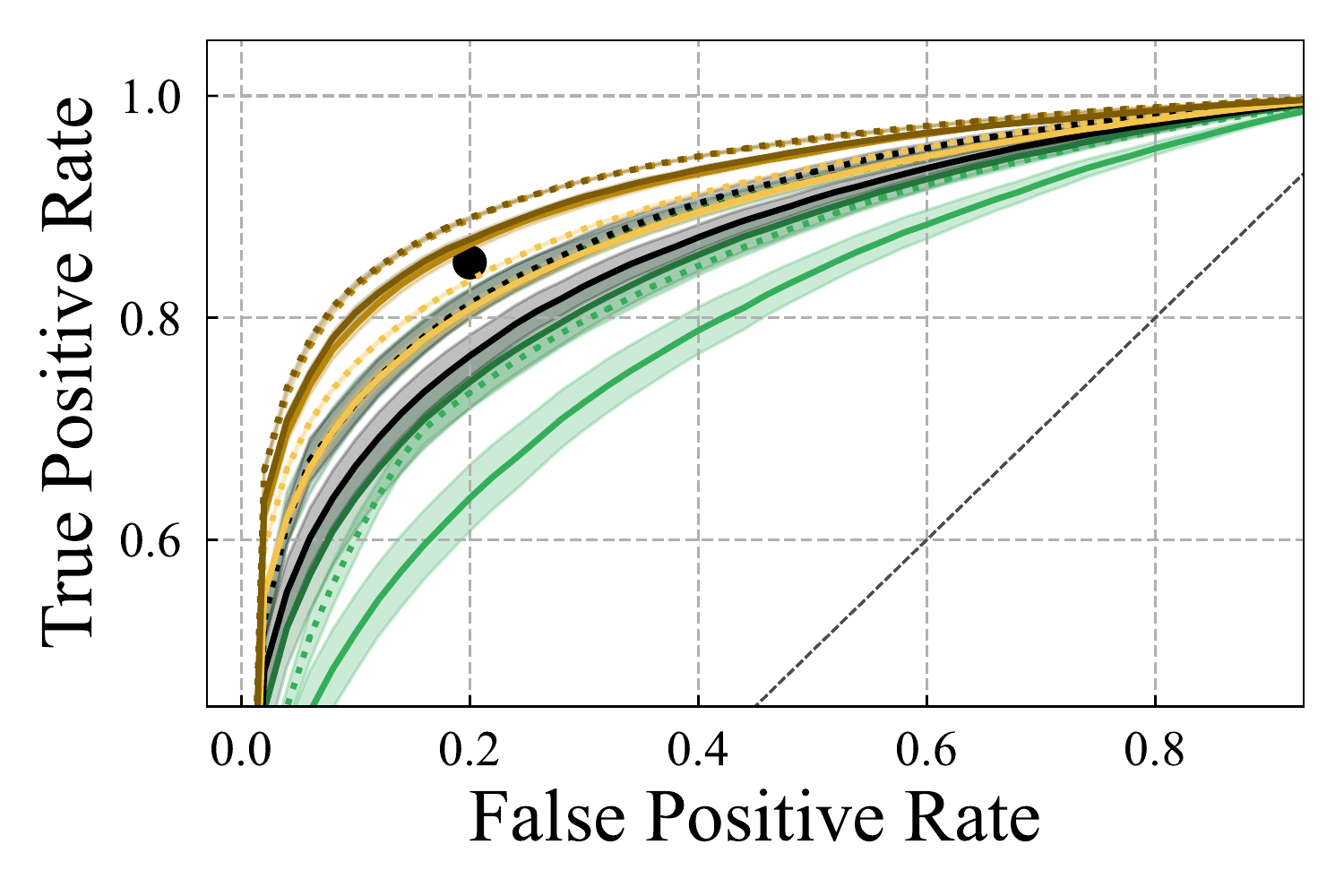}
    \vspace*{-7pt}
    \caption{
        \textbf{ROC: Joint\\$~$}
    }
\end{subfigure}
\begin{subfigure}[l]{0.24\linewidth}
    \hspace*{-10pt}
    \includegraphics[width=\linewidth]{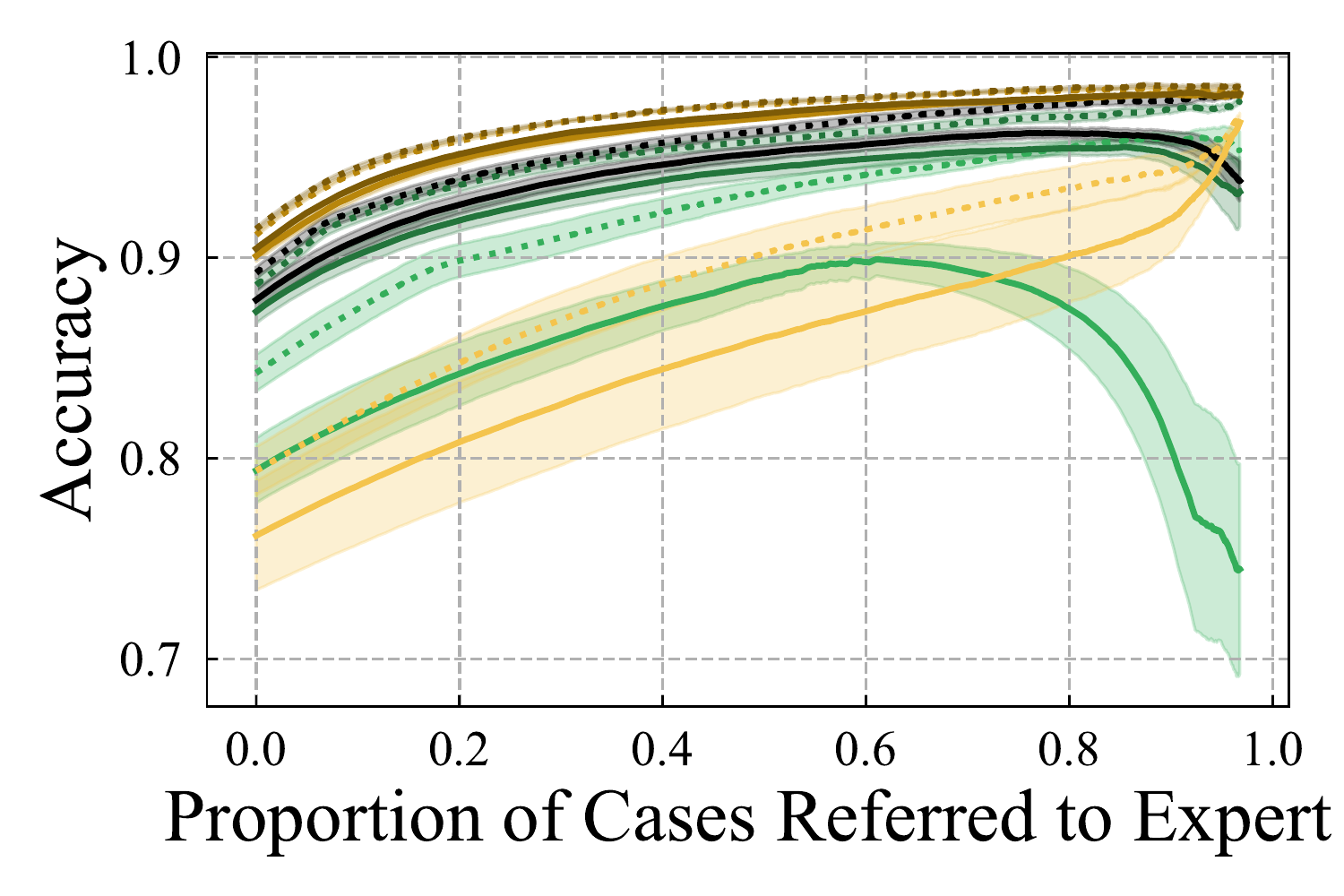}
    \vspace*{-7pt}
    \caption{
        \centering\textbf{Selective Prediction Accuracy: In-Domain}
    }
\end{subfigure}
\begin{subfigure}[r]{0.24\linewidth}
    \hspace*{-10pt}
    \includegraphics[width=\linewidth]{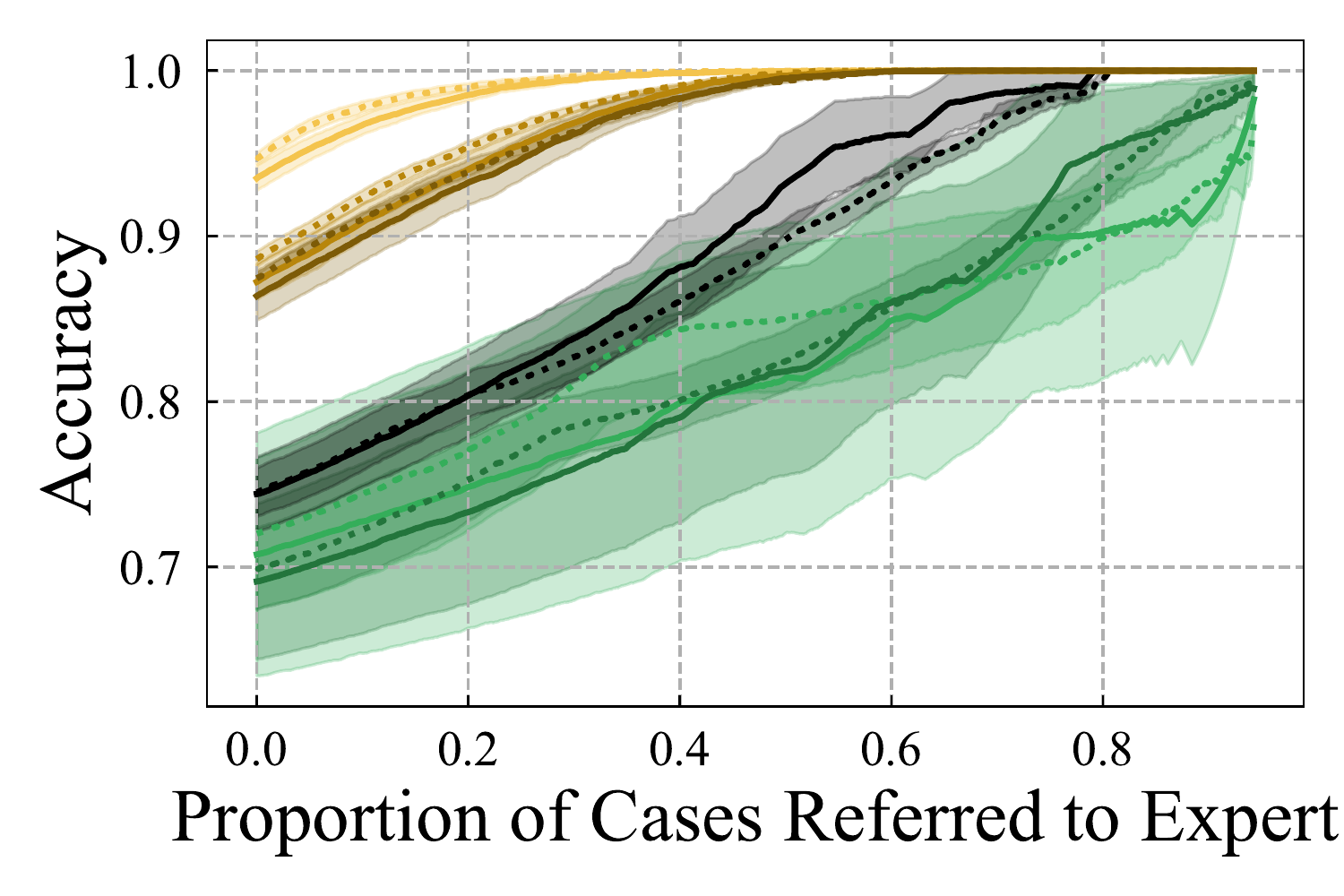}
    \vspace*{-7pt}
    \caption{
        \centering\textbf{Selective Prediction Accuracy: Severity Shift}
    }
\end{subfigure}
\begin{subfigure}{\linewidth}
    \includegraphics[width=\linewidth]{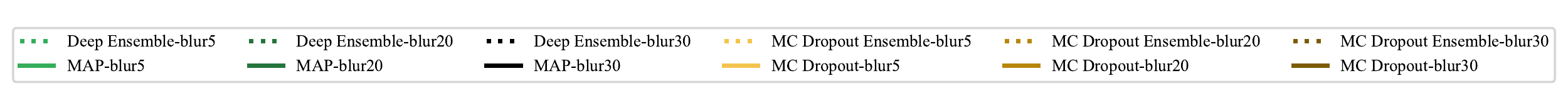}
\end{subfigure}
\vspace*{2pt}
\centering
\caption{
    \textbf{Severity Shift, Varying Blur Constant.} 
    We consider how preprocessing affects model predictive performance and uncertainty quantification on the in-domain test dataset composed only of cases with either no, mild, or moderate diabetic retinopathy, and the \textit{Severity Shift} evaluation set composed only of severe and proliferate cases.
    \textbf{Left:} The receiver operating characteristic curve (ROC) for in-domain diagnosis (\textbf{a}) and for a joint dataset composed of examples from both the in-domain and \textit{Severity Shift} evaluation sets (\textbf{b}).
    The dot in 
    \textbf{black}
    denotes the NHS-recommended 85\% sensitivity and 80\% specificity ratios~\citep{widdowson2016management}.
    \textbf{Right:} Selective prediction on accuracy in the in-domain (\textbf{c}) and \textit{Severity Shift} (\textbf{d}) settings.
    Shading denotes standard error computed over six random seeds.
    We vary the standard deviation hyperparameter of the Gaussian blur kernel through a \texttt{blur\_constant} (e.g., blur5 below corresponds to $\texttt{blur\_constant} = 5$). A higher \texttt{blur\_constant} results in a stronger smoothing of the image as per the preprocessing procedure outlined in \Cref{subsec:app_preprocessing_exps}.
    The default \texttt{blur\_constant} used in other experiments throughout this work is 30.
}
\label{fig:severity_shift_blur}
\end{figure}

\textbf{Severity Shift: Varying Blur Constant (\Cref{fig:severity_shift_blur},~\Cref{tab:metrics_preproc_severity}).}~~~On the in-domain evaluation dataset, higher \texttt{blur\_constant} (corresponding to stronger smoothing) tends to perform better across \map and \mcd, single and ensembled models, and the various referral thresholds. 
However, on the Severity Shift (distributionally shifted evaluation dataset), the \mcd variants perform better with \emph{lower} \texttt{blur\_constant}. 
This highlights the importance for practitioners to test changes in experimental settings, including preprocessing, across a variety of uncertainty quantification methods. 

\begin{figure}[t!]
\centering
\begin{subfigure}[l]{0.24\linewidth}
\hspace*{-10pt}
    \includegraphics[width=\linewidth]{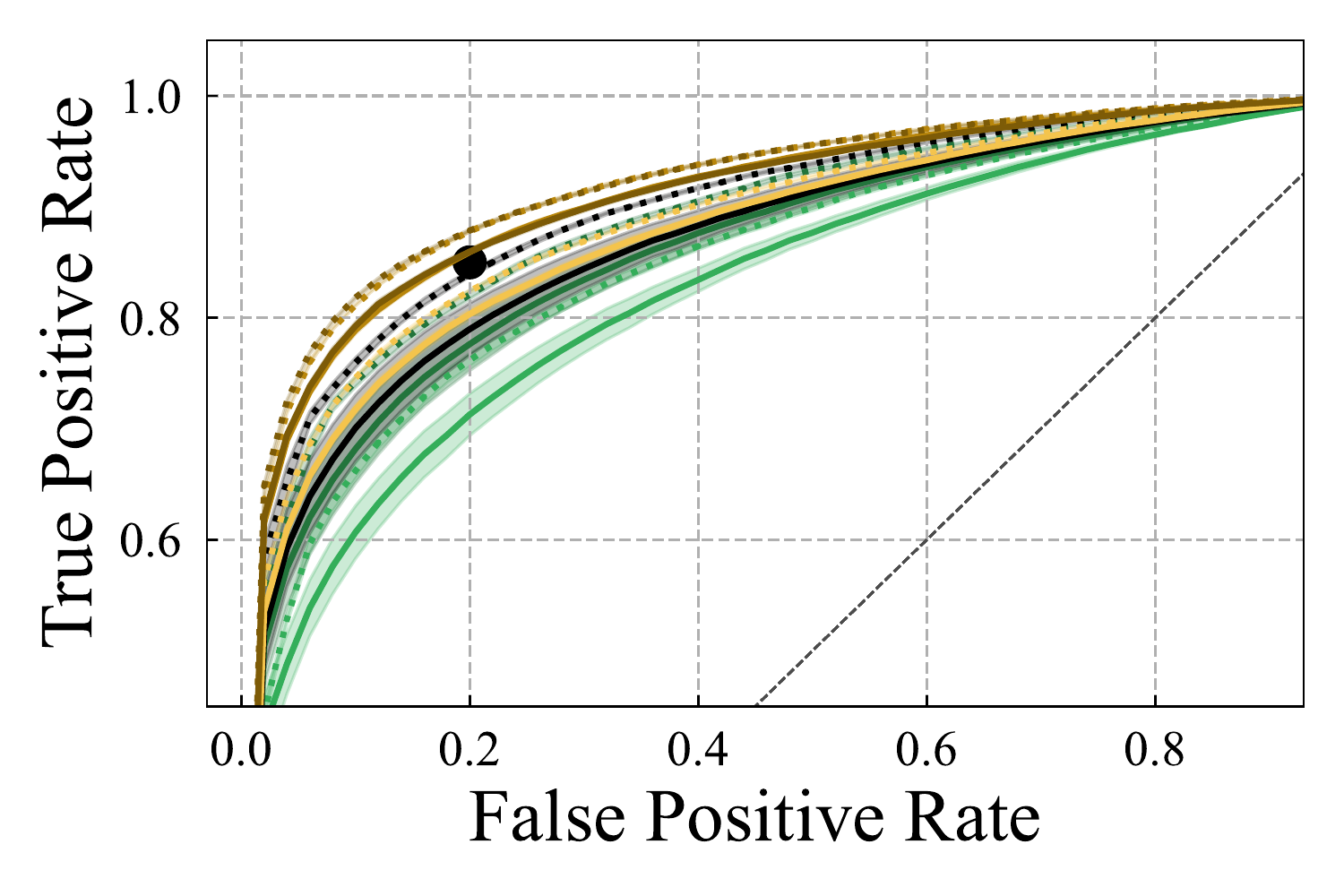}
    \vspace*{-7pt}
    \caption{
        \textbf{ROC: In-Domain\\$~$}
    }
\end{subfigure}
\begin{subfigure}[l]{0.24\linewidth}
    \hspace*{-10pt}
    \includegraphics[width=\linewidth]{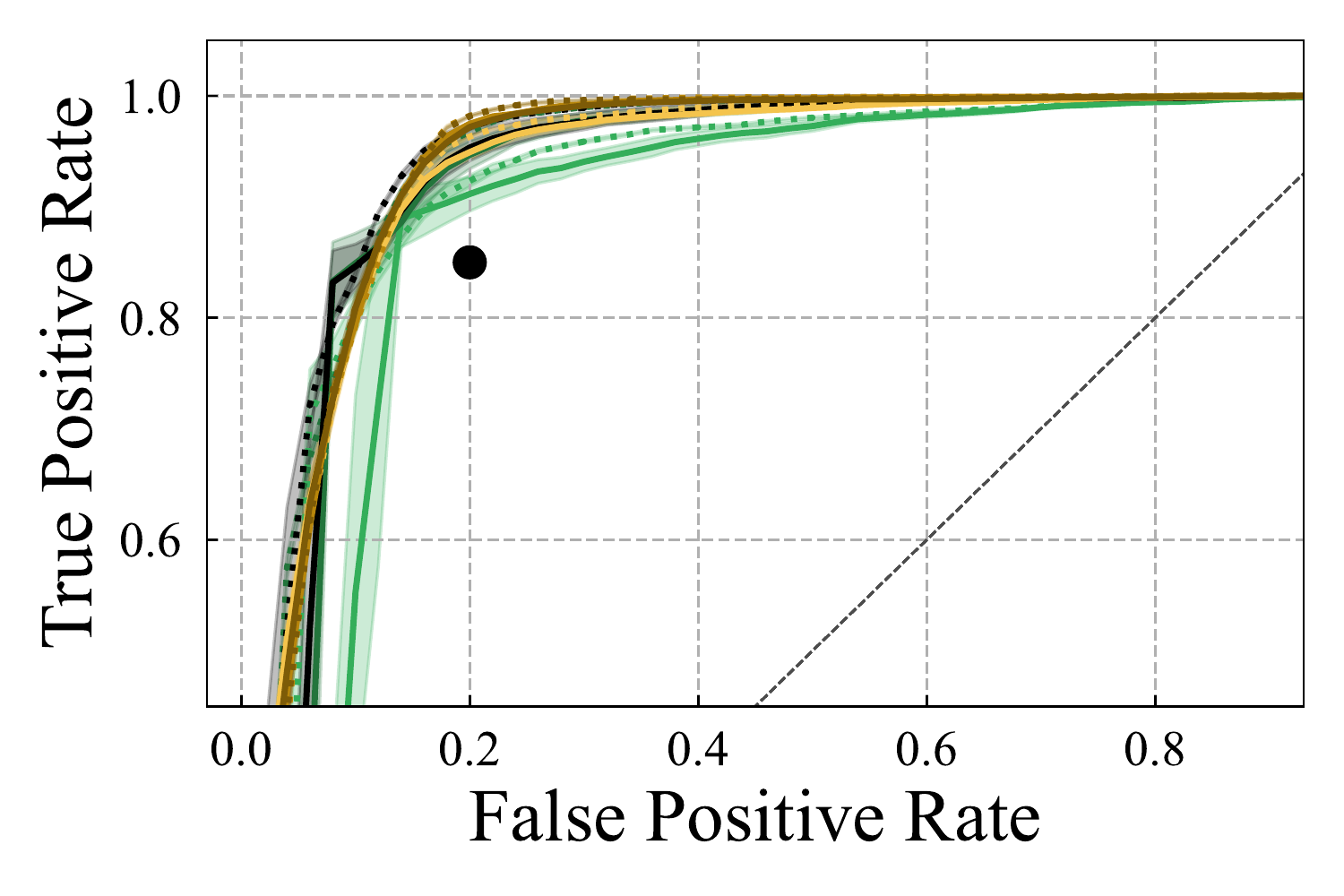}
    \vspace*{-7pt}
    \caption{
        \textbf{ROC: Country Shift\\$~$}
    }
\end{subfigure}
\begin{subfigure}[l]{0.24\linewidth}
    \hspace*{-10pt}
    \includegraphics[width=\linewidth]{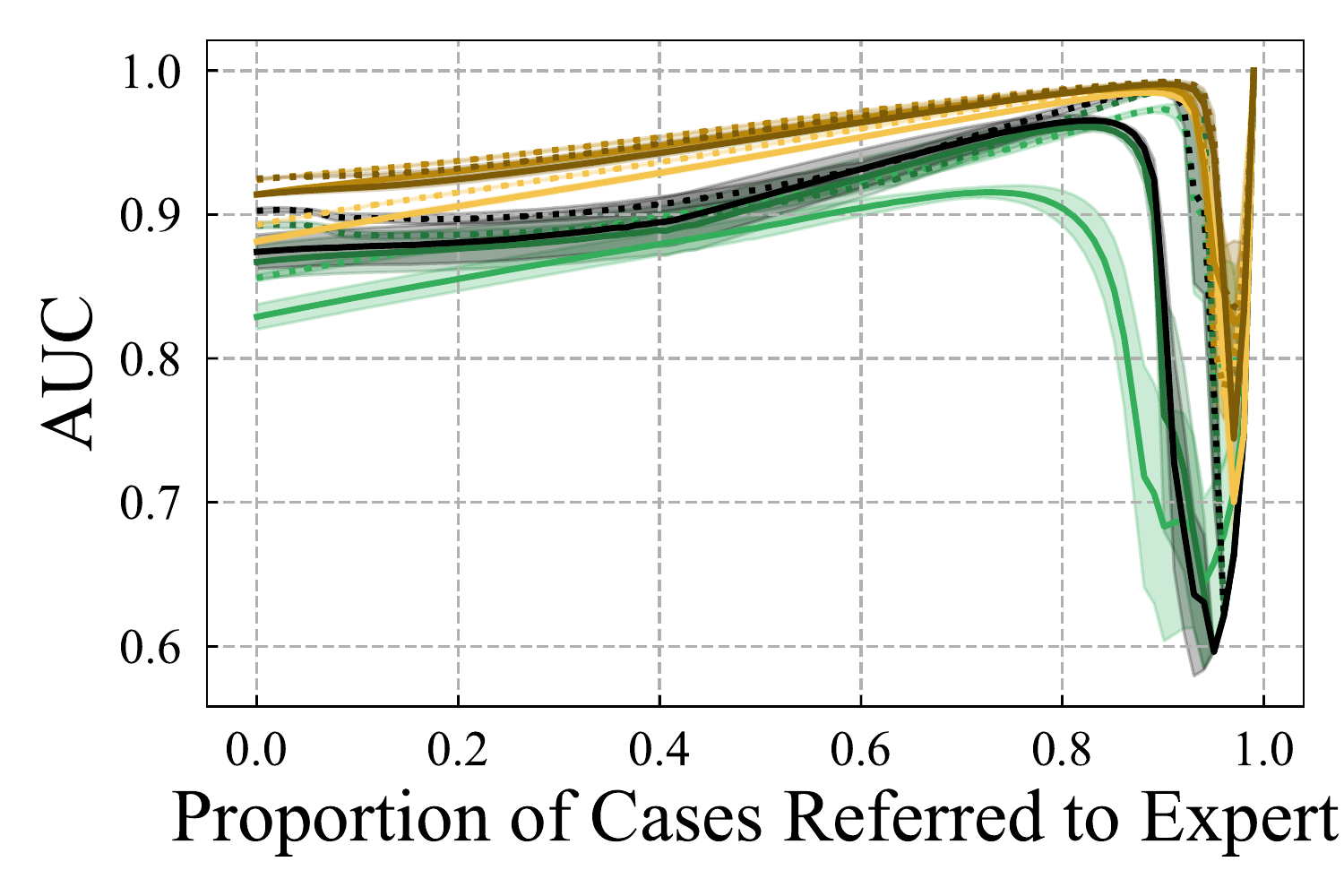}
    \vspace*{-7pt}
    \caption{
        \centering\textbf{Selective Prediction AUC: In-Domain}
    }
\end{subfigure}
\begin{subfigure}[r]{0.24\linewidth}
    \hspace*{-10pt}
    \includegraphics[width=\linewidth]{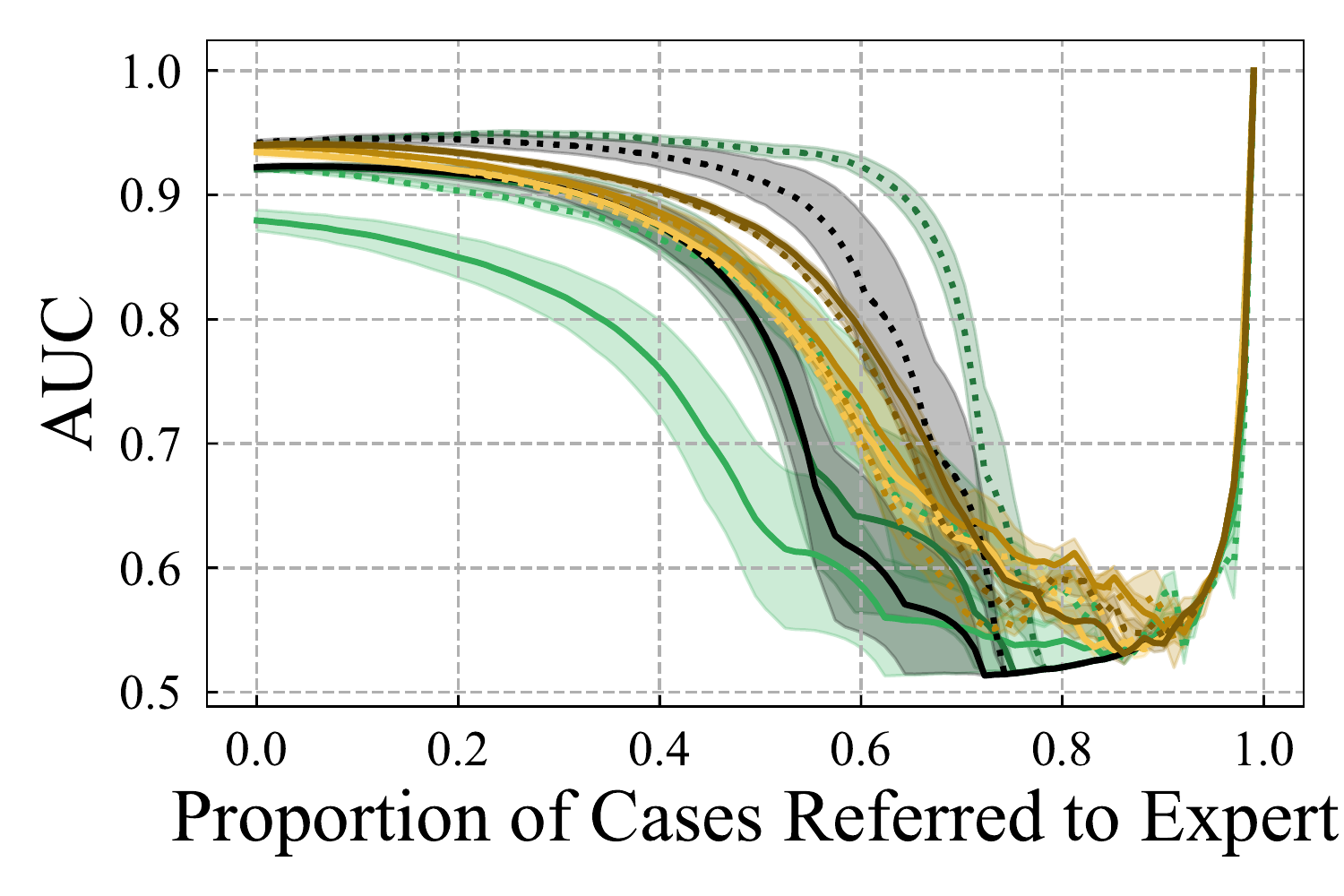}
    \vspace*{-7pt}
    \caption{
        \centering\textbf{Selective Prediction AUC: Country Shift}
    }
\end{subfigure}
\begin{subfigure}{\linewidth}
    \includegraphics[width=\linewidth]{fig/big_font_plots_rebuttal/retention/legend.pdf}
\end{subfigure}
\vspace*{2pt}
\centering
\caption{
    \textbf{Country Shift, Varying Blur Constant.} 
    We consider how preprocessing affects model predictive performance and uncertainty quantification on both in-domain and distributionally shifted data.
    \textbf{Left:} The \emph{receiver operating characteristic curve (ROC)} for in-population diagnosis on the \citet{kaggle_2015} test set (\textbf{a}) and for changing medical equipment and patient populations on the \citet{APTOS_2019} test set (\textbf{b}). 
    The dot in 
    \textbf{black}
    denotes the NHS-recommended 85\% sensitivity and 80\% specificity ratios \cite{widdowson2016management}.
    \textbf{Right:} \emph{selective prediction} on AUC in the \citet{kaggle_2015} (\textbf{c}) and the \citet{APTOS_2019} (\textbf{d}) settings.
    Shading denotes standard error computed over six random seeds.
    We vary the standard deviation hyperparameter of the Gaussian blur kernel through a \texttt{blur\_constant} (e.g., blur5 below corresponds to $\texttt{blur\_constant} = 5$). A higher \texttt{blur\_constant} results in a stronger smoothing of the image as per the preprocessing procedure outlined in \Cref{subsec:app_preprocessing_exps}.
    The default \texttt{blur\_constant} used in other experiments throughout this work is 30.
}
\label{fig:country_shift_blur}
\vspace*{-5pt}
\end{figure}

\textbf{Country Shift: Varying Blur Constant (\Cref{fig:country_shift_blur},~\Cref{tab:metrics_preproc_country}).}~~~Similarly to the Severity Shift results, higher \texttt{blur\_constant} tends to perform better on the in-domain evaluation data across methods and referral rates.
Notably, on the distributionally shifted APTOS data, \textsc{Deep Ensemble} outperforms \textsc{MC Dropout Ensemble}, and $\texttt{blur\_constant} = 20$ significantly improves performance from the default $\texttt{blur\_constant} = 30$ for \textsc{Deep Ensemble} between referral rates 0.4 and 0.7.
For example, for \textsc{Deep Ensemble} at $\tau = 0.7$, we observe $82.2 \pm 2.5$ AUC with $\texttt{blur\_constant} = 20$ versus $67.4 \pm 5.6$ AUC with $\texttt{blur\_constant} = 30$.

\clearpage


\clearpage

\begin{table*}[!htb]
\centering
\caption{
    \textbf{Severity Shift, Varying Blur Constant.}
    We consider how preprocessing affects downstream prediction and uncertainty quality of baseline methods in terms of the area under the receiver operating characteristic curve (AUC) and classification accuracy, as a function of the proportion of data referred to a medical expert for further review.
    All methods are tuned on in-domain validation AUC, and ensembles have $K = 3$ constituent models.
    We vary the standard deviation hyperparameter of the Gaussian blur kernel through a \texttt{blur\_constant} (e.g., blur5 below corresponds to $\texttt{blur\_constant} = 5$). A higher \texttt{blur\_constant} results in a stronger smoothing of the image as per the preprocessing procedure outlined in \Cref{subsec:app_preprocessing_exps}.
    The default \texttt{blur\_constant} used in other experiments is 30.
}
\vspace{-3pt}
\resizebox{1.0\linewidth}{!}{%
\begin{tabular}{@{\extracolsep{2pt}}lcccccc@{}}
\midrule
\midrule
& \multicolumn{2}{c}{No Referral} & \multicolumn{2}{c}{$50\%$ Data Referred} & \multicolumn{2}{c}{$70\%$ Data Referred} \\
\cline{2-3}
\cline{4-5}
\cline{6-7}\\
\textbf{Method}         &
\textbf{AUC (\%) $\uparrow$}            &
\textbf{Accuracy (\%) $\uparrow$}       &
\textbf{AUC (\%) $\uparrow$}            &
\textbf{Accuracy (\%) $\uparrow$}       &
\textbf{AUC (\%) $\uparrow$}            &
\textbf{Accuracy $\uparrow$}       \\
\midrule
\multicolumn{7}{c}{In-Domain (No, Mild, or Moderate DR, Clinical Labels \{0,1,2\})}\\
\midrule

\map (Deterministic)-blur5      & $73.7\pms{1.3}$ & $79.4\pms{1.4}$ & $75.5\pms{3.1}$ & $89.0\pms{0.9}$ & $79.1\pms{3.4}$ & $89.3\pms{1.0}$ \\
\map (Deterministic)-blur10     & $78.7\pms{1.1}$ & $84.6\pms{0.6}$ & $80.0\pms{2.3}$ & $93.4\pms{0.3}$ & $84.5\pms{2.2}$ & $94.1\pms{0.4}$ \\
\map (Deterministic)-blur20     & $79.9\pms{1.3}$ & $87.3\pms{0.5}$ & $77.2\pms{3.4}$ & $94.5\pms{0.4}$ & $80.9\pms{4.1}$ & $95.3\pms{0.3}$ \\
\map (Deterministic)-blur30    & $82.0\pms{1.0}$ & $87.9\pms{0.4}$ & $83.1\pms{1.9}$ & $95.2\pms{0.3}$ & $88.4\pms{1.9}$ & $96.0\pms{0.2}$ \\
\hdashline\noalign{\vskip 0.5ex}
\mcd-blur5      & $84.8\pms{0.4}$ & $76.1\pms{2.3}$ & $91.4\pms{0.3}$ & $86.0\pms{2.4}$ & $94.1\pms{0.4}$ & $88.6\pms{2.2}$ \\
\mcd-blur10     & $86.3\pms{0.1}$ & $84.2\pms{1.3}$ & $92.4\pms{0.4}$ & $93.5\pms{0.8}$ & $95.2\pms{0.2}$ & $95.1\pms{0.6}$ \\
\mcd-blur20     & $88.7\pms{0.3}$ & $90.1\pms{0.2}$ & $92.5\pms{0.5}$ & $97.0\pms{0.1}$ & $95.3\pms{0.3}$ & $97.7\pms{0.1}$ \\
\mcd-blur30    & $89.2\pms{0.2}$ & $90.5\pms{0.1}$ & $92.8\pms{0.6}$ & $97.2\pms{0.0}$ & $95.4\pms{0.4}$ & $97.8\pms{0.0}$ \\
\hdashline\noalign{\vskip 0.5ex}
\textsc{deep ensemble}-blur5    & $78.6\pms{0.6}$ & $84.3\pms{0.8}$ & $75.0\pms{2.6}$ & $93.3\pms{0.5}$ & $75.9\pms{3.3}$ & $94.8\pms{0.3}$ \\
\textsc{deep ensemble}-blur10   & $82.4\pms{0.3}$ & $87.7\pms{0.1}$ & $80.9\pms{1.3}$ & $95.1\pms{0.1}$ & $84.1\pms{1.3}$ & $96.1\pms{0.1}$ \\
\textsc{deep ensemble}-blur20   & $84.2\pms{0.8}$ & $88.6\pms{0.3}$ & $70.9\pms{1.1}$ & $95.8\pms{0.2}$ & $71.4\pms{1.4}$ & $96.7\pms{0.2}$ \\
\textsc{deep ensemble}-blur30  & $85.1\pms{0.7}$ & $89.3\pms{0.2}$ & $82.0\pms{0.9}$ & $96.3\pms{0.2}$ & $85.3\pms{0.9}$ & $97.3\pms{0.2}$ \\
\hdashline\noalign{\vskip 0.5ex}
\mcd \textsc{ensemble}-blur5    & $86.5\pms{0.1}$ & $79.4\pms{1.0}$ & $93.2\pms{0.1}$ & $90.2\pms{1.1}$ & $95.7\pms{0.2}$ & $92.5\pms{0.9}$ \\
\mcd \textsc{ensemble}-blur10   & $87.5\pms{0.0}$ & $86.7\pms{0.6}$ & $\mathbf{93.4\pms{0.2}}$ & $95.4\pms{0.3}$ & $\mathbf{96.0\pms{0.2}}$ & $96.5\pms{0.3}$ \\
\mcd \textsc{ensemble}-blur20   & $90.3\pms{0.0}$ & $91.1\pms{0.1}$ & $\mathbf{93.5\pms{0.2}}$ & $97.6\pms{0.0}$ & $\mathbf{96.0\pms{0.1}}$ & $\mathbf{98.2\pms{0.0}}$ \\
\mcd \textsc{ensemble}-blur30  & $\mathbf{90.6\pms{0.0}}$ & $\mathbf{91.4\pms{0.1}}$ & $93.1\pms{0.2}$ & $\mathbf{97.8\pms{0.0}}$ & $95.7\pms{0.2}$ & $\mathbf{98.2\pms{0.0}}$ \\

\midrule
\multicolumn{7}{c}{Severity Shift (Severe or Proliferate DR, Clinical Labels \{3, 4\})}\\
\midrule

\map (Deterministic)-blur5      & $-$ & $70.8\pms{6.2}$ & $-$ & $81.4\pms{7.9}$ & $-$ & $87.7\pms{7.9}$ \\
\map (Deterministic)-blur10     & $-$ & $77.3\pms{2.2}$ & $-$ & $91.9\pms{2.8}$ & $-$ & $97.2\pms{1.5}$ \\
\map (Deterministic)-blur20     & $-$ & $69.1\pms{4.0}$ & $-$ & $81.8\pms{5.3}$ & $-$ & $88.8\pms{4.4}$ \\
\map (Deterministic)-blur30    & $-$ & $74.4\pms{1.9}$ & $-$ & $93.2\pms{2.6}$ & $-$ & $98.6\pms{1.1}$ \\
\hdashline\noalign{\vskip 0.5ex}
\mcd-blur5      & $-$ & $93.5\pms{0.6}$ & $-$ & $\mathbf{100.0\pms{0.0}}$ & $-$ & $\mathbf{100.0\pms{0.0}}$ \\
\mcd-blur10     & $-$ & $91.0\pms{1.3}$ & $-$ & $99.9\pms{0.0}$ & $-$ & $\mathbf{100.0\pms{0.0}}$ \\
\mcd-blur20     & $-$ & $87.2\pms{0.9}$ & $-$ & $99.7\pms{0.1}$ & $-$ & $\mathbf{100.0\pms{0.0}}$ \\
\mcd-blur30    & $-$ & $86.4\pms{1.3}$ & $-$ & $99.5\pms{0.2}$ & $-$ & $\mathbf{100.0\pms{0.0}}$ \\
\hdashline\noalign{\vskip 0.5ex}
\textsc{deep ensemble}-blur5    & $-$ & $72.0\pms{3.9}$ & $-$ & $85.1\pms{3.7}$ & $-$ & $87.5\pms{3.3}$ \\
\textsc{deep ensemble}-blur10   & $-$ & $80.0\pms{1.2}$ & $-$ & $94.0\pms{1.0}$ & $-$ & $97.8\pms{0.5}$ \\
\textsc{deep ensemble}-blur20   & $-$ & $69.8\pms{2.1}$ & $-$ & $82.4\pms{1.5}$ & $-$ & $89.1\pms{1.5}$ \\
\textsc{deep ensemble}-blur30  & $-$ & $74.5\pms{1.2}$ & $-$ & $89.8\pms{1.0}$ & $-$ & $97.0\pms{0.7}$ \\
\hdashline\noalign{\vskip 0.5ex}
\mcd \textsc{ensemble}-blur5    & $-$ & $\mathbf{94.7\pms{0.3}}$ & $-$ & $\mathbf{100.0\pms{0.0}}$ & $-$ & $\mathbf{100.0\pms{0.0}}$ \\
\mcd \textsc{ensemble}-blur10   & $-$ & $91.9\pms{0.7}$ & $-$ & $\mathbf{100.0\pms{0.0}}$ & $-$ & $\mathbf{100.0\pms{0.0}}$ \\
\mcd \textsc{ensemble}-blur20   & $-$ & $88.6\pms{0.4}$ & $-$ & $99.8\pms{0.0}$ & $-$ & $\mathbf{100.0\pms{0.0}}$ \\
\mcd \textsc{ensemble}-blur30  & $-$ & $87.4\pms{0.3}$ & $-$ & $99.4\pms{0.1}$ & $-$ & $\mathbf{100.0\pms{0.0}}$ \\

\midrule
\midrule
\end{tabular}
}
\label{tab:metrics_preproc_severity}
\end{table*}

\clearpage

\begin{table*}[!htb] %
\vspace{-2mm}
\centering
\caption{
    \textbf{Country Shift, Varying Blur Constant.}
    We consider how preprocessing affects downstream prediction and uncertainty quality of baseline methods in terms of the area under the receiver operating characteristic curve (AUC) and classification accuracy, as a function of the proportion of data referred to a medical expert for further review.
    All methods are tuned on in-domain validation AUC, and ensembles have $K = 3$ constituent models.
    We vary the standard deviation hyperparameter of the Gaussian blur kernel through a \texttt{blur\_constant} (e.g., blur5 below corresponds to $\texttt{blur\_constant} = 5$). A higher \texttt{blur\_constant} results in a stronger smoothing of the image as per the preprocessing procedure outlined in \Cref{subsec:app_preprocessing_exps}.
    The default \texttt{blur\_constant} used in other experiments is 30.
}
\vspace*{-5pt}
\resizebox{1.0\linewidth}{!}{%
\begin{tabular}{@{\extracolsep{2pt}}lcccccc@{}}
\midrule
\midrule
& \multicolumn{2}{c}{No Referral} & \multicolumn{2}{c}{$50\%$ Data Referred} & \multicolumn{2}{c}{$70\%$ Data Referred} \\
\cline{2-3}
\cline{4-5}
\cline{6-7}\\
\textbf{Method}         &
\textbf{AUC (\%) $\uparrow$}            &
\textbf{Accuracy (\%) $\uparrow$}       &
\textbf{AUC (\%) $\uparrow$}            &
\textbf{Accuracy (\%) $\uparrow$}       &
\textbf{AUC (\%) $\uparrow$}            &
\textbf{Accuracy $\uparrow$}       \\
\midrule
\multicolumn{7}{c}{EyePACS Dataset (In-Domain)}\\
\midrule

\map (Deterministic)-blur5      & $82.9\pms{0.7}$ & $80.3\pms{0.8}$ & $89.1\pms{0.6}$ & $91.0\pms{0.7}$ & $91.4\pms{0.3}$ & $91.6\pms{0.6}$ \\
\map (Deterministic)-blur10     & $87.1\pms{0.1}$ & $85.6\pms{0.3}$ & $92.6\pms{0.1}$ & $95.0\pms{0.2}$ & $95.0\pms{0.2}$ & $94.9\pms{0.3}$ \\
\map (Deterministic)-blur20     & $86.7\pms{1.0}$ & $88.0\pms{0.5}$ & $90.5\pms{1.4}$ & $95.6\pms{0.3}$ & $94.4\pms{0.9}$ & $96.3\pms{0.2}$ \\
\map (Deterministic)-blur30    & $87.4\pms{1.0}$ & $88.6\pms{0.6}$ & $91.1\pms{1.4}$ & $95.9\pms{0.3}$ & $94.9\pms{0.8}$ & $96.5\pms{0.2}$ \\
\hdashline\noalign{\vskip 0.5ex}
\mcd-blur5      & $88.1\pms{0.2}$ & $85.9\pms{0.4}$ & $94.0\pms{0.1}$ & $95.0\pms{0.2}$ & $96.5\pms{0.1}$ & $96.4\pms{0.1}$ \\
\mcd-blur10     & $89.0\pms{0.2}$ & $85.5\pms{0.5}$ & $94.7\pms{0.2}$ & $94.9\pms{0.3}$ & $96.9\pms{0.1}$ & $96.3\pms{0.2}$ \\
\mcd-blur20     & $91.4\pms{0.1}$ & $90.2\pms{0.2}$ & $95.7\pms{0.2}$ & $97.3\pms{0.1}$ & $97.5\pms{0.1}$ & $98.0\pms{0.1}$ \\
\mcd-blur30    & $91.4\pms{0.1}$ & $90.9\pms{0.0}$ & $95.3\pms{0.2}$ & $97.4\pms{0.0}$ & $97.4\pms{0.1}$ & $98.1\pms{0.0}$ \\
\hdashline\noalign{\vskip 0.5ex}
\textsc{deep ensemble}-blur5    & $85.6\pms{0.2}$ & $84.6\pms{0.1}$ & $90.9\pms{0.3}$ & $94.3\pms{0.0}$ & $93.6\pms{0.3}$ & $95.8\pms{0.2}$ \\
\textsc{deep ensemble}-blur10   & $88.8\pms{0.0}$ & $88.0\pms{0.1}$ & $94.2\pms{0.1}$ & $96.2\pms{0.0}$ & $96.4\pms{0.1}$ & $97.3\pms{0.0}$ \\
\textsc{deep ensemble}-blur20   & $89.2\pms{0.2}$ & $89.5\pms{0.2}$ & $90.5\pms{0.3}$ & $96.9\pms{0.1}$ & $93.8\pms{0.3}$ & $97.7\pms{0.0}$ \\
\textsc{deep ensemble}-blur30  & $90.3\pms{0.1}$ & $90.3\pms{0.2}$ & $91.7\pms{0.5}$ & $97.2\pms{0.0}$ & $95.0\pms{0.4}$ & $97.9\pms{0.0}$ \\
\hdashline\noalign{\vskip 0.5ex}
\mcd \textsc{ensemble}-blur5    & $89.3\pms{0.0}$ & $87.3\pms{0.1}$ & $94.7\pms{0.0}$ & $95.7\pms{0.1}$ & $97.1\pms{0.0}$ & $96.9\pms{0.0}$ \\
\mcd \textsc{ensemble}-blur10   & $90.1\pms{0.0}$ & $87.4\pms{0.1}$ & $95.4\pms{0.0}$ & $96.0\pms{0.0}$ & $97.3\pms{0.0}$ & $97.0\pms{0.1}$ \\
\mcd \textsc{ensemble}-blur20   & $92.4\pms{0.0}$ & $91.2\pms{0.0}$ & $\mathbf{96.2\pms{0.1}}$ & $97.7\pms{0.0}$ & $\mathbf{97.9\pms{0.0}}$ & $98.3\pms{0.0}$ \\
\mcd \textsc{ensemble}-blur30  & $\mathbf{92.5\pms{0.0}}$ & $\mathbf{91.6\pms{0.0}}$ & $95.8\pms{0.1}$ & $\mathbf{97.8\pms{0.0}}$ & $97.7\pms{0.1}$ & $\mathbf{98.4\pms{0.0}}$ \\

\midrule
\multicolumn{7}{c}{APTOS 2019 Dataset (Shifted)}\\
\midrule

\map (Deterministic)-blur5      & $87.9\pms{0.7}$ & $69.9\pms{1.4}$ & $64.0\pms{5.3}$ & $78.6\pms{1.7}$ & $55.3\pms{3.2}$ & $78.9\pms{1.9}$ \\
\map (Deterministic)-blur10     & $90.2\pms{0.2}$ & $77.0\pms{0.7}$ & $63.1\pms{2.0}$ & $81.1\pms{0.6}$ & $51.1\pms{0.0}$ & $80.0\pms{0.6}$ \\
\map (Deterministic)-blur20     & $92.1\pms{0.2}$ & $85.2\pms{0.3}$ & $79.8\pms{3.8}$ & $87.9\pms{1.5}$ & $60.0\pms{4.6}$ & $86.0\pms{1.2}$ \\
\map (Deterministic)-blur30    & $92.2\pms{0.2}$ & $86.2\pms{0.4}$ & $80.1\pms{2.8}$ & $87.6\pms{1.1}$ & $55.4\pms{3.3}$ & $85.4\pms{0.9}$ \\
\hdashline\noalign{\vskip 0.5ex}
\mcd-blur5      & $93.4\pms{0.2}$ & $78.4\pms{0.7}$ & $82.2\pms{0.4}$ & $84.6\pms{0.1}$ & $62.5\pms{0.6}$ & $88.0\pms{0.5}$ \\
\mcd-blur10     & $93.3\pms{0.2}$ & $77.3\pms{0.9}$ & $79.7\pms{0.3}$ & $83.3\pms{0.3}$ & $59.6\pms{1.0}$ & $87.2\pms{0.4}$ \\
\mcd-blur20     & $93.9\pms{0.1}$ & $84.9\pms{0.4}$ & $83.8\pms{1.2}$ & $86.2\pms{0.6}$ & $63.8\pms{2.4}$ & $87.9\pms{0.2}$ \\
\mcd-blur30    & $94.0\pms{0.2}$ & $86.8\pms{0.2}$ & $87.4\pms{0.3}$ & $88.1\pms{0.2}$ & $65.3\pms{1.3}$ & $88.2\pms{0.3}$ \\
\hdashline\noalign{\vskip 0.5ex}
\textsc{deep ensemble}-blur5    & $92.1\pms{0.1}$ & $70.8\pms{0.6}$ & $82.5\pms{1.7}$ & $85.0\pms{0.3}$ & $63.2\pms{4.2}$ & $87.1\pms{0.6}$ \\
\textsc{deep ensemble}-blur10   & $91.8\pms{0.0}$ & $78.8\pms{0.3}$ & $73.5\pms{0.3}$ & $84.5\pms{0.1}$ & $51.1\pms{0.0}$ & $82.0\pms{0.1}$ \\
\textsc{deep ensemble}-blur20   & $\mathbf{94.1\pms{0.0}}$ & $87.0\pms{0.1}$ & $\mathbf{93.7\pms{0.4}}$ & $\mathbf{93.6\pms{0.3}}$ & $\mathbf{82.2\pms{2.5}}$ & $\mathbf{91.7\pms{0.5}}$ \\
\textsc{deep ensemble}-blur30  & $94.2\pms{0.2}$ & $87.5\pms{0.1}$ & $91.2\pms{1.4}$ & $92.4\pms{0.7}$ & $67.4\pms{5.6}$ & $90.1\pms{0.9}$ \\
\hdashline\noalign{\vskip 0.5ex}
\mcd \textsc{ensemble}-blur5    & $93.7\pms{0.1}$ & $80.1\pms{0.3}$ & $81.9\pms{0.2}$ & $84.7\pms{0.1}$ & $63.2\pms{0.4}$ & $87.2\pms{0.2}$ \\
\mcd \textsc{ensemble}-blur10   & $93.6\pms{0.1}$ & $78.7\pms{0.4}$ & $79.1\pms{0.1}$ & $83.4\pms{0.1}$ & $59.6\pms{0.2}$ & $87.3\pms{0.3}$ \\
\mcd \textsc{ensemble}-blur20   & $94.0\pms{0.0}$ & $86.4\pms{0.3}$ & $83.3\pms{0.7}$ & $85.7\pms{0.3}$ & $58.3\pms{0.9}$ & $87.6\pms{0.1}$ \\
\mcd \textsc{ensemble}-blur30  & $94.1\pms{0.1}$ & $\mathbf{87.6\pms{0.1}}$ & $86.8\pms{0.2}$ & $88.0\pms{0.1}$ & $62.3\pms{0.3}$ & $87.7\pms{0.2}$ \\

\midrule
\midrule
\end{tabular}
}
\label{tab:metrics_preproc_country}
\vspace*{-15pt}
\end{table*}

\clearpage

\end{appendices}

\end{document}